\documentclass[11pt]{article}       
\usepackage{geometry}               
\usepackage{graphicx}
\usepackage{caption}
\usepackage{subcaption}
\usepackage{enumerate}
\usepackage{hyperref}
\usepackage{cite}

\usepackage{amsfonts}
\usepackage{epstopdf}
\usepackage{caption}
\usepackage{amsmath} 
\usepackage{amssymb}  
\usepackage{mathtools}
\usepackage{amsthm} 
\usepackage{bbm}
\usepackage{comment}

\usepackage{bm}
\usepackage{lipsum}
\usepackage{algorithm}
\usepackage{algpseudocode}
\usepackage{color}
\usepackage{array}
\usepackage{multirow}
\usepackage{cite}
\usepackage{url}
\usepackage{booktabs}
\usepackage{tikz}
\usetikzlibrary{shapes.geometric, arrows}

\theoremstyle{plain}
\newtheorem{thm}{Theorem}[section]
\newtheorem{prop}[thm]{Proposition}
\newtheorem{lem}[thm]{Lemma}

\theoremstyle{definition}
\newtheorem{defn}[thm]{Definition}
\newtheorem{assumption}[thm]{Assumption}
\theoremstyle{remark}
\newtheorem{rem}[thm]{Remark}

\newcommand{\E}{\mathbb{E}}

\newcommand{\R}{\mathbb{R}}

\newcommand{\x}{\bm{x}}

\DeclareMathOperator*{\argmin}{arg\,min}

\title{Approximate Thompson Sampling for Learning\\ Linear Quadratic Regulators with $O(\sqrt{T})$ Regret\thanks{The first two authors contributed equally. This work was supported in part by the Information and Communications Technology Planning and Evaluation
(IITP) grants funded by MSIT No. 2022-0-00124, No. 2022-0-00480 and No. RS-2021-II211343, Artificial Intelligence
Graduate School Program (Seoul National University). 
}}

\author{Yeoneung Kim\thanks{
 Y. Kim is with the Department of Applied Artificial Intelligence, Seoul National University of Science and Technology, Seoul, 01811, South Korea. {\tt\small  yeoneung@seoultech.ac.kr}}
 \and
Gihun Kim \and
Jiwhan Park \and
 Insoon Yang
\thanks{G. Kim, J. Park, and I. Yang are with the Department of Electrical and Computer Engineering and ASRI, Seoul National University, Seoul, 08826, South Korea. {\tt\small  \{hoon2680, jiwhanpark, insoonyang\}@snu.ac.kr}}
}

\date{}

\begin{document}
\maketitle

\pagestyle{myheadings}
\thispagestyle{plain}

\begin{abstract}
We propose a novel Thompson sampling algorithm that learns linear quadratic regulators (LQR) with a Bayesian regret bound of $O(\sqrt{T})$. 
Our method leverages Langevin dynamics with a carefully designed preconditioner and incorporates a simple excitation mechanism. 
We show that the excitation signal drives the minimum eigenvalue of the preconditioner to grow over time, thereby accelerating the approximate posterior sampling process. 
Furthermore, we establish nontrivial concentration properties of the approximate posteriors generated by our algorithm. 
These properties enable us to bound the moments of the system state and attain an $O(\sqrt{T})$ regret bound without relying on the restrictive assumptions that are often used in the literature.
\end{abstract}

\section{Introduction}\label{sec:intro}
Balancing the exploration-exploitation trade-off is a fundamental challenge in reinforcement learning (RL) because in most cases, there is no clear criterion to choose between acting to learn about the unknown environment (`exploration') or making a reward-maximizing decision given the information gathered thus far (`exploitation'). This dilemma has been systematically addressed by two principal approaches: \emph{optimism in the face of uncertainty} (OFU) and \emph{Thompson sampling} (TS). OFU-based methods construct confidence sets for the environment or model parameters using the data observed thus far. An optimistic or reward-maximizing set of parameters is then selected from within this confidence set, and a corresponding optimal policy is executed~\cite{lai1985asymptotically}. Algorithms based on OFU have been shown to provide strong theoretical guarantees, particularly in the context of bandit problems~\cite{kearns2002near}. On the other hand, TS is a Bayesian method in which the environment or model parameters are sampled from a posterior distribution that is updated over time using observed data and a prior~\cite{thompson1933likelihood}. An optimal policy with respect to the sampled parameters is then constructed and executed. TS is often more computationally tractable than OFU, as OFU typically requires solving a nonconvex optimization problem over a confidence set in each episode. TS has demonstrated effectiveness in online learning across a wide range of sequential decision-making problems, including multi-armed bandits~\cite{agrawal2012analysis, agrawal2013thompson, kaufmann2012thompson}, Markov decision processes~\cite{osband2013more, osband2016posterior, gopalan2015thompson}, and LQR problems~\cite{ouyang2019posterior, osband2016posterior, abbasi2015bayesian, abeille2017thompson, faradonbeh2020adaptive}.

 In TS-based online learning, posterior sampling becomes challenging in high-dimensional settings. It is also computationally intractable when the posterior distribution lacks a closed-form expression, which occurs when the noise and prior distributions are not conjugate. To address this, Markov Chain Monte Carlo (MCMC) methods---particularly Langevin MCMC---have been proposed~\cite{gilks1995markov, roberts1996exponential, durmus2016sampling, welling2011bayesian}. With these theoretical foundations, there have been attempts to leverage Langevin MCMC to effectively solve contextual bandit problems~\cite{huix2023tight,xu2022langevin,mazumdar2020thompson} and MDPs~\cite{ishfaq2023provable,karbasi2023langevin}. Nevertheless, Langevin MCMC is computationally intensive. To mitigate this issue, various acceleration techniques have been studied (see~\cite{welling2011bayesian, li2019stochastic, mou2019high, ding2021random, lu2019accelerating} and references therein). In particular, preconditioning has been shown to be effective for improving sampling efficiency~\cite{welling2011bayesian, girolami2011riemann, dalalyan2017theoretical, dwivedi2018log}. Motivated by these findings, we incorporate preconditioned Langevin MCMC into TS for LQR problems.

\subsection{Related work}
There is a rich body of literature regarding regret analysis for online learning of LQR problems, which are categorized as follows.

\textbf{Certainty equivalence (CE)}:
The certainty equivalence principle~\cite{landau1998adaptive} has been widely adopted for learning dynamical systems with unknown transitions, wherein the optimal policy is designed under the assumption that the estimated system parameters accurately represent the true parameters. The performance of CE-based methods has been extensively studied across various settings, including online learning~\cite{simchowitz2020naive, dean2018regret, mania2019certainty, jedra2022minimal}, sample complexity analysis~\cite{dean2020sample}, finite-time stabilization~\cite{faradonbeh2018finite}, and asymptotic regret bounds~\cite{faradonbeh2020adaptive}.

\textbf{Optimism in the face of uncertainty (OFU)}: \cite{abbasi2011regret,ibrahimi2012efficient} proposed OFU-based learning algorithms that iteratively select high-performing control actions while constructing confidence sets. These methods achieve a frequentist regret bound of $\tilde O(\sqrt{T})$, but are often computationally impractical due to the complexity of the resulting constraints. To address this issue, subsequent works~\cite{cohen2019learning,abeille2020efficient} translated the nonconvex optimization problem inherent in OFU into a semidefinite programming (SDP) formulation, attaining the same $\tilde O(\sqrt{T})$ regret bound with high probability. Alternatively, \cite{faradonbeh2020adaptive,faradonbeh2020input} introduced randomized control actions to avoid constructing confidence sets, while still achieving an asymptotic regret bound of $\tilde{O}(\sqrt{T})$. More recently, \cite{lale2022reinforcement} proposed an algorithm that rapidly stabilizes the system and attains a $\tilde{O}(\sqrt{T})$ frequentist regret bound without requiring a stabilizing control gain matrix. 

\textbf{Thompson sampling (TS)}: It has been shown that the upper bound for the frequentist regret under Gaussian noise can be as large as $\tilde O(T^{2/3})$~\cite{abeille2017thompson}, which was later improved to $\tilde O(\sqrt{T})$ in~\cite{abeille2018improved} using a TS-based approach; however, this result is limited to \textit{scalar} systems. Subsequently, \cite{kargin2022thompson} extended the analysis to multidimensional systems, achieving a $\tilde O(\sqrt{T})$ frequentist regret bound. Nonetheless, the Gaussian noise assumption remains essential for establishing these guarantees. For the Bayesian regret bound, prior results \cite{ouyang2019posterior,gagrani2022modified} demonstrate the potential of TS-based algorithms to achieve a $\tilde O(\sqrt{T})$ Bayesian regret bound. However, these methods are subject to several limitations. Specifically, both the noise and the prior distribution over system parameters are assumed to be Gaussian, ensuring conjugacy between the prior and posterior. Additionally, the columns of the system parameter matrix are assumed to be mutually independent.

\textbf{Comparison with~\cite{mazumdar2020thompson}:}
Our work builds on the ideas introduced in~\cite{mazumdar2020thompson}, which focuses on multi-armed bandits. However, key differences arise due to the fundamentally different nature of LQR problems. For example, in the bandit setting, the strong log-concavity of the reward function ensures linear growth of the likelihood function as more data is collected. This property plays a crucial role in their analysis. In contrast, such growth does not occur in LQR problems, prompting us to introduce an adaptive preconditioner to improve computational efficiency. Moreover, the Lipschitz smoothness of the log-reward function in~\cite{mazumdar2020thompson} facilitates the analysis of the gap between exact and approximate posteriors—a simplification that does not hold in the LQR setting.

\subsection{Contributions}
In this paper, we propose a computationally efficient approximate Thompson sampling algorithm for learning linear quadratic regulators (LQR) with a Bayesian regret bound of $O(\sqrt{T})$.\footnote{It is worth noting that the frequentist regret bound does not imply the Bayesian regret bound of the same order as the high-probability frequentist regret is converted into $\mathbb{E}[\mathrm{Regret}] \approx O  ((1-\delta)\sqrt{T \log(1/\delta)}+ \delta \exp(T) )$ with confidence $\delta>0$. Here, simply taking $\delta=\exp(-T)$ will increase the order of $T$ in the leading term. To achieve the $O(\sqrt{T})$ Bayesian regret by taking the expectation on all feasible values of system parameters, it is necessary to estimate the exponential growth of the system state over the time horizon. As this growth can quickly lead to a polynomial-in-time regret bound, one crucial aspect of addressing this challenge is the need for controlling the tail probability in an effective manner. By ensuring that the tail probability is controlled properly, we mitigate the risk of exponential growth of system state, thereby maintaining stability and performance within acceptable bounds. Thus, obtaining a tight estimate of the tail probability is instrumental when employing Langevin MCMC for TS.} Our algorithm is based on carefully designed Langevin dynamics that achieve an improved convergence rate. The regret analysis is conducted under the assumption that the system noise follows a strongly log-concave distribution—a relaxation of the Gaussian noise assumption commonly adopted in prior works. To the best of our knowledge, our method achieves the tightest known Bayesian regret bound for online LQR learning, improving upon the existing $\tilde{O}(\sqrt{T})$ bounds\footnote{Here, $\tilde{O}(\cdot)$ hides logarithmic factors.} in the literature~\cite{abeille2018improved, ouyang2019posterior, gagrani2022modified}.

It is worth noting that in~\cite{ouyang2019posterior, gagrani2022modified}, the system noise is assumed to follow independent and identically distributed Gaussian. Moreover, the columns of the system parameter matrix are assumed to be mutually independent and Gaussian in the prior, which is key to both the tractability of their regret analysis and the simplification of posterior updates. In contrast, our work not only achieves a tighter regret bound but also relaxes these restrictive assumptions. While we adopt the assumption on system parameters from~\cite{abeille2018improved}, we go beyond their analysis by establishing a regret bound that holds for multi-dimensional systems.

The two key components of our method are: $(i)$ a preconditioned unadjusted Langevin algorithm (ULA) for approximate Thompson sampling, and $(ii)$ a simple excitation mechanism. The proposed excitation mechanism injects a noise signal into the control input at the end of each episode, which causes the minimum eigenvalue of the preconditioner to increase over time, thereby accelerating the posterior sampling process. We identify appropriate step sizes and iteration counts for the preconditioned Langevin MCMC and demonstrate both an accelerated convergence rate for approximate Thompson sampling and improved learning performance. Specifically, we show that the sampled system parameters converge to the true parameters at a rate of $\tilde{O}(t^{-\frac{1}{4}})$. This improvement yields a tighter bound on the system state norm, which in turn contributes to achieving the improved regret bound of $O(\sqrt{T})$.

\section{Preliminaries}
\label{sec:pre}

\subsection{Linear-Quadratic Regulators}

Consider a linear stochastic system of the form
\begin{equation}\label{sys}
x_{t+1} = Ax_t + B u_t + w_t, \quad t = 1, 2, \ldots,
\end{equation}
where $x_t \in \mathbb{R}^n$ is the system input, and $u_t \in \mathbb{R}^{n_u}$ is the control input. 
The disturbance $w_t \in \mathbb{R}^n$ is an independent and identically distributed (i.i.d.) zero-mean random vector with covariance matrix $\mathbf{W}$.
Throughout the paper, let $I_n$ denote the $n$ by $n$ identity matrix, let $|v|_P := \sqrt{v^{\top}Pv}$ be the weighted $2$-norm of a vector $v$ with respect to a positive semidefinite matrix $P$, let $|v|$ indicate the Euclidean norm, and let $|A|$ represent the spectral norm of a matrix $A$.

\begin{assumption}\label{ass:noise}
For every $t=1, 2, \ldots$, the random vector $w_t$ satisfies the following properties:
\begin{enumerate}
\item The probability density function (pdf) of noise $p_w (\cdot)$ is known and twice differentiable. Additionally, $\underline m I_n \preceq -\nabla^2  \log p_w(\cdot)  \preceq \overline m I_n.$ for some $\underline m, \overline m>0$.\footnote{The density of a multivariate normal distribution whose covariance $\Sigma$ lies between $\underline m$ and $\overline m$ satisfies this assumption.}
\item $\mathbb{E} [w_t ] = 0$ and $\mathbb{E} [w_t w_t^\top ] = \mathbf{W}$, where $\mathbf{W}$ is positive definite.
 \end{enumerate}
\end{assumption}
Our paper deals with a broader class of disturbances compared to existing methods~\cite{abeille2018improved,ouyang2019posterior,gagrani2022modified}, as any multivariate Gaussian distribution satisfies the assumption.

Let $d:= n+n_u$ and ${\Theta}$ be the system parameter matrix defined by
${\Theta} := \begin{bmatrix}  {\Theta}(1) & \cdots & {\Theta} (n)\end{bmatrix} := \begin{bmatrix} 
A & B
\end{bmatrix}^{\top} \in \mathbb{R}^{d \times n}$,
where ${\Theta} (i) \in \mathbb{R}^d$ is the $i$th column of ${\Theta}$. 
We also let 
${\theta} := \mathrm{vec}( \Theta) := ( {\Theta} (1),  {\Theta} (2), \ldots, {\Theta}(n)) \in \mathbb{R}^{dn}$ 
denote the vectorized version of ${\Theta}$. 
We often refer to $\theta$ as the parameter vector.

Let $h_t := (x_1, u_1, \ldots, x_{t-1}, u_{t-1}, x_t)$ be the \emph{history} of observations made up to time $t$, and let $H_t$ denote the collection of such histories at stage $t$. 
A (deterministic) policy $\pi_t$ maps history $h_t$ to action $u_t$, i.e., $\pi_t (h_t) = u_t$. The set of admissible policies is defined as 
$\Pi := \{\pi = (\pi_1, \pi_2, \ldots) \mid \pi_t: H_t \to \mathbb{R}^{n_u} \mbox{ is measurable $\forall t$} \}.$

The stage-wise cost is chosen to be a quadratic function of the form
$c (x_t, u_t) := x_t^\top Q x_t + u_t^\top R u_t$,
where $Q \in \mathbb{R}^{n\times n}$ is symmetric positive semidefinite and $R \in \mathbb{R}^{n_u\times n_u}$ is symmetric positive definite. 
The cost matrices $Q$ and $R$ are assumed to be known.\footnote{This assumption is common in the literature~\cite{abbasi2011regret,faradonbeh2020adaptive,abeille2020efficient,jedra2022minimal,kargin2022thompson,dean2020sample}.}
We consider the infinite-horizon average cost LQ setting with the following cost function:
\begin{equation}
J_\pi (\theta) := \limsup_{T \to \infty} \frac{1}{T} \mathbb{E}_\pi \bigg [
\sum_{t=1}^T c (x_t, u_t)
\bigg ].
\end{equation}
Given $\theta \in \mathbb{R}^{d  n}$, $\pi_*(x;\theta)$ denotes an optimal policy  if it exists, and the corresponding optimal cost is given by
$J (\theta) = \inf_{\pi \in \Pi} J_\pi (\theta)$. 
It is well known that the optimal policy and cost can be obtained using the Riccati equation under the standard stabilizability and observability assumptions (e.g.,~\cite{bertsekas2011dynamic}). 

\begin{thm}\label{lqr}
Suppose that $(A, B)$ is stabilizable,
and $(A, Q^{1/2})$ is observable. 
Then, the following algebraic Riccati equation (ARE) has a unique positive definite solution $P^*(\theta)$:
\begin{equation}\label{ric}
\begin{split}
P^*(\theta) &= Q + A^\top P^*(\theta) A - A^\top P^*(\theta) B (R + B^\top P^*(\theta) B)^{-1} B^\top P^*(\theta) A.
\end{split}
\end{equation}
Furthermore, the optimal cost function is given by
$J (\theta) = \mathrm{tr} (\mathbf{W} P^*(\theta))$,
which is  continuously differentiable with respect to $\theta$,
and
the optimal policy is uniquely obtained as
$\pi_* (x; \theta) = K(\theta) x$,
where the control gain matrix $K(\theta)$ is given by $K(\theta) := - (R + B^\top P^*(\theta) B)^{-1} B^\top P^*(\theta) A$. 
\end{thm}
The optimal policy, called the \emph{linear-quadratic regulator} (LQR), is an asymptotically stabilizing controller: it drives the closed-loop system state to the origin, that is, the spectrum of $A + BK (\theta)$ is contained in the interior of a unit circle~\cite{bertsekas2011dynamic}.

\subsection{Online learning of LQR}

The theory of LQR is applicable when the true system parameters ${\theta_*} := \mathrm{vec}( \Theta_*):= \mathrm{vec}(\begin{bmatrix} 
A_* & B_*
\end{bmatrix}^\top)$
are fully known and stabilizable.
However, we consider the case where the true parameter vector $\theta_*$ is unknown. 
Online learning is a popular approach to addressing this case~\cite{abbasi2011regret}. 
The performance of an online learning algorithm is typically measured by regret. In particular, we consider the Bayesian setting where the prior distribution $p_1$ of the true system parameter random variable $\bar{\theta}_*$ is assumed to be given, and define the Bayesian regret over $T$ stages as:
\begin{equation}\label{regret}
R(T) := \mathbb{E} \bigg [
\sum_{t=1}^T ( c (x_t, u_t) - J (\bar{\theta}_*))
\bigg ].
\end{equation}
The expectation is taken with respect to the distributions of system noise $(w_1, w_2, \ldots, w_T)$, the internal randomness of the learning algorithm, and the prior distribution since we only have the belief of true system parameters in the form of the prior distribution.
\subsection{Thompson sampling}

Thompson sampling (TS) or posterior sampling has been used in a large class of online learning problems~\cite{russo2017tutorial}. The naive TS algorithm for learning LQR  starts with sampling a system parameter from the posterior $\mu_k$ at the beginning of episode $k$. Considering this sample parameter as true, the control gain matrix $K(\theta_{k})$ is computed by solving the ARE~\eqref{ric}. During the episode, the control gain matrix is used to produce control action $u_t=K(\theta_k)x_t$, where $x_t$ is the system state observed at time $t$.  Along the way, the state-input data is collected and the posterior is updated using the dataset. 
We will use dynamic episodes meaning that the length of the episode increases as the learning proceeds. Specifically, the $k$th episode starts at $t=\frac{k(k+1)}{2}$ and the sampled system parameter is used   throughout the episode.

The posterior update is performed using Bayes’ rule and it preserves the log-concavity of distributions. To see this we let $z_t:=(x_t,u_t)\in\R^d$ and write $
p(x_{t+1} |z_t, \theta) = p_w (x_{t+1} - \Theta^{\top}  z_t )$,
which is log-concave with respect to $\theta$ under Assumption~\ref{ass:noise}. Hence, the posterior at stage $t$ is given as
\begin{equation}\label{posterior}
\begin{split}
p(\theta | h_{t+1}) &\propto p(x_{t+1} | z_t, \theta ) p(\theta |h_t) = p_w (x_{t+1} - \Theta^{\top}  z_t ) p(\theta  | h_t).
\end{split}
\end{equation}
Thus, if $p(\theta  | h_t)$ is log-concave, then so is $p(\theta | h_{t+1})$.

However, sampling from the posterior is computationally intractable particularly
when the distributions at hand are not conjugate. Without conjugacy, posterior distribution does not have a closed-form expression. A popular approach to resolving this issue is using Markov chain Monte Carlo (MCMC) type algorithm that can be used for posterior sampling in an approximate but tractable way as described in the following subsection.

\subsection{The unadjusted Langevin algorithm (ULA)}

Consider the problem of sampling from a probability distribution with density $p (x) \propto e^{-U(x)}$, where the potential $U: \mathbb{R}^{n_x} \to \mathbb{R}$ is twice differentiable. The Langevin dynamics take the form
\begin{equation}\label{eq:cont}
\mathrm{d}X_\tau = - \nabla U(X_\tau) \mathrm{d}\tau + \sqrt{2} \mathrm{d} B_\tau,
\end{equation}
where  $B_\tau$ is standard Brownian motion in $\mathbb{R}^{n_x}$.
It is well-known that given an arbitrary $X_0$, the pdf of $X_{\xi}$ converges to the target pdf $p(x)$ as $\xi \to \infty$~\cite{mou2019high,pavliotis2014stochastic}. 
To approximate $X_\tau$, we apply the Euler--Maruyama discretization to the Langevin diffusion, yielding the \emph{unadjusted Langevin algorithm} (ULA):
\begin{equation}\label{eq:ula}
X_{j+1} = X_j - \gamma_j \nabla U(X_j) + \sqrt{2\gamma_j} W_j,
\end{equation}
where $(W_j)_{j\geq 1}$ are i.i.d. standard $n_x$-dimensional Gaussian random vectors, and  $(\gamma_j)_{j\geq 1}$ are step  sizes. 
While Metropolis--Hastings corrections are often used to mitigate discretization error~\cite{roberts1996exponential,bou2013nonasymptotic}, small step sizes can eliminate the need for such adjustments. In this work, we propose adaptive step sizes and iteration counts that ensure improved concentration properties, as discussed in Section~\ref{sec:alg}.

The condition number of the Hessian of the potential is a key factor in determining the rate of convergence. More precisely, the following concentration
property of ULA holds, which is a modification of Theorem 5 in ~\cite{mazumdar2020thompson}. 
\begin{rem}\label{rem:shared}
It is important to note that if $X_0\sim e^{-U}$, then $X_t \sim ~ e^{-U}$ in~\eqref{eq:cont} for all $t$. Thus, we can regard the noise sequence in~\eqref{eq:ula} to achieve $X_N$ for $N\in\mathbb{N}$ as a realization of the continuous Brownian motion in~\eqref{eq:cont} up to time $\tau=\sum_{j=0}^{N-1}{\gamma_j}$, which is further specified in Appendix~\ref{app:naive_ula}.
\end{rem}
\begin{thm}\label{thm:naive_ula}
Suppose that the pdf $p(x) \propto e^{-U(x)}$ is strongly log-concave and $\lambda_{\min} I\preceq \nabla^2 U(x) \preceq \lambda_{\max}I$ for all $x$, where $\lambda_{\max},\lambda_{\min}>0$. Let the stepsize be given by $\gamma_j \equiv \gamma = O\big (\frac{\lambda_{\min}}{\lambda_{\max}^2}\big )$  and the number of iterations $N$ satisfy $N = \Omega  \big ((\frac{\lambda_{\max}}{\lambda_{\min}} )^2\big ).$\footnote{$a_n = O(b_n)$ means $\limsup_{n \to \infty} |a_n/b_n |< \infty$, and $a_n = \Omega(b_n)$ indicates $\liminf_{n \to \infty}|a_n/b_n |>0$.}
Given $X_0 \in \argmin U(x)$, let $p_N$ denote the pdf of $X_N$ obtained by iterating~\eqref{eq:ula}. Then, 
$\mathbb{E}_{x \sim p, \tilde{x} \sim p_N}\big[
| x- \tilde{x} |^2
\big ]^{\frac{1}{2}} \leq O\big( \sqrt{\frac{1}{\lambda_{\min}}}\big)$,
where $x=x_{\gamma N}$ is a solution to~\eqref{eq:cont} with $X_0 \sim e^{-U(x)}$ and the joint probability distribution of $x\sim p$ and $\tilde x\sim p_N$ is obtained via the shared Brownian motion.
\end{thm}

\section{Online Learning Algorithm}

The naive TS approach for learning LQR has two main weaknesses. The first arises from the potential selection of a destabilizing controller, which can cause the system state to grow exponentially and lead to unbounded regret. To address this issue, we control the probability of the state exhibiting excessively large norms. The second weakness stems from inefficiencies in the sampling process when the system noise and prior distributions are not conjugate. In such cases, ULA offers an alternative for posterior approximation, but it is often extremely slow. To accelerate the sampling process, we introduce a preconditioning technique. 

\subsection{Preconditioned ULA for approximate posterior sampling}

One of the  key components of our learning algorithm is approximate posterior sampling via preconditioned Langevin dynamics. The potential in ULA is chosen as $U_t(\theta) := - \log p( \theta | h_{t})$, where $p (\theta | h_{t})$ denotes the posterior distribution of the true system parameter given the history up to $t$. Unfortunately, a direct implementation of ULA to TS for LQR is inefficient as it requires a large number of iterations. To accelerate the convergence of Langevin dynamics, we propose a preconditioning technique.\footnote{Preconditioning techniques have been used for Langevin algorithms in different contexts; see, e.g., \cite{li2016preconditioned, lu2020continuum, bras2022langevin}.}

To describe the preconditioned Langevin dynamics, we choose a positive definite matrix $P$, referred to as a \emph{preconditioner}. 
The change of variables $\theta' = P^{\frac{1}{2}}\theta$ yields $
\mathrm{d}\theta_\tau = -{P^{-1}  \nabla U_t (\theta_\tau)}\mathrm{d} \tau + \sqrt{2P^{-1}} \mathrm{d}  B_\tau$. 
Applying the Euler--Maruyama discretization with constant stepsize $\gamma$ yields the preconditioned ULA:
\begin{equation}\label{preULA}
\theta_{j+1} = \theta_j - \gamma P^{-1} \nabla U_t(\theta_j) + \sqrt{2\gamma P^{-1}}  W_j,
\end{equation}
where $(W_j)_{j\geq 1}$ is an i.i.d. sequence of standard $dn$-dimensional Gaussian random vectors.

Given the data $z_t = (x_t, u_t)$ collected, the preconditioner in our setting is defined as
\begin{equation}\label{eq:preconditioner}
P_t:=\lambda I_{dn}+\sum_{s=1}^{t-1} \mathrm{blkdiag}\{z_s z_s^\top\}_{i=1}^n,
\end{equation}
where $\mathrm{blkdiag}\{A_i\}_{i=1}^n\in\mathbb{R}^{dn\times dn}$ denotes the block diagonal matrix of the $A_i$, and $\lambda > 0$ is a constant determined by the prior. 
Then, the curvature of the Hessian of the potential is bounded when scaled along the spectrum of the preconditioner, which is shown in the following lemma:

\begin{lem}\label{concavity}
Suppose Assumption~\ref{ass:noise} holds and the potential of the prior satisfies $\nabla^2_\theta U_1(\cdot)= \lambda I_{dn}$ for some $\lambda>0$. Then, for all $\theta$ and $t$, we have
$m I_{dn} \preceq  P_t^{-\frac{1}{2}}\nabla^2 U_t(\theta)P_t^{-\frac{1}{2}}  \preceq M I_{dn}$,
where $m=\min\{ \underline m, 1\}$ and $M = \max\{\overline m, 1\}$.
\end{lem}
The proof of this lemma can be found in Appendix~\ref{app:sec_a}.  It follows from Lemma~\ref{concavity} and Theorem~\ref{thm:naive_ula} that we can rescale the number of
iterations required for the convergence of ULA while ensuring improved accuracy in the concentration of the sampled system parameter. In fact, we show later that the number of required iterations scales only with $n$. To demonstrate the effect of preconditioning, note that Lemma~\ref{concavity} implies 
$m \lambda_{\min} (P_t)I_{dn} \preceq \nabla^2 U_t \preceq M \lambda_{\max}(P_t)I_{dn}$.
Theorem~\ref{thm:naive_ula} then implies that $O\big ( (\lambda_{\max} (P_t)/\lambda_{\min} (P_t))^2 \big )$ iterations are needed to achieve an error bound of $O\big ({1}/{\sqrt{\lambda_{\min} (P_t)}} \big )$.
Our algorithm improves this bound to $O\big ({1}/{\sqrt{\max\{\lambda_{\min} (P_t),t\}}}\big )$.
Throughout the paper, we use the notation $\mathbf{U}_k := U_{t_k}$ to explicitly indicate the dependence on the current episode $k$.

\begin{rem}
Our preconditioner can be viewed as an adaptive scaling mechanism analogous to the Fisher information matrix in natural policy gradient methods. This connection arises because the empirical covariance matrix captures the local curvature of the posterior distribution, effectively conditioning the Langevin dynamics for more efficient sampling.
\end{rem}

\subsection{Algorithm}\label{sec:alg}

We begin by introducing the following log-concavity condition on the prior, centered arbitrarily. This condition is a slight relaxation of the assumption in~\cite{ouyang2019posterior}.


\begin{assumption}\label{ass:init_bound}
The prior $p_1$ satisfies $\nabla^2_\theta U_1(\cdot) = \lambda I_{dn}$ for $U_1(\cdot):=-\log p_1(\cdot)$ and some $\lambda\geq1$
\end{assumption}
The initialization of the preconditioner \( P_t \) plays a crucial role in the efficiency of the sampling process. If \( P_0 \) is too small, the algorithm may suffer from slow exploration due to small step sizes in the Langevin dynamics. Conversely, if \( P_0 \) is too large, the algorithm may place excessive trust in the prior, potentially slowing adaptation to the true system parameters. Our choice of \( P_0 = \lambda I \) with a moderate \(\lambda\) ensures a balance between these effects. For mathematical convenience, it suffices to set $\lambda>0$, but we assume $\lambda \geq 1$ to simplify the analysis.

Following~\cite{abeille2018improved}, we consider an admissible set of parameters defined as $\mathcal{C}:=\{\theta \in \mathbb{R}^{dn}:|\theta|\leq S,|A+BK(\theta)|\leq \rho<1, J(\theta)\leq M_J\}$ for some constants $S,\rho,M_J>0$ where ${\theta} = \mathrm{vec}(\begin{bmatrix} 
A & B
\end{bmatrix}^{\top})$. 
To sample from the posterior distribution, we restrict the sample to lie within $\mathcal{C}$ via rejection sampling. This ensures that for any sampled system parameter $ \theta \in \mathcal{C}$, there exists a positive constant $M_{P^*}$ such that $| P^*(\theta) | \leq M_{P^*}$~\cite{abeille2017thompson}. 
Consequently, $|[I \quad K(\theta)^\top ]| \leq M_K$ for some $M_K>1$, and therefore,
$|A_*+B_*K(\theta)|\leq M_\rho$ for some $M_\rho \geq 1$.

\begin{algorithm}[h]
\caption{Thompson sampling with Langevin dynamics for LQR
}
\label{alg:ula}
\begin{algorithmic}[1]
\State {\bf Input:} $p_1$;
\State {\bf Initialization:} $t \leftarrow 1$, $t_0 \leftarrow 0$,
$x_1 \leftarrow 0$, $\mathcal{D} \leftarrow \emptyset$,
$\mathbf{U}_0 \leftarrow U_1$,
$\tilde{{\theta}}_0 \leftarrow \argmin U_1 (\theta)$, $\theta_{\min, 0} \leftarrow \tilde{\theta}_0$;
\For{Episode $k = 1,2,\ldots$}
\State $T_{k} \leftarrow k+1$, and $t_k \leftarrow t$;
\State $\mathbf{U}_k (\cdot) := \mathbf{U}_{k-1} (\cdot) -\sum_{(z_t, x_{t+1}) \in \mathcal{D}} \log p_w(x_{t+1} -\Theta^\top z_t)$;
\State $\mathcal{D} \leftarrow \emptyset$;
\State $\theta_{\min, k} \in \argmin \mathbf{U}_k (\theta)$;
\State Compute the preconditioner $\tilde{P}_k$, the step  size $\tilde{\gamma}_k$, and the number of iterations $\tilde{N}_k$ as \eqref{preconditioner_stepsize_iteration};
\While{True}
\State ${\theta}_0 \leftarrow \theta_{\min, k}$;
\For{Step $j=0, 1, \ldots, \tilde{N}_k-1$}
\State Sample $\theta_{j+1} \sim \mathcal{N} \big( \theta_j - \tilde{\gamma}_k \tilde{P}_k^{-1} \nabla \mathbf{U}_k (\theta_j), 2\tilde{\gamma}_k \tilde{P}_k^{-1}\big)$;
\EndFor

\If{$\theta_{\tilde{N}_k} \in \mathcal{C}$}
\State $\tilde\theta_k\leftarrow\theta_{\tilde{N}_k}$
\State Break;
\EndIf

\EndWhile
\State Compute the gain matrix $K_k := K(\tilde{\theta}_k)$;
\While{ $t \leq t_k + T_{k}-1$}
\State Execute control $u_t = K_k x_t + \nu_t$ for $\nu_t$ satisfying Assumption~\ref{ass:perturb};
\State Observe new state $x_{t+1}$, and update $\mathcal{D} \leftarrow \mathcal{D} \cup \{(z_t, x_{t+1})\}$;
\State  $t \leftarrow t+1$;
\EndWhile
\EndFor
\end{algorithmic}
\end{algorithm}

Our proposed algorithm is presented in Algorithm~\ref{alg:ula}. We employ dynamic episode scheduling, as it has been shown to be effective in the literature~\cite{abbasi2011regret, ouyang2019posterior, abeille2017thompson}. In the algorithm, $t_k$ and $T_k$ denote the start time and the length of episode $k$, respectively. By definition, $t_1=1$ and $t_{k+1} = t_k + T_k$. The episode length is chosen as $T_k = k+1$.
To update the posterior—or equivalently, the potential—at episode $k$, we use the dataset $\mathcal{D}:= \{(z_t, x_{t+1})\}_{t_{k-1} \leq t \leq t_{k} - 1}$ collected during the previous episode. 
It follows from \eqref{posterior} that the potential can be updated as Line 5,
where $\mathbf{U}_0$ is initialized as $U_1$, the potential of the prior.
Approximate TS is then performed using the preconditioned ULA with the preconditioner, step size, and number of iterations chosen as $\tilde{P}_{k}:= {P}_{t_k}$, $\tilde{\gamma}_k:={\gamma}_{t_k}$ and $\tilde{N}_k:=\max(1, \lceil{N}_{t_k}\rceil)$, where
\begin{equation}\label{preconditioner_stepsize_iteration}
     \begin{split}
     &{P}_{t}:=\lambda I_{dn} + \sum_{s=1}^{t-1} \mathrm{blkdiag}\{z_sz_s^\top\}_{i=1}^n, \; {\gamma}_t := \frac{m \lambda_{\min,t}}{16 M^2 \max\{\lambda_{\min,t},t \}}, \;  {N}_t := \frac{4\log_2 \left(\frac{\max\{\lambda_{\min,t},t\}}{\lambda_{\min,t}}\right)}{m \gamma_t}.
     \end{split}
\end{equation}
Here, $\lambda_{\min,t}$ and $\lambda_{\max,t}$ denote the minimum and maximum eigenvalues of ${P}_t$. This choice is based on a detailed analysis of the concentration properties of ULA, as established in Proposition~\ref{prop:ula}. The additional operations on $N_{t_k}$ ensure $\tilde N_k\in\mathbb{N}$, avoiding the possibility of infinite rejection when $\tilde N_k=0$. In the algorithm, we obtain the unique minimizer $\theta_{\min, t}$ using Newton's method.

After performing the preconditioned ULA update $\tilde{N}_k$ times, we check whether $\theta_{\tilde{N}_k} \in \mathcal{C}$. If so, the sampled parameter is accepted and the corresponding control gain matrix is computed via ARE~\eqref{ric}. 
To ensure that the rejection step ends in a finite number of iterations,
we assume that there exists a small positive constant $\epsilon$ such that, for each episode $k$,  
$\mathrm{Pr}(\tilde{\theta}_k \in \mathcal{C}) \geq 1 - \epsilon$ under the posterior distribution.
Although this assumption may appear restrictive, it has been empirically validated in all of our examples, as shown in Appendix~\ref{app:additional}.

\begin{figure}[h]\label{alg:infusing}
    \centering
    \begin{tikzpicture}[scale = 1.2]
    \draw[-latex] (1,0) -- (7,0) ;
    \foreach \x in  {1,2,3,4,5,6} \draw[shift={(\x,0)},color=black] (0pt,3pt) -- (0pt,-3pt);
    \foreach \x\xtext in {1/$t_1=1$,2,3/$t_2=3$,4,5,6/$t_3=6$} \draw[shift={(\x,0)},color=black] (0pt,0pt) -- (0pt,-3pt)
    node[below] {\xtext};
    \draw ; \node[] at (2,0.4) {$T_1$};
    \draw ; \node[] at (4.5,0.4) {$T_2$};
    
    
    \draw[->] (2, -20 pt)node[below]{$\nu_1$} -- (2,-15pt);
    \draw[->] (5, -20 pt)node[below]{$\nu_2$} -- (5,-15pt);

    
    \draw [blue, thick, -stealth] (1,0) |- (2.9,0.2);
    \draw [blue, thick, -stealth] (3,0) |- (5.9,0.2);
    
    \end{tikzpicture}
    \caption{Infusing noise for enhanced exploration}
    \label{fig:horizon}
\end{figure}

A novel component of our algorithm is the injection of a noise signal into the control input $u_t$ at the end of each episode as illustrated in Figure~\ref{fig:horizon}. 
This perturbation enhances exploration.
The external noise signal is assumed to satisfy the following:

\begin{assumption}\label{ass:perturb}
The random variable $\nu_s\in\mathbb{R}^{n_u}$ is $\bar L_\nu$-sub-Gaussian,\footnote{A distribution is $L_\nu$-sub-Gaussian if $\mathrm{Pr}(|\nu| > y )< C  \text{exp}(-\frac{1}{2L_{\nu}^2} y^2)$ for some $C>0$.}     and satisfies $\nu_s=0$ if $s\in [t_j,t_{j+1}-2]$ for $j\geq 2$. Moreover, $\mathbb{E}[\nu_s]=0$ and $\mathbf{W}':=\mathbb{E}[\nu_s \nu_s^{\top}]$ is a positive definite matrix whose maximum and minimum eigenvalues are identical to those of $\mathbf{W}$.\footnote{The assumption on the maximum and minimum eigenvalues of $\mathbf{W'}$ is made for simplicity in the proof of Proposition~\ref{prop:pers-ex} which concerns the growth of $\lambda_{\min}(P_t)$.} 
\end{assumption}
Since our algorithm does not rely on a predefined stabilizing set of parameters, one may be concerned that the control policies generated during the early learning phase could exhibit instability due to limited data. To address this issue, our excitation mechanism ensures that the preconditioner matrix grows over time, thereby improving the concentration properties of the sampled system parameters, as shown in the following section.

\section{Concentration Properties}

To show that Algorithm~\ref{alg:ula} achieves an $O(\sqrt{T})$ regret bound, 
we first examine the concentration properties of the exact and approximate posterior distributions given the history up to a fixed time $t$ for the potential $U_t(\theta)=U_1(\theta) - \sum_{s=1}^{t-1} \log p_w(x_{s+1}-\Theta^\top z_s)$. When $t$ is chosen as $t_k$, we recover the case corresponding to Algorithm~\ref{alg:ula}.
As illustrated in Figure~\ref{fig:flow}, 
the concentration results established in this section enable us to bound the moments of the system state, which is essential for attaining the desired regret bound in Section~\ref{sec:regret}.

\begin{figure*}[h]
\centering
\tikzstyle{concentration} = [rectangle, rounded corners, 
minimum width=1.5cm, 
minimum height=1.5cm,
text centered, 
draw=black, 
fill=white!30]

\tikzstyle{excitation} = [rectangle, rounded corners, 
minimum width=1.5cm, 
minimum height=1.5cm,
text centered, 
draw=black, 
fill=white!30]

\tikzstyle{state} = [rectangle, rounded corners, 
minimum width=1.5cm, 
minimum height=1.5cm,
text centered, 
draw=black, 
fill=white!30]

\tikzstyle{etc} = [rectangle, rounded corners, 
minimum width=1.5cm, 
minimum height=1.5cm,
text centered, 
draw=black, 
fill=white!30]

\tikzstyle{main} = [rectangle, rounded corners, 
minimum width=1.5cm, 
minimum height=1.5cm,
text centered, 
draw=black, 
fill=white!30]

\tikzstyle{io} = [trapezium, 
trapezium stretches=true, 
trapezium left angle=70, 
trapezium right angle=110, 
minimum width=3cm, 
minimum height=1cm, text centered, 
draw=black, fill=blue!30]

\tikzstyle{process} = [rectangle, 
minimum width=3cm, 
minimum height=1cm, 
text centered, 
text width=3cm, 
draw=black, 
fill=orange!30]

\tikzstyle{decision} = [diamond, 
minimum width=3cm, 
minimum height=1cm, 
text centered, 
draw=black, 
fill=green!30]
\tikzstyle{arrow} = [thick,->,>=stealth]

\begin{tikzpicture}[node distance=2.5cm,font=\sffamily\scriptsize]

\node (state_poly) [state] 
{\begin{tabular}{c}
    Theorem ~\ref{thm:state_poly}: \\
    Polynomial time
    bound\\ for  system state
\end{tabular}};

\node (concent_prob) [concentration, above of = state_poly, yshift=-0.5cm] {\begin{tabular}{c}
    Prop. ~\ref{prop:ula} \& Prop. ~\ref{prop:half_exact_true}:\\
    Comparison between\\ exact and approximate\\ posteriors
\end{tabular}};

\node (concent) [concentration, right of=state_poly, xshift= 1.25cm]
{\begin{tabular}{c}
    Theorem ~\ref{thm:exact_appox}: \\
    Concentration of\\  approximate posteriors
    \end{tabular}};

\node (per_ex) [excitation, above of=concent, yshift=-0.5cm]
{\begin{tabular}{c}
    Prop. ~\ref{prop:pers-ex}:\\
    Decay of $\frac{1}{\lambda_{\min,t}}$\\
    as $t\rightarrow\infty$
\end{tabular}};

\node (state_bound) [state, right of=concent,xshift= 1cm]
{\begin{tabular}{c}
    Theorem ~\ref{thm:improved}: \\  
    Bounding\\ the moments of\\  system state
\end{tabular}};
\node (regret) [main, right of=state_bound,xshift= 1cm] 
{\begin{tabular}{c}
    Theorem ~\ref{thm:regret}:\\ Regret bound\\  
    $R(T) \leq O(\sqrt{T})$
\end{tabular}};
\node (bellman) [etc, above of=regret,yshift= -0.5cm] 
{\begin{tabular}{c}
    Bellman's principle  
\end{tabular}};

\draw [arrow] (concent_prob) -- (state_poly);
\draw [arrow] (state_poly) -- (concent);
\draw [arrow] (state_poly.south) to [out=-10,in=-170] (state_bound.south);
\draw [arrow] (per_ex) -- (concent);
\draw [arrow] (concent) -- (state_bound);
\draw [arrow] (state_bound) -- (regret);
\draw [arrow] (bellman) -- (regret);
\end{tikzpicture}
\caption{Flow chart of our theoretical results.}\label{fig:flow}
\end{figure*}

\subsection{Comparing exact and approximate posteriors}

Let $\mu_t$ denote the \emph{exact posterior} distribution defined by $\mu_t \propto \exp (-U_t)$.\footnote{Throughout this subsection, in the definition of the potential $U_t$, we let $(z_s)_{s\geq1}$ be an $\mathbb{R}^d$-valued stochastic process adapted to a filtration $(\mathcal{F}_t)_{t\geq0}$, where each $z_s$ is assumed to be $\mathcal{F}_{s-1}$-measurable for all $s\geq1$.}
For the approximate posterior, recall the preconditioned ULA that generates
$\theta_{j+1} \sim \mathcal{N} \left( \theta_j - \gamma_t P_t^{-1} \nabla U_t (\theta_j), 2\gamma_t P_t^{-1}\right)$ starting from $\theta_0 \in  \argmin U_t (\cdot)$.
After repeating this update for $N_t$ steps, we obtain $\theta_{N_t}$.
We let $\tilde{\mu}_t$ denote the \emph{approximate posterior}, defined as the distribution of $\theta_{N_t}$. 
We first compare the exact and approximate posteriors. 
The result quantifies the concentration depending on the moment $p$. The higher moment bound for $p>2$ is used to characterize a set of system parameters with which the state does not grow exponentially as illustrated in the following subsection, while the bound for $p=2$ is necessary for our regret analysis.
Throughout the paper, the joint distribution between $\theta_t\sim\mu_t$ and $\tilde\theta_t\sim\tilde \mu_t$ is characterized via a shared Brownian path driving both the continuous Langevin diffusion and the discrete ULA dynamics with the preconditioner, as demonstrated in Remark~\ref{rem:shared}.
\begin{prop}\label{prop:ula}
Suppose Assumptions~\ref{ass:noise} and~\ref{ass:init_bound} hold. 
Then, the exact posterior $\mu_t$ and the approximate posterior $\tilde{\mu}_t$ obtained via preconditioned ULA satisfy 
\[
\mathbb{E}_{\theta_t \sim \mu_t,\, \tilde{\theta}_t \sim \tilde{\mu}_t}\big[
| \theta_t - \tilde{\theta}_t |^p_{P_t} \mid h_{t}
\big ] \leq D_p
\]
for all $p\geq 2$,
where $D_p = \big(\frac{pdn}{m}\big)^{\frac{p}{2}}\big(2^{2p+1}+5^p\big)$. When $p=2$, we further have
\begin{equation}\label{bd:comp}
\mathbb{E}_{\theta_t \sim \mu_t,\, \tilde{\theta}_t \sim \tilde{\mu}_t}\big[
| \theta_t - \tilde{\theta}_t |^2 \mid h_{t}
\big ]^{\frac{1}{2}} \leq \sqrt{\frac{D}{\max\{\lambda_{\min,t}, t\}}},
\end{equation}
where $D = 114\frac{dn}{m}$ and $\lambda_{\min,t}$ denotes the minimum eigenvalue of $P_t$.
\end{prop}

The proof of this proposition is contained in Appendix~\ref{app:prop:ula}.
Without the preconditioner, it would have been inevitable to obtain a result weaker than Proposition~\ref{prop:ula}; Theorem~\ref{thm:naive_ula} would yield a convergence rate of $O({1}/{\sqrt{\lambda_{\min,t}}})$, which is an LQR version of \cite[Theorem 5]{mazumdar2020thompson}.
We infused the time step $t$ into the step size required for ULA so that the right-hand side of \eqref{bd:comp} decreases with $t$. 
Thus, $\max\{\lambda_{\min,t}, t\} \geq \lambda_{\min,t}$ contributes to an improved concentration property.

Another important observation is a concentration bound for the exact posterior. 
This concentration property is essential for characterizing a confidence set used in the proof of Theorem~\ref{thm:state_poly}.
\begin{prop}\label{prop:half_exact_true}
Suppose Assumptions~\ref{ass:noise} and~\ref{ass:init_bound} hold. 
Then, the following inequality
\begin{equation}
\begin{split}
    &\mathbb{E}_{\theta_t \sim \mu_t}\big[
    | \theta_t - \theta_* |^p_{P_t} \mid h_{t}
    \big ]^{\frac{1}{p}} \leq 2p \sqrt{\frac{8nM^2}{m^3}\log \bigg(\frac{n}{\delta}\bigg(\frac{\lambda_{\max,t}}{\lambda}\bigg)^{\frac{d}{2}}\bigg)+C}, \quad t >0
\end{split}
\end{equation}
holds with probability at least $1-\delta$ for
any $0<\delta < 1$ and $p \geq 2$, where
the constant $C>0$ depends only on $p$, $m$, $n$, $d$, and $\lambda$, and $\lambda_{\max,t}$ denotes the maximum eigenvalue of $P_t$.\footnote{Here, the probability $1 - \delta$ is with respect to the randomness of the trajectory $(z_s)_{s \geq 1}$.}
\end{prop}

The proof of this proposition can be found in Appendix~\ref{app:prop:half_exact_true}.

\subsection{Bounding expected state norms by a polynomial of time}\label{sec:polystate}

A key result we derive from Propositions~\ref{prop:ula} and~\ref{prop:half_exact_true} is that the system state grows at most polynomially in expectation over time. 
To show this property, we modify the confidence set construction and self-normalization technique developed for the OFU approach~\cite{abbasi2011improved,abbasi2011regret}.
Our key idea is to construct a set that contains the system parameters sampled via ULA with high probability.
The higher-moment bounds from Propositions~\ref{prop:ula} and ~\ref{prop:half_exact_true} are crucial to our analysis as Markov-type inequalities can be exploited for any $p$. We then partition the probability space of the stochastic process into two sets, ``good'' and ``bad,'' as in the OFU approach.

\begin{thm}\label{thm:state_poly}
Suppose Assumptions \ref{ass:noise},\ref{ass:init_bound} and \ref{ass:perturb} hold.
For $T >0$, $p\geq 2$, and a random trajectory $(x_s)_{s=1}^T$ generated by Algorithm~\ref{alg:ula}, we have
\[
\mathbb{E} \Big [\max_{j\leq t}|x_j|^p  \Big ] \leq Ct^{\frac{7}{2}p(d+1)}, \quad t \geq 1,
\]
where the constant $C>0$ depends only on $p$, $m$, $n$, $n_u$, $\bold{W}$, $M_\rho$ and $\lambda$.
\end{thm}

The proof of this theorem can be found in Appendix~\ref{app:polystate}.
It is worth emphasizing that this polynomial-time bound is attained without using predefined sets of parameters that make the true system stabilizable. 
In Section~\ref{sec:regret}, we will further improve the result to a uniform bound, which plays a critical role in our regret analysis.

\subsection{Concentration of exact and approximate posteriors}

Leveraging the previous results on the concentration and the expected state norms, we can deduce that the minimum eigenvalue of the preconditioner actually grows in time. 
Exploiting this property and Theorem~\ref{thm:state_poly}, an improved concentration property of the exact posterior follows. Finally, the triangle inequality yields the desired result, the concentration of the approximate posterior around the true system parameter.

We begin by characterizing the growth of the minimum eigenvalue of the preconditioner which results from injecting a random noise signal $\nu_s$ to perturb the action at the end of each episode.
To derive this result, we decompose the preconditioner in each episode into two parts—a random matrix and a self-normalized matrix-valued process—as in~\cite{jedra2022minimal}. 
Specifically, by Lemma~\ref{lem:preconditioner_expansion}, 
\begin{equation}\nonumber
\begin{split}
&\sum z_s z_s^\top=\sum  \underbrace{(L_s\psi_s)(L_s\psi_s)^{\top}}_{\text{random matrix part}} - \underbrace{\left(\sum  y_s (L_s \psi_s)^{\top}\right)^{\top}\left(\sum y_sy_s^{\top}+  I_d\right)^{-1}\left(\sum   y_s (L_s \psi_s)^{\top}\right)}_{\text{self-normalization}} -  I_d,
\end{split}
\end{equation}
where $y_s := 
\begin{bmatrix}
    A_*x_{s-1} + B_*u_{s-1}          \\
    K_j(A_*x_{s-1} + B_*u_{s-1})
\end{bmatrix}$, 
$L_s := 
\begin{bmatrix}
    I_n & 0\\
    K_j & I_{n_u}
\end{bmatrix}$,
 $\psi_s :=
\begin{bmatrix}
    w_{s-1}\\
    \nu_{s}
\end{bmatrix}$, and $K_j$ is  the control gain matrix used in the $j$th episode.
The random matrix part contributes the growth of the minimum eigenvalue of the preconditioner with high probability.
More precisely, the following proposition holds:

\begin{prop}\label{prop:pers-ex}
Suppose that Assumptions \ref{ass:noise}--\ref{ass:perturb} hold. For  $k\geq k_0(m,n,n_u,\lambda,M_K,M_\rho,\mathbf{W})$, we have
\[
\mathbb{E}\bigg[\frac{1}{\lambda_{\min,t_{k+1}}^p}\bigg] \leq C k^{-p}, \quad p\geq 2,\] where $t_{k+1}$ is the start time of episode $k+1$ in Algorithm~\ref{alg:ula}, 
$\lambda_{\min, t_{k+1}}$ denotes the minimum eigenvalue of $\tilde{P}_{k+1} := P_{t_{k+1}}$, and the constant $C > 0$ depends only on $p, n, n_u, \bold{W}, M_K$ and $\lambda$.
\end{prop}

The proof of this proposition can be found in Appendix~\ref{app:prop:pers-ex}.
Recalling the probabilistic bound for $|\theta_t - \theta_*|_{P_t}$ from Proposition~\ref{prop:half_exact_true}, we observe that
$|\theta_t - \theta_* |$ is controlled by  $1/\sqrt{\lambda_{\min,t}}$ and the self-normalization term.
Using Theorem~\ref{thm:state_poly}, we can show that the latter is dominated by the former, which grows at most polynomially in time due to Proposition~\ref{prop:pers-ex}. 
Consequently, the following improved concentration bound holds for the exact posterior.

\begin{thm}\label{thm:exact_appox}
Suppose Assumptions \ref{ass:noise}--\ref{ass:perturb} hold. Then, the exact posterior $\mu_{t}$ and the approximate posterior $\tilde \mu_t$ realized from the shared Brownian motion satisfy
\[
\mathbb{E}\big [\mathbb{E}_{\theta_t \sim \mu_{t}}[|\theta_t-\theta_*|^p \mid h_{t}] \big ] \leq C\Big(t^{-\frac{1}{4}}\sqrt{\log t}\Big)^p, \mbox{ and }
\mathbb{E}\big[\mathbb{E}_{\tilde{\theta}_t \sim \tilde{\mu}_t}[| \tilde{\theta}_t - \theta_* |^p \mid h_{t} ]\big] \leq C\Big(t^{-\frac{1}{4}}\sqrt{\log t}\Big)^p
\] 
for all $t\geq 1$ and $p\geq 2$, where the outer expectation is taken over all histories, and the constant $C > 0$ depends only on $p, n, n_u, \bold{W}$, $M_K$, $M_\rho$, and $\lambda$.
\end{thm}

The proof of this theorem can be found in Appendix~\ref{app:thm:exact_appox}.

\section{Regret Bound} \label{sec:regret}

To further improve the bound in Theorem~\ref{thm:state_poly}, we decompose the moment of the system state into two parts based on the following cases: $|\tilde \theta_t-\theta_*|\leq \epsilon_0$ and $|\tilde \theta_t-\theta_*| > \epsilon_0$, where $\epsilon_0$ is a positive constant. 
When $\epsilon_0$ is sufficiently small, 
we have $|A_*+B_*K(\tilde \theta_t)|<1$, and thus the first part can be easily handled. 
For the second part, we invoke the Markov inequality to balance the growth of the state with the tail probability by choosing an appropriate value of $p$. 
This intuitive argument can be made rigorous using Theorems~\ref{thm:state_poly} and~\ref{thm:exact_appox}, leading to the following result.

\begin{thm}\label{thm:improved}
Suppose that Assumptions \ref{ass:noise}-\ref{ass:perturb} hold.
For any $T > 0$ and a random trajectory $(x_s)_{s=1}^T$ generated by Algorithm~\ref{alg:ula}, we have
\[
\mathbb{E} [|x_t|^q ] < C, \quad q = 2, 4,
\]
where  the constant $C >0$ depends only on $p, n, n_u, \bold{W}, M_K, M_\rho, \epsilon_0$, and $\lambda$. {Here, $\epsilon_0$ is a positive constant such that $|\theta - \theta_*| \leq \epsilon_0$ implies $|A_* + B_* K(\theta)| < 1$.} 
\end{thm}
The proof of this theorem can be found in Appendix~\ref{app:thm:improved}.


Finally, we establish our main result: Algorithm~\ref{alg:ula} achieves an $O(\sqrt{T})$ Bayesian regret bound.

\begin{thm}\label{thm:regret}
Suppose that Assumptions \ref{ass:noise}-\ref{ass:perturb} hold. Then, the Bayesian regret~\eqref{regret} of Algorithm~\ref{alg:ula} is bounded as follows:
\[
R(T) \leq O(\sqrt{T}).
\]
\end{thm}
The proof of this theorem can be found in Appendix~\ref{app:thm:regret}.
The regret bound is empirically verified by the results of our experiments. See Appendix~\ref{app:exp} for our empirical analyses.

\section{Concluding Remarks}
We proposed a novel approximate Thompson sampling algorithm for learning LQR with an improved $O(\sqrt{T})$ regret bound.  
Our method does not require the noise to be Gaussian or the columns of $\Theta$ to be independent.  
This relaxation of restrictive assumptions is enabled by a carefully designed preconditioned ULA and the use of perturbed control actions only at the end of each episode.

As a future research direction, it may be possible to extend our algorithm to settings with noise distributions having non-log-concave potentials.  
In our work, the log-concavity of the posterior potential is preserved under the considered noise models, which enables acceleration of the sampling process through preconditioning.  
To handle more general classes of noise, alternative techniques beyond the current ULA framework may be necessary.  
Recently,~\cite{cheng2018sharp} derived sharp non-asymptotic convergence rates for Langevin dynamics in nonconvex settings.  
We plan to investigate the incorporation of such results into our framework.
\numberwithin{equation}{section}
 
\appendix

\section{Proofs}
\subsection{Proof of Theorem~\ref{thm:naive_ula}}\label{app:naive_ula}

To prove Theorem~\ref{thm:naive_ula}, we use the following lemma. 

\begin{lem}\label{lem:naive_ula}
Suppose Assumption~\ref{ass:noise} holds.
Let $X \in \mathbb{R}^{n_x}$ be a random variable with probability density function $p(x) \propto e^{-U(x)}$, where $\lambda_{\min} I_{n_x}\preceq \nabla^2 U  \preceq  \lambda_{\max}I_{n_x}$
for $\lambda_{\max},\lambda_{\min}>0$. Let
$\{ Y_j\}$, $Y_j  \in \mathbb{R}^{n_x}$, be generated by the ULA as
\[
Y_{j+1} = Y_j - \gamma \nabla U(Y_j) + \sqrt{2\gamma} W_j,
\]
where $Y_0$ is a random variable with an arbitrary density function.
If  $\gamma \leq \frac{ \lambda_{\min}}{16\lambda_{\max}^2}$, then we have
\[
\mathbb{E} [ | Y_j - X |^2 ] < 2^{-\frac{\lambda_{\min}\gamma    j}{4}} \mathbb{E} [
| Y_0 - X |^2 
] + 
2^8 \frac{n_x \lambda_{\max}^2 }{\lambda_{\min}^2} \gamma,
\]
where $X$ and $Y_j$ are understood via the shared Brownian motion in continuous and discretized stochastic differential equations as demonstrated in Remark~\ref{rem:shared}.
\end{lem}

\begin{proof}
Let $\{Z_{\tau}\}_{\tau \geq 0}$ be a continuous interpolation of $\{Y_j\}$, defined by
\begin{equation}\label{eq:naive_sde_z}
\left \{
\begin{array}{ll}
\mathrm{d} Z_{\tau} = -  \nabla U(Y_j) \mathrm{d}{\tau} + \sqrt{2} \mathrm{d} B_{\tau}& \mbox{for }\tau \in [j\gamma, (j+1) \gamma) \\
Z_{\tau} = Y_j & \mbox{for } \tau = j\gamma.
\end{array}
\right.
\end{equation}
Note that $\lim_{\tau \nearrow j\gamma} Z_{\tau} = Y_j =  \lim_{\tau \searrow j\gamma} Z_{\tau}$ for each $j$, and thus $\{Z_{\tau}\}$ is a continuous process. 
We introduce another stochastic process $\{X_{\tau}\}$, defined by 
\[
\mathrm{d} X_{\tau} = - \nabla U( X_{\tau}) \mathrm{d} \tau + \sqrt{2} \mathrm{d} B_{\tau},
\]
where $X_0$ is a random variable with pdf $p(x) \propto e^{-U(x)}$.
By Lemma~\ref{lem:ld}, $X_{\tau}$ has the same pdf $p(x)$ for all $\tau$. 
We use the same Brownian motion $B_{\tau}$ to define both $\{Z_{\tau}\}$ and $\{X_{\tau}\}$. 
Fix an arbitrary $j$. 
Differentiating $|Z_{\tau} - X_{\tau}|^2$ with respect to $\tau \in [j\gamma, (j+1)\gamma)$ yields
\begin{equation}\nonumber
\begin{split}
\frac{\mathrm{d} |Z_{\tau} - X_{\tau}|^2}{\mathrm{d}{\tau}} &= 2(Z_{\tau} - X_{\tau})^\top \bigg (
\frac{\mathrm{d} Z_{\tau}}{\mathrm{d} {\tau}} - \frac{\mathrm{d} X_{\tau}}{\mathrm{d} {\tau}}
\bigg )\\
&= 2 (Z_{\tau} - X_{\tau})^\top ( - \nabla U (Y_j) +  \nabla U(Z_{\tau}))
+ 2 (Z_{\tau} - X_{\tau})^\top (  - \nabla U(Z_{\tau}) +  \nabla U(X_{\tau})).
\end{split}
\end{equation}
Therefore, we have
\begin{equation}\nonumber
\begin{split}
&2 (Z_{\tau} - X_{\tau})^\top ( - \nabla U (Y_j) + \nabla U(Z_{\tau}))
+ 2 (Z_{\tau} - X_{\tau})^\top (  -  \nabla U(Z_{\tau}) +  \nabla U(X_{\tau}))\\
&\leq 2 (Z_{\tau} - X_{\tau})^\top( - \nabla U (Y_j) +  \nabla U(Z_{\tau}))
- 2\lambda_{\min} (Z_{\tau} - X_{\tau})^\top  (Z_{\tau} - X_{\tau})\\
&\leq  2 |Z_{\tau} - X_{\tau}| |\nabla U(Z_{\tau}) - \nabla U (Y_j) |
-2 \lambda_{\min} |Z_{\tau} - X_{\tau}|^2,
\end{split}
\end{equation}
where the first inequality follows from the strong convexity of $U$.
On the other hand, using Young's inequality, we have 
\[
|Z_{\tau} - X_{\tau}|  |  \nabla U(Z_{\tau}) - \nabla U (Y_j) | 
\leq \frac{\lambda_{\min}| Z_{\tau} - X_{\tau}|^2}{2}  + \frac{|   \nabla U(Z_{\tau}) - \nabla U (Y_j) |^2}{2  \lambda_{\min}}.
\]
Combining all together, we deduce that
\[
\frac{\mathrm{d} |Z_{\tau} - X_{\tau}|^2}{\mathrm{d}{\tau}} \leq - \lambda_{\min}| Z_{\tau} - X_{\tau}|^2 + \frac{1}{\lambda_{\min}} |   \nabla U(Z_{\tau}) - \nabla U (Y_j) |^2,
\]
which implies 
\[
\frac{\mathrm{d}}{\mathrm{d} {\tau}} (e^{\lambda_{\min}  {\tau}} | Z_{\tau} - X_{\tau} |^2) \leq 
\frac{e^{\lambda_{\min} {\tau}}}{\lambda_{\min}} |   \nabla U(Z_{\tau}) - \nabla U (Y_j) |^2.
\]
Integrating both sides from $j\gamma$ to $(j+1)\gamma$ and then multiplying $e^{-\lambda_{\min} (j+1) \gamma}$, we have
\begin{align*}
| Z_{(j+1)\gamma} - X_{(j+1)\gamma} |^2 &\leq 
e^{-\lambda_{\min}  \gamma} | Z_{j\gamma} - X_{j\gamma} |^2\\
&\quad+\frac{1}{\lambda_{\min}} \int_{j\gamma}^{(j+1) \gamma} 
e^{-\lambda_{\min}( (j+1)\gamma - s)} |   \nabla U(Z_s) - \nabla U (Y_j) |^2 \mathrm{d} s.
\end{align*}
Since $X_t$ and $X$ have the same pdf, 
we have
\begin{align}
\mathbb{E} [| Z_{(j+1)\gamma} - X |^2 ] &\leq 
e^{-\lambda_{\min} \gamma} \mathbb{E} [ | Z_{j\gamma} - X |^2]
+\frac{1}{\lambda_{\min}} \int_{j\gamma}^{(j+1) \gamma} 
 \mathbb{E} [ | \nabla U(Z_s) - \nabla U (Y_j) |^2 ]\mathrm{d} s \nonumber\\
 &\leq e^{-\lambda_{\min}  \gamma} \mathbb{E} [ | Z_{j\gamma} - X |^2]
+\frac{\lambda_{\max}^2}{\lambda_{\min} } \int_{j\gamma}^{(j+1) \gamma} 
\mathbb{E} [ | Z_s-Y_j|^2 ] \mathrm{d} s,\label{eq:naive_bd}
\end{align}
where the first inequality follows from $e^{-\lambda_{\min} ( (j+1)\gamma - s)}  \leq 1$ and the second inequality follows from the Lipschitz smoothness of $U$.

To bound~\eqref{eq:naive_bd}, we handle its first and second terms   separately. 
Regarding the second term, we first integrate the SDE \eqref{eq:naive_sde_z} from $j\gamma$ to $s \in [j \gamma, (j+1)\gamma )$  to obtain 
\begin{equation}\nonumber
 Z_s - Y_j   =    -(s - j\gamma)  \nabla U(Y_j) + \sqrt{2} (B_s - B_{j\gamma}).  
\end{equation}
The second term of \eqref{eq:naive_bd} can then be bounded  by
\begin{equation}\label{eq:naive_bd0}
\begin{split}
&\int_{j\gamma}^{(j+1) \gamma} 
\mathbb{E} [ | Z_s-Y_j|^2 ] \mathrm{d} s 
= \int_{j\gamma}^{(j+1) \gamma}\mathbb{E} [ | 
-(s - j\gamma)  \nabla U(Y_j) + \sqrt{2} (B_s - B_{j\gamma})
|^2 ] \mathrm{d} s \\
&\leq 2\bigg [
 \int_{j\gamma}^{(j+1) \gamma} \mathbb{E} [ | 
(s - j\gamma)  \nabla U(Y_j) |^2] \mathrm{d} s
+ 2 \int_{j\gamma}^{(j+1) \gamma} \mathbb{E} [|B_s - B_{j\gamma} |^2] \mathrm{d}s \bigg ].
\end{split}
\end{equation}
For $s \in [j\gamma, (j+1) \gamma)$, we note that $|s - j\gamma| \leq \gamma$, and thus
\begin{equation}\label{eq:naive_bd1}
\begin{split}
\int_{j\gamma}^{(j+1) \gamma} \mathbb{E} [ | 
(s - j\gamma)  \nabla U(Y_j) |^2] \mathrm{d} s
&\leq \gamma^2 \int_{j\gamma}^{(j+1) \gamma}  \mathbb{E} [
| \nabla U(Y_j) |^2 
] \mathrm{d} s \\
&= \gamma^3 \mathbb{E} [
| \nabla U(Y_j) |^2
] \\
&= \gamma^3 \mathbb{E} [
| \nabla U(Y_j) -  \nabla U(x_{\min}) |^2
]  \\
&\leq \gamma^3 \lambda_{\max}^2  \mathbb{E} [ | Y_j - x_{\min} |^2 ], 
\end{split}
\end{equation}
where $x_{\min}$ is a minimizer of  $U$.
It follows from  \cite[Lemma 9]{mazumdar2020thompson} that
\begin{equation}\label{eq:naive_bd2}
\begin{split}
\mathbb{E} [ | Y_j - x_{\min}|^2] & \leq 2\mathbb{E} [ | Y_j - X|^2] + 10^2 \frac{n_x}{\lambda_{\min}}.
\end{split}
\end{equation}
Moreover,  \cite[Lemma 8]{mazumdar2020thompson} yields
\begin{equation}\label{eq:naive_bd3}
\int_{j\gamma}^{(j+1)\gamma} \mathbb{E} [
| B_s - B_{j\gamma}|^2
]\mathrm{d}s \leq \frac{4n_x}{e} \gamma^2.
\end{equation}

Combining \eqref{eq:naive_bd0}--\eqref{eq:naive_bd3}, we obtain that 
\begin{equation}\nonumber
\begin{split}
\int_{j\gamma}^{(j+1)\gamma} \mathbb{E} [
| Z_s - Y_j |^2
]\mathrm{d} s &\leq 
2^2 \lambda_{\max}^2 \gamma^3 \mathbb{E} [ | Y_j - X |^2 ] + 2(10 \lambda_{\max})^2 \gamma^3 \frac{n_x}{\lambda_{\min}} +  \frac{16n_x}{e}\gamma^2\\
&< 2^2 \lambda_{\max}^2 \gamma^3 \mathbb{E} [ | Y_j - X |^2 ] + 2^5 n_x \gamma^2,
\end{split}
\end{equation}
where the second inequality follows from $\gamma \leq \frac{\lambda_{\min}}{16 \lambda_{\max}^2}$.

Substituting this bound into \eqref{eq:naive_bd}, we have
\begin{equation}\nonumber
\begin{split}
\mathbb{E} [| Z_{(j+1)\gamma} - X |^2 ]  &< e^{-\lambda_{\min}  \gamma} \mathbb{E} [ | Z_{j\gamma} - X |^2]
+2^2\frac{\lambda_{\max}^4}{\lambda_{\min}} \gamma^3 \mathbb{E} [ | Y_j - X |^2 ]
+ 2^5 n_x \frac{\lambda_{\max}^2}{\lambda_{\min}} \gamma^2\\
&\leq  \bigg (
1 - \frac{\lambda_{\min} }{4}\gamma
\bigg )^2 \mathbb{E} [ | Y_j - X|^2]+2^2\frac{\lambda_{\max}^4}{\lambda_{\min}} \gamma^3 \mathbb{E} [ | Y_j - X |^2 ]
+ 2^5 n_x \frac{\lambda_{\max}^2}{\lambda_{\min} } \gamma^2,
\end{split}
\end{equation}
where the second inequality follows from the fact that 
$e^{-x} \leq 1 - \frac{x}{2}$ for $x \in [0,1]$. 
To further simplify the upper-bound, we use the following  two inequalities: 
$2^2\frac{\lambda_{\max}^4}{\lambda_{\min}} \gamma^3 = \frac{\lambda_{\min}}{64 } \big (\frac{16 \lambda_{\max}^2}{\lambda_{\min}}\big )^2\gamma^3 
\leq \frac{\lambda_{\min}}{64 }  \gamma$ and 
$\big (
1 - \frac{\lambda_{\min}}{4}\gamma
\big )^2 + \frac{\lambda_{\min}}{64}  \gamma \leq \big (
1 - \frac{\lambda_{\min}}{8}\gamma
\big )^2$.
Consequently, $\mathbb{E} [| Z_{(j+1)\gamma} - X |^2 ]$ is bounded as
\[
\mathbb{E} [| Z_{(j+1)\gamma} - X |^2 ] < \bigg (
1 - \frac{\lambda_{\min} }{8}\gamma
\bigg )^2\mathbb{E} [ | Y_j - X|^2]  + 2^5 n_x \frac{\lambda_{\max} ^2}{\lambda_{\min} } \gamma^2.
\]
Invoking this inequality repeatedly yields
\begin{equation}\nonumber
\begin{split}
\mathbb{E} [| Z_{(j+1)\gamma} - X |^2 ] &<
\bigg (
1 - \frac{\lambda_{\min} }{8}\gamma
\bigg )^{2(j+1)} \mathbb{E} [ | Y_0 - X|^2]
+ \sum_{i=0}^j \bigg (
1 - \frac{\lambda_{\min} }{8}\gamma
\bigg )^{2i} 2^5 n_x \frac{\lambda_{\max} ^2}{\lambda_{\min} } \gamma^2\\
&< \bigg (
1 - \frac{\lambda_{\min} }{8}\gamma
\bigg )^{2(j+1)} \mathbb{E} [ | Y_0 - X|^2]
+ \frac{1}{1 - (
1 - \frac{\lambda_{\min} }{8}\gamma
)}2^5 n_x \frac{\lambda_{\max} ^2}{\lambda_{\min} } \gamma^2\\
&= \bigg (
1 - \frac{\lambda_{\min} }{8}\gamma
\bigg )^{2(j+1)} \mathbb{E} [ | Y_0 - X|^2]
+ 2^8 n_x \frac{\lambda_{\max} ^2}{\lambda_{\min} ^2} \gamma.
\end{split}
\end{equation}
Since $(1 - \frac{\lambda_{\min} }{8} \gamma ) \leq (\frac{1}{2})^{\frac{\lambda_{\min} }{8} \gamma}$ and  $Z_{(j+1)\gamma} = Y_{j+1}$, we conclude that 
\[
\mathbb{E} [| Y_{j+1} - X |^2 ] = \mathbb{E} [| Z_{(j+1)\gamma} - X |^2 ] < \bigg (
\frac{1}{2}
\bigg )^{\frac{\lambda_{\min}\gamma (j+1) }{4}} \mathbb{E} [ | Y_0 - X|^2]
+ 2^8 n_x \frac{\lambda_{\max}^2 }{\lambda_{\min}^2} \gamma.
\]
Replacing $j+1$ with $j$, the result follows.
\end{proof}

\begin{proof}[Proof of Theorem~\ref{thm:naive_ula}]

We now prove Theorem~\ref{thm:naive_ula}. It follows from \cite[Lemma 10]{mazumdar2020thompson} that 
\[
\mathbb{E}_{x \sim p}\big[|x-x_{\min}|^2\big]^{\frac{1}{2}} \leq 5 \sqrt{\frac{2n_x}{\lambda_{\min}}},
\]
where $x_{\min}$ is a minimizer of $U$. Using Lemma~\ref{lem:naive_ula} with $n_x = dn$ and 
the initial distribution $X_0 \sim \delta_{x_{\min}}$, we obtain that
\[
\mathbb{E}_{x \sim p, \tilde{x} \sim p_N}\big[|x-\tilde x|^2\big] < 2^{-\frac{\lambda_{\min} \gamma  N}{4}}
\mathbb{E}_{x \sim p}\big[|x-x_{\min}|^2\big] + 
2^8 \frac{n_x \lambda_{\max}^2}{ \lambda_{\min}^2 } \gamma.
\]
Taking the stepsize and the number of steps as 
$\gamma = \frac{\lambda_{\min}}{16\lambda_{\max}^2}$ and $N = \frac{64\lambda_{\max}^2}{\lambda_{\min}^2}$, respectively,
the first and second terms on the RHS of the inequality above are bounded as
\begin{equation}\nonumber
\begin{split}
2^{-\frac{\lambda_{\min} \gamma  N}{4}}
\mathbb{E}_{x \sim p}\big[|x-x_{\min}|^2\big] &= 
\frac{1}{2} \mathbb{E}_{x \sim p}\big[|x-x_{\min}|^2\big]\leq 25 \frac{n_x}{\lambda_{\min}},
\end{split}
\end{equation}
and
\[
2^8 \frac{n_x \lambda_{\max}^2 }{\lambda_{\min}^2 } \gamma \leq 
2^4 \frac{n_x}{\lambda_{\min}},
\]
respectively. 
Therefore, we conclude that
\[
\mathbb{E}_{x \sim p, \tilde{x} \sim p_N}\big[|x-\tilde x|^2\big]^{\frac{1}{2}} < \sqrt{41\frac{n_x}{\lambda_{\min}}} = O\bigg (\sqrt{\frac{1}{\lambda_{\min}}} \bigg )
\]
as desired.
\end{proof}

\subsection{Proof of Lemma~\ref{concavity}}\label{app:sec_a}
\begin{proof}
By direct calculation, we first observe that
\begin{equation*}
    \nabla^2_{\theta} \log p_w(x_{s+1}-\Theta^\top z_s) = \nabla^2_{w_s} \log p_w(x_{s+1}-\Theta^\top z_s) \otimes z_s z_s^{\top},
\end{equation*}
where $\otimes$ denotes Kronecker product.
Then, the Hessian $\nabla^2_{\theta}U_t$ is given by 
\begin{equation*}
    \nabla^2_{\theta}U_t = \lambda I_{dn} - \sum_{s=1}^{t-1}\nabla^2_{w_s} \log p_w(x_{s+1}-\Theta^\top z_s) \otimes z_s z_s^{\top}.
\end{equation*} 
Under Assumption~\ref{ass:noise}, for any state action pair $z_s=(x_s,u_s)$, we have
\begin{equation*}
  \underline m\text{blkdiag}(\{ z_s z_s^{\top}\}_{i=1}^{n})   \preceq -\nabla^2_{w_s} \log p_w(x_{s+1}-\Theta^\top z_s) \otimes z_s z_s^{\top} \preceq \overline m \text{blkdiag}(\{z_s z_s^{\top}\}_{i=1}^{n}),
\end{equation*}
which implies that
\begin{equation*}
\begin{split}
   \min\{ \underline m, 1\}\bigg( \lambda I_{dn}+\sum_{s=1}^{t-1}\text{blkdiag}(\{z_s z_s^{\top}\}_{i=1}^{n})\bigg) &\preceq  \nabla^2_{\theta}U_t\\
   &\preceq \max\{ \overline m, 1\}\bigg( \lambda I_{dn}+\sum_{s=1}^{t-1}\text{blkdiag}(\{z_s z_s^{\top}\}_{i=1}^{n})\bigg). 
   \end{split}
   \end{equation*}
Finally, letting the preconditioner $P_t := \lambda I_{dn} +\sum_{s=1}^{t-1}\text{blkdiag}(\{z_s z_s^{\top}\}_{i=1}^{n})$, the result follows.
\end{proof}

\subsection{Proof of Proposition~\ref{prop:ula}}
\label{app:prop:ula}

To prove Proposition~\ref{prop:ula}, we first introduce the following two lemmas regarding the stationarity of the preconditioned Langevin diffusion and the non-asymptotic behavior of the preconditioned ULA. 

\begin{lem}\label{lem:ld}
Suppose that Assumption~\ref{ass:noise} holds.
Let $X_{\tau} \in \mathbb{R}^{n_x}$ denote the solution of the preconditioned Langevin equation
\[
\mathrm{d} X_{\tau} = -P^{-1} \nabla U( X_{\tau}) \mathrm{d} {\tau} + \sqrt{2} P^{-\frac{1}{2}} \mathrm{d} B_{\tau},
\]
where $X_0$ is distributed according to $p(x) \propto e^{-U(x)}$, and 
 $P \in \mathbb{R}^{n_x \times n_x}$ is an arbitrary  positive definite matrix. 
Then, $X_{\tau}$ has the same probability density $p(x)$ for all ${\tau} \geq 0$. 
\end{lem}

\begin{proof}
Consider the following Fokker-Planck equation associated with the preconditioned Langevin equation:
\begin{equation}\label{fp}
\frac{\partial q (x, {\tau})}{\partial {\tau}} = -\sum_{i=1}^{n_x} \frac{\partial}{\partial x_i} \big ( 
[P^{-1} \nabla \log p (x) ]_i q (x, {\tau})
\big )
+ \sum_{i=1}^{n_x} \sum_{j=1}^{n_x} \frac{\partial^2}{\partial x_i \partial x_j} \big ( [P^{-1}]_{ij} q(x, {\tau})\big ).
\end{equation}
It is well known that $q(x,{\tau})$ is the probability density function of $X_{\tau}$. 
We can check that $p(x)$ is a solution of the Fokker-Planck equation by plugging $q(x, {\tau}) = p(x)$ into \eqref{fp}.
Specifically, 
\begin{equation}
\begin{split}
&-\sum_{i=1}^{n_x} \frac{\partial}{\partial x_i} \big ( 
[P^{-1} \nabla \log p (x) ]_i p(x) \big ) +  \sum_{i=1}^{n_x} \sum_{j=1}^{n_x} \frac{\partial^2}{\partial x_i \partial x_j} \big ( [P^{-1}]_{ij} p(x) \big )\\ &= 
-\sum_{i=1}^{n_x} \frac{\partial}{\partial x_i} \bigg ( \sum_{j=1}^{n_x}
[P^{-1}]_{ij} \frac{\partial}{\partial x_j} p (x)   \bigg) +  \sum_{i=1}^{n_x} \sum_{j=1}^{n_x} \frac{\partial^2}{\partial x_i \partial x_j} \big ( [P^{-1}]_{ij} p(x) \big ) = 
0 = \frac{\partial p(x)}{\partial {\tau}}.
\end{split}
\end{equation}
Since the Fokker-Planck equation has a unique smooth solution~\cite{pavliotis2014stochastic}, we conclude that $q(x, t) \equiv p(x)$ for all $t$, and the result follows. 
\end{proof}

\begin{lem}\label{lem:ula}
Suppose Assumption~\ref{ass:noise} holds.
Let $X \in \mathbb{R}^{n_x}$ be a random variable with probability density function $p(x) \propto e^{-U(x)}$, and the stochastic process $\{ Y_j\}$, $Y_j  \in \mathbb{R}^{n_x}$, be generated by the preconditioned ULA as
\[
Y_{j+1} = Y_j - \gamma P^{-1} \nabla U(Y_j) + \sqrt{2\gamma P^{-1}} W_j,
\]
where $Y_0$ is a random variable with an arbitrary density function, and 
$P \in \mathbb{R}^{n_x \times n_x}$ is a positive definite matrix with minimum eigenvalue $\lambda_{\min}$ and maximum eigenvalue $\lambda_{\max}$.
If  $\gamma \leq \frac{m \lambda_{\min}}{16 M^2 \max\{\lambda_{\min},t\}}$ and  $m I_{n_x} \preceq P^{-\frac{1}{2}}\nabla^2U P^{-\frac{1}{2}}\preceq MI_{n_x}$, then we have  
\[
\mathbb{E} [ | Y_j - X |_P^p ] < \bigg (
\frac{1}{2}
\bigg )^{\frac{m \gamma j}{4}} \mathbb{E} [ | Y_0 - X|_P^p]
+ 2^{4p+1} (pn_x)^{\frac{p}{2}} \frac{M^p}{m^p} \gamma^{\frac{p}{2}}
\]
for any $p\geq 2$ where $X$ and $Y_j$ are understood via the shared Brownian motion in continuous and discretized stochastic differential equations as demonstrated in Remark~\ref{rem:shared}.
\end{lem}

\begin{proof}
Let $\{Z_{\tau}\}_{{\tau}\geq 0}$ be a continuous interpolation of $\{Y_j\}$, defined by
\begin{equation}\label{eq:sde_z}
\left \{
\begin{array}{ll}
\mathrm{d} Z_{\tau} = - P^{-1} \nabla U(Y_j) \mathrm{d}{\tau} + \sqrt{2P^{-1}} \mathrm{d} B_{\tau}& \mbox{for }{\tau} \in [j\gamma, (j+1) \gamma) \\
Z_{\tau} = Y_j & \mbox{for } {\tau} = j\gamma.
\end{array}
\right.
\end{equation}
Note that $\lim_{{\tau} \nearrow j\gamma} Z_{\tau} = Y_j =  \lim_{{\tau} \searrow j\gamma} Z_{\tau}$ for each $j$, and thus $\{Z_{\tau}\}$ is a continuous process. 
We introduce another stochastic process $\{X_{\tau}\}$, defined by 
\[
\mathrm{d} X_{\tau} = -P^{-1} \nabla U( X_{\tau}) \mathrm{d} {\tau} + \sqrt{2} P^{-\frac{1}{2}} \mathrm{d} B_{\tau},
\]
where $X_0$ is a random variable with pdf $p(x) \propto e^{-U(x)}$.
By Lemma~\ref{lem:ld}, $X_{\tau}$ has the same pdf $p(x)$ for all ${\tau}$. 
We use the same Brownian motion $B_{\tau}$ to define both $\{Z_{\tau}\}$ and $\{X_{\tau}\}$.

Fix an arbitrary $j$.
For any $p \geq 2$, 
differentiating $|Z_{\tau} - X_{\tau}|_P^p = |P^{\frac{1}{2}}(Z_{\tau} - X_{\tau})|^p$ with respect to ${\tau} \in [j\gamma, (j+1)\gamma)$, we have
\begin{align*}
\frac{\mathrm{d} |Z_{\tau} - X_{\tau}|_P^p}{\mathrm{d}{\tau}} &= p|P^{\frac{1}{2}}(Z_{\tau} - X_{\tau})|^{p-2}(Z_{\tau} - X_{\tau})^\top P \bigg (
\frac{\mathrm{d} Z_{\tau}}{\mathrm{d} {\tau}} - \frac{\mathrm{d} X_{\tau}}{\mathrm{d} {\tau}}
\bigg )
\\&= p|P^{\frac{1}{2}}(Z_{\tau} - X_{\tau})|^{p-2} (Z_{\tau} - X_{\tau})^\top ( - \nabla U (Y_j) +  \nabla U(Z_{\tau}))
\\& \quad + p|P^{\frac{1}{2}}(Z_{\tau} - X_{\tau})|^{p-2} (Z_{\tau} - X_{\tau})^\top (  - \nabla U(Z_{\tau}) +  \nabla U(X_{\tau})).
\end{align*}
Noting that $m  I_{n_x} \preceq P^{-\frac{1}{2}}\nabla^2 U P^{-\frac{1}{2}}  \preceq  M  I_{n_x}$, we have
\begin{equation}\nonumber
\begin{split}
&p|P^{\frac{1}{2}}(Z_{\tau} - X_{\tau})|^{p-2}\big[ (Z_{\tau} - X_{\tau})^\top ( - \nabla U (Y_j) + \nabla U(Z_{\tau}))
+ (Z_{\tau} - X_{\tau})^\top (  -  \nabla U(Z_{\tau}) +  \nabla U(X_{\tau}))\big]\\
&\leq p|P^{\frac{1}{2}}(Z_{\tau} - X_{\tau})|^{p-2} \big[(Z_{\tau} - X_{\tau})^\top P^{\frac{1}{2}} P^{-\frac{1}{2}}( - \nabla U (Y_j) +  \nabla U(Z_{\tau}))
- m (Z_{\tau} - X_{\tau})^\top P (Z_{\tau} - X_{\tau})\big]\\
&=p|P^{\frac{1}{2}}(Z_{\tau} - X_{\tau})|^{p-2}\big[  |Z_{\tau} - X_{\tau}|_P  |  P^{-\frac{1}{2}} \nabla U(Z_{\tau}) -P^{-\frac{1}{2}} \nabla U (Y_j) |
- m  |Z_{\tau} - X_{\tau}|_P^2\big],
\end{split}
\end{equation}
where the first inequality follows from the mean value theorem.
Now, recall the generalized Young's inequality, $ab \leq \frac{s^\alpha a^\alpha}{\alpha} + \frac{s^{-\beta}b^\beta}{\beta}$ for $s>0$, $a,b,\alpha,\beta>0$ such that $\frac{1}{\alpha}+\frac{1}{\beta}=1$.
Choosing $s=(\frac{pm}{2(p-1)})^{(p-1)/p}$, $\alpha=\frac{p}{p-1}$, and $\beta=p$ yields
\begin{align*}
&|Z_{\tau} - X_{\tau}|_P^{p-1}  |  P^{-\frac{1}{2}} \nabla U(Z_{\tau}) -P^{-\frac{1}{2}} \nabla U (Y_j) | \\
&\leq \frac{p-1}{p}\frac{pm}{2(p-1)} | Z_{\tau} - X_{\tau}|_P^p  + \frac{1}{p}\frac{1}{(\frac{pm}{2(p-1)})^{p-1}}|  P^{-\frac{1}{2}} \nabla U(Z_{\tau}) -P^{-\frac{1}{2}} \nabla U (Y_j) |^p.
\end{align*}
Combining all together with $\frac{pm}{2(p-1)}\geq \frac{m}{2}$, we have
\[
\frac{\mathrm{d} |Z_{\tau} - X_{\tau}|_P^p}{\mathrm{d}\tau} \leq - \frac{pm}{2}  | Z_{\tau} - X_{\tau}|_P^p + \frac{2^{p-1}}{m^{p-1}} |  P^{-\frac{1}{2}} \nabla U(Z_{\tau}) -P^{-\frac{1}{2}} \nabla U (Y_j) |^p,
\]
which implies that
\[
\frac{\mathrm{d}}{\mathrm{d} {\tau}} (e^{\frac{pm}{2}  {\tau}} | Z_{\tau} - X_{\tau} |_P^p) \leq 
e^{\frac{pm}{2}  {\tau}}\frac{2^{p-1}}{m^{p-1}} |  P^{-\frac{1}{2}} \nabla U(Z_{\tau}) -P^{-\frac{1}{2}} \nabla U (Y_j) |^p.
\]
Integrating both sides from $j\gamma$ to $(j+1)\gamma$ and then multiplying both sides by $e^{-\frac{pm}{2} (j+1) \gamma}$, we obtain that
\begin{align*}
&| Z_{(j+1)\gamma} - X_{(j+1)\gamma} |_P^p\\
&\leq 
e^{-\frac{pm}{2}  \gamma} | Z_{j\gamma} - X_{j\gamma} |_P^p
+\frac{2^{p-1}}{m^{p-1}} \int_{j\gamma}^{(j+1) \gamma} 
e^{-\frac{pm}{2} ( (j+1)\gamma - s)} |  P^{-\frac{1}{2}} \nabla U(Z_s) -P^{-\frac{1}{2}} \nabla U (Y_j) |^p \mathrm{d} s.
\end{align*}
Since $X_{\tau}$ and $X$ have the same pdf due to Lemma~\ref{lem:ld}, we have
\begin{align}
&\mathbb{E} [| Z_{(j+1)\gamma} - X |^p_P ] \nonumber\\ &\leq 
e^{-\frac{pm}{2} \gamma} \mathbb{E} [ | Z_{j\gamma} - X |_P^p]
+\frac{2^{p-1}}{m^{p-1}} \int_{j\gamma}^{(j+1) \gamma} 
 \mathbb{E} [ |  P^{-\frac{1}{2}} \nabla U(Z_s) -P^{-\frac{1}{2}} \nabla U (Y_j) |^p ]\mathrm{d} s \nonumber\\
 & = e^{-\frac{pm}{2} \gamma} \mathbb{E} [ | Z_{j\gamma} - X |_P^p]
+\frac{2^{p-1}}{m^{p-1}} \int_{j\gamma}^{(j+1) \gamma} 
 \mathbb{E} [ |  P^{-\frac{1}{2}} (\int_0^1 \nabla^2 U(Y_j + t(Z_s-Y_j))\mathrm{d}t)P^{-\frac{1}{2}}P^\frac{1}{2}(Z_s-Y_j)  |^p ]\mathrm{d} s \nonumber\\
 &\leq e^{-\frac{pm}{2}  \gamma} \mathbb{E} [ | Z_{j\gamma} - X |_P^p]
+\frac{2^{p-1}M^{p}}{m^{p-1}} \int_{j\gamma}^{(j+1) \gamma} 
\mathbb{E} [ | P^{\frac{1}{2}}(Z_s-Y_j) |^p ] \mathrm{d} s,\label{eq:bd}
\end{align}
where the first inequality follows from $e^{-m ( (j+1)\gamma - s)}  \leq 1$ and the second inequality follows since $M$ is an upper bound for $|P^{-\frac{1}{2}}\nabla^2UP^{-\frac{1}{2}}|$ from the assumption in the lemma. 
To bound~\eqref{eq:bd}, we handle the first and second terms, separately. 

For the second term, we integrate \eqref{eq:sde_z} from $j\gamma$ to $s \in [j \gamma, (j+1)\gamma )$  to obtain 
\begin{equation} \nonumber
 Z_s - Y_j   =    -(s - j\gamma) P^{-1} \nabla U(Y_j) + \sqrt{2P^{-1}} (B_s - B_{j\gamma}).  
\end{equation}
Ignoring the constant coefficient, the second term of \eqref{eq:bd} is then bounded by
\begin{equation}\label{eq:bd0}
\begin{split}
&\int_{j\gamma}^{(j+1) \gamma} 
\mathbb{E} [ | P^{\frac{1}{2}}(Z_s-Y_j)|^p ] \mathrm{d} s 
\\
&= \int_{j\gamma}^{(j+1) \gamma}\mathbb{E} [ | 
-(s - j\gamma) P^{-\frac{1}{2}} \nabla U(Y_j) + \sqrt{2} (B_s - B_{j\gamma})
|^p ] \mathrm{d} s \\
&\leq 2^{p-1}\bigg [
 \int_{j\gamma}^{(j+1) \gamma} \mathbb{E} [ | 
(s - j\gamma) P^{-\frac{1}{2}} \nabla U(Y_j) |^p] \mathrm{d} s
+ 2^{\frac{p}{2}} \int_{j\gamma}^{(j+1) \gamma} \mathbb{E} [|B_s - B_{j\gamma} |^p] \mathrm{d}s \bigg ].
\end{split}
\end{equation}
For $s \in [j\gamma, (j+1) \gamma)$, we note that $|s - j\gamma| \leq \gamma$, and thus
\begin{equation}\label{eq:bd1}
\begin{split}
\int_{j\gamma}^{(j+1) \gamma} \mathbb{E} [ | 
(s - j\gamma) P^{-\frac{1}{2}} \nabla U(Y_j) |^p] \mathrm{d} s
&\leq \gamma^p \int_{j\gamma}^{(j+1) \gamma}  \mathbb{E} [
|P^{-\frac{1}{2}} \nabla U(Y_j) |^p
] \mathrm{d} s \\
&= \gamma^{p+1} \mathbb{E} [
|P^{-\frac{1}{2}} \nabla U(Y_j) |^p
] \\
&= \gamma^{p+1} \mathbb{E} [
|P^{-\frac{1}{2}} \nabla U(Y_j) - P^{-\frac{1}{2}} \nabla U(x_{\min}) |^p
]  \\
&\leq \gamma^{p+1} M^p  \mathbb{E} [ | Y_j - x_{\min} |_P^p ], 
\end{split}
\end{equation}
where $x_{\min}$ is a minimizer of potential $U$.

Let $\tilde{X} := P^{\frac{1}{2}}X$. Its pdf is denoted by by $ \tilde{p}(\tilde{x})$.
Then, for any $p\geq 2$,
\begin{equation}
\begin{split}
\mathbb{E} [ | Y_j - x_{\min}|_P^p] 
&\leq 2^{p-1} (
\mathbb{E} [ | Y_j - X|_P^p]
+\mathbb{E} [ | \tilde{X} - \tilde{x}_{\min}|^p]
),
\end{split}
\end{equation}
where $\tilde{x}_{\min} = P^{\frac{1}{2}} x_{\min}$. 
Since $\tilde{p}(\tilde{x}) = \det(P^{-\frac{1}{2}}) p (P^{-\frac{1}{2}}\tilde{x})$, we have
$-\nabla^2_{\tilde{x}} \log \tilde{p}(\tilde{x}) = -P^{-\frac{1}{2}} \nabla^2_x \log p(P^{-\frac{1}{2}}\tilde x) P^{-\frac{1}{2}}$.
Thus, $\tilde{p}$ is $m$-strongly log-concave. 
It follows from \cite[Lemma 9]{mazumdar2020thompson} that 
\begin{equation}\label{eq:bd2}
\begin{split}
\mathbb{E} [ | Y_j - x_{\min}|_P^p] & \leq 2^{p-1}\mathbb{E} [ | Y_j - X|_P^p] + \frac{10^p}{2} \big(\frac{pn_x}{m}\big)^{\frac{p}{2}}.
\end{split}
\end{equation}
On the other hand, \cite[Lemma 8]{mazumdar2020thompson} yields that
\begin{equation}\label{eq:bd3}
\int_{j\gamma}^{(j+1)\gamma} \mathbb{E} [
| B_s - B_{j\gamma}|^p
]\mathrm{d}s \leq 2\big(\frac{pn_x}{e}\big)^{\frac{p}{2}} \gamma^{\frac{p}{2}+1}.
\end{equation}
Combining \eqref{eq:bd0}--\eqref{eq:bd3}, we obtain that \begin{equation}
\begin{split}
&\int_{j\gamma}^{(j+1)\gamma} \mathbb{E} [
| Z_s - Y_j |_P^p
]\mathrm{d} s \\
&\leq 
2^{2p-2} M^p \gamma^{p+1} \mathbb{E} [ | Y_j - X |_P^p ] + 2^{p-2}(10 M)^p \gamma^{p+1} \big(\frac{pn_x}{m}\big)^{\frac{p}{2}} +  2^{\frac{3p}{2}}\big(\frac{pn_x}{e}\big)^{\frac{p}{2} }\gamma^{\frac{p}{2}+1}\\
&\leq 2^{2p-2} M^p \gamma^{p+1} \mathbb{E} [ | Y_j - X |_P^p ] + 2^{3p} (pn_x)^{\frac{p}{2}} \gamma^{\frac{p}{2}+1},
\\& 
\end{split}
\end{equation}
where the second inequality follows from $\gamma \leq \frac{m \lambda_{\min}}{16 M^2\max\{\lambda_{\min},t\}} \leq \frac{m}{16 M^2}$. 
Plugging this inequality into \eqref{eq:bd} yields
\begin{align*}
&\mathbb{E} [| Z_{(j+1)\gamma} - X |_P^p ]  \\
&\leq e^{-\frac{pm}{2}  \gamma} \mathbb{E} [ | Z_{j\gamma} - X |_P^p]
+2^{3p-3}\frac{M^{2p}}{m^{p-1}} \gamma^{p+1} \mathbb{E} [ | Y_j - X |_P^p ]
+ 2^{4p-1} (pn_x)^{\frac{p}{2}} \frac{M^p}{m^{p-1} } \gamma^{\frac{p}{2}+1}.
\end{align*}

To further simplify the first two terms on the right-hand side, we use the following inequalities:
\begin{align*}
2^{3p-3}\frac{M^{2p}}{m^{p-1}} \gamma^{p+1} &= \frac{m}{2^{p+3} } \bigg (\frac{16 M ^2 \max\{\lambda_{\min},t\}}{m \lambda_{\min}}\bigg )^p\bigg(\frac{\lambda_{\min}}{\max\{\lambda_{\min},t\}} \bigg)^p \gamma^{p+1}
\leq \frac{m}{32 }  \gamma\\
e^{-\frac{pm}{2}\gamma} + \frac{m }{32}  \gamma 
&\leq
e^{-m\gamma} + \frac{m }{32}  \gamma \leq 1-\frac{m}{2}\gamma+\frac{m }{32}  \gamma < 1-\frac{m}{4}\gamma,
\end{align*}
where the second line follows from the fact that 
$e^{-x} \leq 1 - \frac{x}{2}$ for $x \in [0,1]$.
Consequently, $\mathbb{E} [| Z_{(j+1)\gamma} - X |_P^p ]$ is bounded as
\[
\mathbb{E} [| Z_{(j+1)\gamma} - X |_P^p ] < \bigg (
1 - \frac{m}{4}\gamma
\bigg )\mathbb{E} [ | Y_j - X|_P^p]  + 2^{4p-1} (pn_x)^{\frac{p}{2}} \frac{M^p}{m^{p-1} } \gamma^{\frac{p}{2}+1}.
\]
Invoking the bound repeatedly, we obtain that
\begin{equation}
\begin{split}
\mathbb{E} [| Z_{(j+1)\gamma} - X |_P^p ] &<
\bigg (
1 - \frac{m}{4}\gamma
\bigg )^{(j+1)} \mathbb{E} [ | Y_0 - X|_P^p]
+ \sum_{i=0}^j \bigg (
1 - \frac{m}{4}\gamma
\bigg )^{i} 2^{4p-1} (pn_x)^{\frac{p}{2}} \frac{M^p}{m^{p-1} } \gamma^{\frac{p}{2}+1}\\
&< \bigg (
1 - \frac{m}{4}\gamma
\bigg )^{(j+1)} \mathbb{E} [ | Y_0 - X|_P^p]
+ \frac{1}{1 - (
1 - \frac{m}{4}\gamma
)}2^{4p-1} (pn_x)^{\frac{p}{2}} \frac{M^p}{m^{p-1} } \gamma^{\frac{p}{2}+1}\\
&= \bigg (
1 - \frac{m}{4}\gamma
\bigg )^{(j+1)} \mathbb{E} [ | Y_0 - X|_P^p]
+ 2^{4p+1} (pn_x)^{\frac{p}{2}} \frac{M^p}{m^p} \gamma^{\frac{p}{2}}\nonumber.
\end{split}
\end{equation}
Since $(1 - \frac{m }{4} \gamma ) \leq (\frac{1}{2})^{\frac{m }{4} \gamma}$,  $Z_{(j+1)\gamma} = Y_{j+1}$, we conclude that
\[
\mathbb{E} [| Y_{j+1} - X |_P^p ] = \mathbb{E} [| Z_{(j+1)\gamma} - X |_P^p ] < \bigg (
\frac{1}{2}
\bigg )^{\frac{m \gamma (j+1) }{4}} \mathbb{E} [ | Y_0 - X|_P^p]
+ 2^{4p+1} (pn_x)^{\frac{p}{2}} \frac{M^p}{m^p} \gamma^{\frac{p}{2}}.
\]
Replacing $j+1$ with $j$, the result follows.
\end{proof}

We are now ready to prove Proposition~\ref{prop:ula}.

\begin{proof}[Proof of Proposition~\ref{prop:ula}]
For simplicity, the following notation
is used throughout the proof: for a positive definite matrix $P$, we let
\[
E^p_P (\mu, \tilde\mu | h) := \mathbb{E}_{x \sim \mu, \tilde x \sim \tilde\mu} [ | x - \tilde x|_P^p | h].
\]
We also let $\lambda_{\max,t}$ and $\lambda_{\min,t}$ denote the maximum and minimum eigenvalues of $P_t$, respectively.

Since $\mu_t$ is $m$-strongly log-concave distribution, 
it follows from \cite[Lemma 10]{mazumdar2020thompson} that 
\begin{equation}\label{eq:lem}
E^p_{P_t} ( \mu_t, \delta (\theta_{\min, t}) | h_{t}) \leq 5^p \bigg(\frac{pdn}{m}\bigg)^{\frac{p}{2}}
\end{equation}
for all $t$.
We then use Lemma~\ref{lem:ula} with $n_x = dn$ and 
the initial distribution $\theta_0 \sim \delta_ {\theta_{\min, t}}$ in Algorithm~\ref{alg:ula} 
to obtain that 
\[
E^p_{P_t}(\mu_{t}, \tilde{\mu}_{t}| h_{t}) < 2^{-\frac{m \gamma_t  N_t}{4}}
E^p_{P_t}(\mu_{t}, \delta(\theta_{\min, t})| h_{t}) + 
2^{4p+1} (pdn)^{\frac{p}{2}} \frac{M^p}{m^p} \gamma_t^{\frac{p}{2}}.
\]
In Algorithm~\ref{alg:ula}, the stepsize and number of iterations are chosen to be 
$\gamma_t = \frac{m \lambda_{\min,t}}{16M^2\max\{\lambda_{\min,t},t\}}$ and $N_t = \frac{4\log_2 (\max\{\lambda_{\min,t},t\}/\lambda_{\min,t})}{m \gamma_t}$. 
Thus, the first and second terms on the right-hand side of  the inequality above are bounded as
\begin{equation}\nonumber
\begin{split}
2^{-\frac{\gamma_t m N_t}{4}}
E_{P_t}^p(\mu_{t}, \delta(\theta_{\min, t})| h_{t}) &= 
2^{-\log_2 (\max\{\lambda_{\min,t},t\}/\lambda_{\min,t})} E_{P_k}^p(\mu_{t}, \delta(\theta_{\min, t})| h_{t})\\
&\leq 5^p \bigg(\frac{pdn}{m}\bigg)^{\frac{p}{2}}
 \bigg(\frac{\lambda_{\min,t}}{\max\{\lambda_{\min,t},t\}}\bigg),\\
\end{split}
\end{equation}
and
\[
2^{4p+1} (pdn)^{\frac{p}{2}} \frac{M^p}{m^p} \gamma_t^{\frac{p}{2}} \leq 
2^{2p+1} \frac{(pdn)^{\frac{p}{2}}}{m^{\frac{p}{2}}} \bigg(\frac{\lambda_{\min,t}}{\max\{\lambda_{\min,t},t\}}\bigg)^{\frac{p}{2}},
\]
respectively.
Therefore, we conclude that
\begin{align*}
E^p_{P_t}(\mu_{t}, \tilde{\mu}_{t}| h_{t}) &< \bigg(\frac{pdn}{m}\bigg)^{\frac{p}{2}}\bigg(5^{p}\frac{\lambda_{\min,t}}{\max\{\lambda_{\min,t},t\}}+2^{2p+1}\bigg(\frac{\lambda_{\min,t}}{\max\{\lambda_{\min,t},t\}}\bigg)^{\frac{p}{2}}  \bigg)\\
&\leq \bigg(\frac{pdn}{m}\bigg)^{\frac{p}{2}}\big(2^{2p+1}+5^{p}\big).
\end{align*}

For the special case with $p=2$, a simpler bound is attained. 
Using the inequality 
\[
\lambda_{\min,t}\mathbb{E}_{\theta_t\sim\mu_{t},\tilde\theta_t\sim\tilde \mu_{t}}{[|\theta_t-\tilde{\theta}_t|^2 \mid h_{t}]} \leq  E^2_{P_t}(\mu_{t},\tilde \mu_{t}\mid h_{t}),
\]
one can deduce that
\begin{align*}
\mathbb{E}_{\theta_t\sim\mu_{t},\tilde\theta_t\sim\tilde \mu_{t}}{[|\theta_t-\tilde{\theta}_t|^2 \mid h_{t}]}^{\frac{1}{2}} &<
\sqrt{\frac{1}{\lambda_{\min,t}}\bigg(\frac{2dn}{m}\bigg)\bigg(5^{2}\frac{\lambda_{\min,t}}{\max\{\lambda_{\min,t},t\}}+2^{5}\frac{\lambda_{\min,t}}{\max\{\lambda_{\min,t},t\}}\bigg)}\\
&= 
 \sqrt{\frac{{D}}{{\max\{\lambda_{\min,t},t\}}}},
\end{align*}
where $ D = 114\frac{dn}{m}$.
\end{proof}

\subsection{Proof of Proposition~\ref{prop:half_exact_true}}
\label{app:prop:half_exact_true}
\begin{proof}
Fix an arbitrary $t$. Given $\theta_0 \in \mathbb{R}^{dn}$, let $\theta_\tau \in \mathbb{R}^{dn}$ denote the solution of the following SDE:
\[
\mathrm{d}\theta_\tau = -P_t^{-1}\nabla U_t(\theta_\tau)\mathrm{d}\tau + \sqrt{2}P_t^{-\frac{1}{2}}\mathrm{d}B_\tau,
\]
where $P_t= \lambda {I_{dn}} + \sum_{s=1}^{t-1} \mathrm{blkdiag}(\{z_sz_s^{\top}\}_{i=1}^n)$ and $U_t = U_1 +U_t'$ with $U_t' =\sum_{s=1}^{t-1}\log p_w(x_{s+1}-\Theta^{\top}z_s)$.
Define $V(\tau)$ as
\[
V(\tau) = \frac{1}{2}e^{\alpha \tau}|\theta_\tau-\theta_*|_{P_t}^2,
\]
for a fixed $\alpha>0$. 
Applying Ito's lemma to $V(\tau)$ yields
\[
V(\tau) - V(0) = F_1 + F_2 + F_3,
\]
where 
\begin{equation}\nonumber
\begin{split}
    &F_1 = \int_0^\tau e^{\alpha \eta}\nabla_{\theta}U_t(\theta_\eta)^{\top}(\theta_*-\theta_\eta) \mathrm{d} \eta + \frac{\alpha}{2}\int_0^\tau e^{\alpha \eta}|\theta_\eta-\theta_*|_{P_t}^2 \mathrm{d} \eta,\\
    &F_2 = dn \int_0^\tau e^{\alpha \eta}\mathrm{d} \eta,\\
    &F_3 = \sqrt{2}\int_0^\tau e^{\alpha \eta} (\theta_\eta - \theta_*)^{\top}P_t^{\frac{1}{2}}\mathrm{d}B_\eta.
\end{split}
\end{equation}

We first expand $F_1$ as follows:
\begin{equation}\nonumber
\begin{split}
    F_1 &= \int_0^\tau e^{\alpha \eta}\nabla_{\theta}U_t(\theta_\eta)^{\top}(\theta_*-\theta_\eta)\mathrm{d}\eta + \frac{\alpha}{2}\int_0^\tau e^{\alpha \eta}|\theta_\eta-\theta_*|^2_{P_t} \mathrm{d}\eta\\
    &=- \int_0^\tau e^{\alpha \eta}(\nabla_{\theta}U_t(\theta_\eta)-\nabla_{\theta}U_t(\theta_*))^{\top}(\theta_\eta-\theta_*)\mathrm{d}\eta + \frac{\alpha}{2}\int_0^\tau e^{\alpha \eta}|\theta_\eta-\theta_*|_{P_t}^2 \mathrm{d}\eta \\
    &  \quad +\int_0^\tau e^{\alpha \eta}\nabla_{\theta}U_1(\theta_*)^{\top}(\theta_*-\theta_\eta)\mathrm{d}\eta+  \int_0^\tau e^{\alpha \eta} \nabla_{\theta} U'_t(\theta_*)^{\top}(\theta_*-\theta_\eta)\mathrm{d}\eta\\
    &\leq -m \int_0^\tau e^{\alpha \eta}(\theta_\eta-\theta_*)^{\top}P_t(\theta_\eta-\theta_*)\mathrm{d}\eta + \frac{\alpha}{2}\int_0^\tau e^{\alpha \eta}|\theta_\eta-\theta_*|_{P_t}^2 \mathrm{d}\eta\\
    & \quad  +\int_0^\tau e^{\alpha \eta}\nabla_{\theta}U_1(\theta_*)^{\top}(\theta_*-\theta_\eta)\mathrm{d}\eta+ \int_0^\tau e^{\alpha \eta} \nabla_{\theta} U'_t(\theta_*)^{\top}(\theta_*-\theta_\eta)\mathrm{d}\eta\\
    &\leq \frac{\alpha-2m}{2}\int_0^\tau e^{\alpha \eta}|\theta_\eta-\theta_*|_{P_t}^2 \mathrm{d}\eta +  \int_0^\tau e^{\alpha \eta} \nabla_{\theta} U_1(\theta_*)^{\top}(\theta_*-\theta_\eta)\mathrm{d}\eta\\
    &\quad + \int_0^\tau e^{\alpha \eta} \nabla_{\theta} U'_t(\theta_*)^{\top}(\theta_*-\theta_\eta)\mathrm{d}\eta.
\end{split}
\end{equation}
It follows from Young's inequality that the second and third terms on the right-hand side can be bounded as follows:
\begin{equation*}
\begin{split}
    \int_0^\tau e^{\alpha \eta} \nabla_{\theta} U_1(\theta_*)^{\top}(\theta_*-\theta_\eta)\mathrm{d}\eta &\leq \int_0^\tau e^{\alpha \eta}|P_t^{-\frac{1}{2}}\nabla_{\theta} U_1(\theta_*)||P_t^{\frac{1}{2}}(\theta_*-\theta_\eta)|\mathrm{d}\eta\\
    & \leq \frac{1}{m}\int_0^\tau e^{\alpha \eta}|P_t^{-\frac{1}{2}}\nabla_{\theta} U_1(\theta_*)|^2 \mathrm{d}\eta +\frac{m}{4}\int_0^\tau e^{\alpha \eta}|\theta_*-\theta_\eta|_{P_t}^2 \mathrm{d}\eta,
\end{split}
\end{equation*}
and
\begin{equation*}
\begin{split}
    \int_0^\tau e^{\alpha \eta} \nabla_{\theta} U'_t(\theta_*)^{\top}(\theta_*-\theta_\eta)\mathrm{d}\eta &\leq \int_0^\tau e^{\alpha \eta}|P_t^{-\frac{1}{2}}\nabla_{\theta} U'_t(\theta_*)||P_t^{\frac{1}{2}}(\theta_*-\theta_\eta)|\mathrm{d}\eta\\
    & \leq \frac{1}{m}\int_0^\tau e^{\alpha \eta}|P_t^{-\frac{1}{2}}\nabla_{\theta} U'_t(\theta_*)|^2 \mathrm{d}\eta +\frac{m}{4}\int_0^\tau e^{\alpha \eta}|\theta_*-\theta_\eta|_{P_t}^2 \mathrm{d}\eta.
\end{split}
\end{equation*}
Putting everything together, we have
\begin{align*}
F_1 &\leq \frac{\alpha-m}{2}\int_0^\tau e^{\alpha \eta}|\theta_\eta-\theta_*|_{P_t}^2 \mathrm{d}\eta +\frac{1}{m} \int_0^\tau e^{\alpha \eta} |P_t^{-\frac{1}{2}}\nabla_{\theta} U_1(\theta_*)|^2\mathrm{d}\eta\\
&\quad +\frac{1}{m} \int_0^\tau e^{\alpha \eta} |P_t^{-\frac{1}{2}}\nabla_{\theta} U'_t(\theta_*)|^2\mathrm{d}\eta.
\end{align*}
Let $\alpha = m$. We then obtain that
\begin{align*}
F_1 &\leq \frac{1}{m} \int_0^\tau e^{\alpha \eta} |P_t^{-\frac{1}{2}}\nabla_{\theta} U_1(\theta_*)|^2\mathrm{d}\eta +\frac{1}{m}\int_0^\tau e^{\alpha \eta}|P_t^{-\frac{1}{2}}\nabla_{\theta} U'_t(\theta_*)|^2 \mathrm{d}\eta\\
& \leq C_0 e^{\alpha \tau}+\frac{1}{m}\int_0^\tau e^{\alpha \eta}|P_t^{-\frac{1}{2}}\nabla_{\theta} U'_t(\theta_*)|^2 \mathrm{d}\eta
\end{align*}
for some positive constant $C_0$ depending only on $m, n, d$ and $\lambda$. 

On the other hand, $F_2$ is bounded as 
\[
F_2 = dn\int_0^\tau e^{\alpha \eta}\mathrm{d}\eta = \frac{dn}{\alpha}(e^{\alpha \tau}-1) \leq \frac{dn}{\alpha}e^{\alpha \tau} = \frac{dn}{m}e^{\alpha \tau}.
\]

Regarding $F_3$, we use the Burkholder-Davis-Gundy inequality \cite{ren2008burkholder} to obtain that for a fixed $\Delta>0$
\begin{equation}\nonumber
\begin{split}
    \mathbb{E}\Big[\sup_{0\leq \tau \leq \Delta} |F_3|\Big] &\leq 4 \mathbb{E}\bigg[ \bigg(\int_0^\Delta e^{2\alpha \eta}|\theta_\eta-\theta_*|_{P_t}^2 \mathrm{d}\eta\bigg)^{\frac{1}{2}}\bigg]\\
    &\leq 4 \mathbb{E}\bigg[\bigg(\sup_{0\leq \tau \leq \Delta}e^{\alpha \tau}|\theta_\tau-\theta_*|_{P_t}^2 \int_0^\Delta e^{\alpha \eta}\mathrm{d} \eta\bigg)^{\frac{1}{2}}\bigg]\\
    & = 4 \mathbb{E}\bigg[\bigg(\sup_{0\leq \tau \leq \Delta}e^{\alpha \tau}|\theta_\tau-\theta_*|_{P_t}^2 \bigg(\frac{e^{\alpha \Delta}-1}{\alpha}\bigg)\bigg)^{\frac{1}{2}}\bigg]\\
    & \leq \mathbb{E} \bigg[\bigg(\frac{16 e^{\alpha \Delta}}{\alpha}\bigg)^{\frac{1}{2}}\Big (\sup_{0\leq \tau \leq \Delta}e^{\alpha \tau}|\theta_\tau-\theta_*|_{P_t}^2 \Big )^{\frac{1}{2}} \bigg],
\end{split}
\end{equation}
where the expectation is taken with respect to $\theta_\tau$.
By Young's inequality, we further have
\begin{equation}\nonumber
\begin{split}
    \mathbb{E} \bigg[\bigg(\frac{16 e^{\alpha \Delta}}{\alpha}\bigg)^{\frac{1}{2}}\Big (\sup_{0\leq \tau \leq \Delta}e^{\alpha \tau}|\theta_\tau-\theta_*|_{P_t}^2 \Big )^{\frac{1}{2}} \bigg]&\leq \mathbb{E}\bigg[\frac{16 e^{\alpha \Delta}}{\alpha} + \frac{1}{4}\sup_{0\leq \tau \leq \Delta}e^{\alpha \tau}|\theta_\tau-\theta_*|_{P_t}^2  \bigg]\\
    &= \frac{16}{m} e^{\alpha \Delta} + \frac{1}{2}\mathbb{E}\bigg[\sup_{0 \leq \tau \leq \Delta}V(\tau)\bigg].
\end{split}
\end{equation}

Putting everything together, we finally have the following bound for $V$:
\begin{equation}\label{sup_v}
\begin{split}
\mathbb{E}\bigg[\sup_{0 \leq \tau \leq \Delta} V(\theta_\tau) \bigg]&=\mathbb{E}\bigg[\sup_{0 \leq \tau \leq \Delta}(F_1+F_2+F_3) \bigg] + V(0)\\
& \leq \mathbb{E}\bigg[\sup_{0 \leq \tau \leq \Delta}F_1\bigg]+\mathbb{E}\bigg[\sup_{0 \leq \tau \leq \Delta}F_2\bigg]+\mathbb{E}\bigg[\sup_{0 \leq \tau \leq \Delta}F_3\bigg] + V(0)\\
& \leq \mathbb{E}\bigg[C_0+\frac{1}{m^2}|P_t^{-\frac{1}{2}}\nabla_{\theta} U'_t(\theta_*)|^2 + \frac{dn+16}{m} \bigg]e^{\alpha \Delta} + \frac{1}{2}\mathbb{E}\bigg[\sup_{0 \leq \tau \leq \Delta}V(\tau)\bigg] + V(0),
\end{split}
\end{equation}
which implies that
\[
\mathbb{E}\bigg[\sup_{0 \leq \tau \leq \Delta} V(\tau) \bigg] \leq
2\bigg(C_0+\frac{1}{m^2}|P_t^{-\frac{1}{2}}\nabla_{\theta} U'_t(\theta_*)|^2 + \frac{dn+16}{m} \bigg)e^{\alpha \Delta} + 2V(0).
\]
We then have
\begin{equation*}
\begin{split}
    \mathbb{E}[|\theta_{\Delta}-\theta_*|_{P_t}|h_t]  = \mathbb{E}[\sqrt{2}e^{-\frac{1}{2}\alpha \Delta}V(\theta_{\Delta})^{\frac{1}{2}}]
    & \leq \sqrt{2} e^{-\frac{1}{2}\alpha \Delta} \bigg(\mathbb{E}\Big [\sup_{0\leq \tau \leq \Delta}V(\tau) \Big ]\bigg)^{\frac{1}{2}}\\
    & \leq 2\sqrt{C_0+\frac{1}{m^2}|P_t^{-\frac{1}{2}}\nabla_{\theta} U'_t(\theta_*)|^2 + \frac{dn+16}{m} + V(0)e^{-\alpha\Delta}}.
\end{split}
\end{equation*}
Letting $\Delta \to \infty$ and using Fatou's lemma, we have
\[
\mathbb{E}_{\theta_t \sim \mu_t}[|\theta_t-\theta_*|_{P_t}|h_t] \leq
2\sqrt{C_0+\frac{1}{m^2}|P_t^{-\frac{1}{2}}\nabla_{\theta} U'_t(\theta_*)|^2 + \frac{dn+16}{m}}.
\]

For a random vector $X$ having a log-concave pdf,
 \cite[Theorem 5.22]{lovasz2003logconcave} yields that
\[
\mathbb{E}[|X|^p]^{\frac{1}{p}} \leq 2p \mathbb{E}[|X|]
\]
for any $p>0$.
We now observe that $y:=P_t^{\frac{1}{2}}(\theta_t-\theta_*)$ has a log-concave pdf since its potential $U_t({P_t}^{-\frac{1}{2}} y +\theta_*)$ is convex. 
Therefore, it follows that
\begin{align}
    \mathbb{E}_{\theta_t \sim \mu_t}[|\theta_t-\theta_*|^p_{P_t}|h_t]&\leq (2p)^p\mathbb{E}_{\theta_t \sim \mu_t}[|\theta_t-\theta_*|_{P_t}|h_t]^p \notag \\
    &\leq (2p)^p\bigg(4C_0+\frac{4}{m^2}|P_t^{-\frac{1}{2}}\nabla_{\theta} U'_t(\theta_*)|^2 + \frac{4dn+64}{m}\bigg)^{\frac{p}{2}} \label{half_1}.
\end{align}

Let $Z:=\begin{bmatrix} z_1& \cdots &z_{t-1} \end{bmatrix}^\top$. 
Then,  $\frac{\partial U'_t(\theta_*)}{\partial \Theta_{ij}} =- \sum_{s=1}^{t-1} Z_{si} \frac{\partial \log p_w(w_s)}{\partial w_s(j)}$, where the $j$th component of $w_s$ is denoted by $w_s(j)$. 
Therefore, $P_t$ can be written as $P_t= \lambda I_{dn}+\mathrm{blkdiag}\{Z^\top Z\}_{i=1}^n = I_n \otimes (Z^{\top}Z + \lambda I_d)$, and it is straightforward to check that $P_t^{-1} = I_n \otimes (Z^{\top}Z + \lambda I_d)^{-1} $.
Letting $\theta_\ell:= \Theta_{ij}$ for $\ell=(j-1)d+i$, 
we deduce that
\begin{equation*}
\begin{split}
    |P_t^{-\frac{1}{2}}\nabla_{\theta} U'_t(\theta_*)|^{2}&=\sum_{\ell,k=1}^{dn} \frac{\partial U'_t(\theta_*)}{\partial \theta_{\ell}} (P_t^{-1})_{\ell k} \frac{\partial U'_t(\theta_*)}{\partial \theta_{k}}\\
    & = \sum_{i',i=1}^d\sum_{j',j=1}^{n} \frac{\partial U'_t(\theta_*)}{\partial {\Theta_{i'j'}}} (P^{-1}_t)_{(j'-1)d+i',(j-1)d+i} \frac{\partial U'_t(\theta_*)}{\partial {\Theta_{ij}}}\\
    &= \sum_{j=1}^{n}\sum_{s',s=1}^{t-1}\frac{\partial \log p_w(w_{s'})}{\partial w_{s'}(j)}(Z(Z^{\top}Z + \lambda I_d)^{-1}Z^{\top})_{s's}\frac{\partial \log p_w(w_s)}{\partial w_s(j)}.
\end{split}
\end{equation*}

We are now ready to leverage the self-normalization technique, Lemma~\ref{lem:self_norm} in Section~\ref{app:lem:self_norm}. 
For a fixed $j$, we let $X_s=z_s$ and $V_t =\lambda I_d +\sum_{s=1}^{t-1} z_s z_s^\top$, $S_t = \sum_{s=1}^{t-1}\frac{\partial\log p_w(w_{s})}{\partial w_{s}(j)}z_s$ and take the probability bound $\delta$ as $\frac{\delta}{n}$ in the statement of the lemma.
Consequently, the inequality
\begin{equation*}
    \sum_{s,s'=1}^{t-1}\frac{\partial \log p_w(w_{s'})}{\partial w_{s'}(j)}(Z(Z^{\top}Z + \lambda I_d)^{-1}Z^{\top})_{s's}\frac{\partial \log p_w(w_s)}{\partial w_s(j)} \leq 2\frac{M^2}{m}\log\bigg(\frac{n}{\delta}\bigg(\frac{\sqrt[n]{\det(P_t)}}{\det(\lambda I_{dn})}\bigg)^{\frac{1}{2}}\bigg)
\end{equation*}
holds with probability at least 1-$\frac{\delta}{n}$ for each $j$.
Combining these for all $j=1, \ldots, n$ with \eqref{half_1}, we conclude that
\begin{align*}
    \mathbb{E}_{\theta_t \sim \mu_t}[|\theta_t-\theta_*|^p_{P_t}|h_t]&\leq (2p)^p\bigg( \bigg(\sum_{j=1}^n \frac{8M^2}{m^3}\log\bigg(\frac{n}{\delta}\bigg(\frac{\sqrt[n]{\det(P_t)}}{\det(\lambda I_d)}\bigg)^{\frac{1}{2}}\bigg)\bigg) + \frac{4dn+64}{m}+ 4C_0\bigg)^{\frac{p}{2}}\\
    &\leq (2p)^p\bigg( 8\frac{nM^2}{m^3}\log \bigg(\frac{n}{\delta}\bigg(\frac{\lambda_{\max,t}}{\lambda}\bigg)^{\frac{d}{2}}\bigg) +C\bigg)^{\frac{p}{2}}
\end{align*}
holds with probability no less than $1-\delta$ for some positive constant $C$ depending only on $m,n,d$ and $\lambda$, as desired. 
\end{proof}

\subsection{Proof of Theorem~\ref{thm:state_poly}}\label{app:polystate}

Before proving Theorem~\ref{thm:state_poly},
we introduce some auxiliary results on the behavior of $M_t := \Theta_* - \tilde \Theta_t \in \mathbb{R}^{d \times n}$, where $\tilde \Theta_t$ is a matrix whose vectorization is $\tilde \theta_t \in \mathbb{R}^{dn}$.
One of the fundamental ideas is to identify critical columns of $M_t$ representing the column space of $M_t$.
We follow the argument presented in  \cite[Appendix D]{abbasi2011regret}. 
For $\mathcal{B} \subset \mathbb{R}^d$ and $v\in\mathbb{R}^d$, let $\pi(v,\mathcal{B})$ denote the projection of the vector $v$ onto the space $\mathcal{B}$.
Similarly, we let $\pi(M,\mathcal{B})$ denote the column-wise projection of $M$ onto $\mathcal{B}$.
We then construct a sequence of subspaces $\mathcal{B}_t$ for $t=T,\ldots,1$ in the following way.
Let $\mathcal{B}_{T+1} =\emptyset$.
For step $t$, we begin by setting $\mathcal{B}_{t}=\mathcal{B}_{t+1}$.
Given $\epsilon > 0$, 
while $|\pi(M_t,\mathcal{B}_t^\perp)|_F>d\epsilon$,\footnote{Here, $|\cdot|_F$ denotes the Frobenius norm} we pick a column $v$ from $M_t$ satisfying $\pi(v,\mathcal{B}_t^\perp)>\epsilon$ and update $\mathcal{B}_t\leftarrow\mathcal{B}_t\oplus \{v\}$.
Thus, for each step $t$, we have
\begin{equation}\label{perpend}
|\pi(M_t,\mathcal{B}_t^\perp)|\leq |\pi(M_t,\mathcal{B}_t^\perp)|_F\leq d\epsilon.
\end{equation}

\begin{defn}
Let $\mathcal{T}_T= \{ t_1, \ldots, t_m\}$, $t_1>t_2>...>t_m$, be the set of timesteps at which subspaces $\mathcal{B}_t$ expand. Clearly, $|\mathcal{T}_T| \leq n$ since $M_t$ has $n$ columns. We also let $i(t):=\max\{i\leq |\mathcal{T}_T|:t_i \geq t\}$.
\end{defn}

A key insight of this procedure is to discover a sequence of subspaces $\mathcal{B}_t$ supporting $M_t$'s.
In this way, we derive the following bounds for the projection of any vector $x$ onto $\mathcal{B}_t$~\cite[Lemma 17]{abbasi2011regret}:
\begin{equation}\label{lem:17}
U\epsilon^{2d} |\pi(x,\mathcal{B}_t)|^2 \leq \sum_{j=1}^{i(t)}|M_{t_j}^\top x|^2,
\end{equation}
where $U = \frac{U_0}{H}$ with 
$U_0 =\frac{1}{16^{d-2}\max\{1,S^{2(d-2)}\}}$.
Here, $H$ is chosen to be a positive number strictly larger than $\max\{16,\frac{4S^2\tilde M^2}{dU_0}\}$, where $\bar{L} = \frac{1}{\sqrt{2m}}$ and $\tilde M$ is defined as
\begin{align*}
\tilde M&=\sup_Y e(T(T+1))^{-1/\log{\delta}}\\
&\quad\quad\times\bigg(10\sqrt{\frac{dn}{m}\log{\bigg(\frac{1}{\delta}\bigg)}}+ 2\log\bigg(\frac{1}{\delta}\bigg) \sqrt{\frac{8M^2n}{m^3}\log \bigg(\frac{nT(T+1)}{\delta}\bigg(d+\frac{TY^2}{\lambda}\bigg)^{\frac{d}{2}}\bigg)+C}\bigg)/Y.
\end{align*}

As mentioned in Section~\ref{sec:polystate},
we decompose an event into a good set and a bad set. 
Let  $\Omega$ denote the probability space representing all randomness incurred from the noise and the preconditioned ULA.
Given $0<\delta < 1$ in Proposition~\ref{prop:half_exact_true}, 
we define the events $E_t$ and $F_t$ as
\begin{align*}
E_t &= \{w \in \Omega: |\tilde\theta_s-\theta_*|_{P_s} \leq \beta_s(\delta)  \; \forall s \leq t\},\\
F_t &= \bigg\{w  \in \Omega:   |x_{s}| \leq \alpha_s,\,\max_{j \leq s}|\nu_j| \leq d \bar L_{\nu}\sqrt{2\log\bigg(\frac{2s^2(s+1)}{\delta}\bigg)} \; \forall s \leq t\bigg\},
\end{align*}
where 
\[ \beta_s(\delta) := e(s(s+1))^{-\frac{1}{\log \delta}}\left[10\sqrt{\frac{dn}{m} \log \bigg (\frac{1}{\delta}\bigg )}+ 2 \log \bigg (\frac{1}{\delta}\bigg ) \sqrt{\frac{8M^2n}{m^3}\log \bigg(\frac{n s (s+1)}{\delta}\bigg(\frac{\lambda_{\max,s}}{\lambda}\bigg)^{\frac{d}{2}}\bigg)+ C} \right ]\] 
with the constant $C$ from Proposition~\ref{prop:half_exact_true},
and 
\[\alpha_s := \frac{1}{1-\rho}\bigg(\frac{M_{\rho}}{\rho}\bigg)^d\left[{G}\Big (\max_{j\leq s}|z_j| \Big)^{\frac{d}{d+1}}\beta_s(\delta)^{\frac{1}{2(d+1)}}+ d(\bar L+S\bar L_{\nu})\sqrt{2\log\bigg(\frac{2s^2(s+1)}{\delta}\bigg)}\right]\]
with the constants $S$,
$\rho$ and $M_\rho$ defined in the beginning of Section~\ref{sec:alg}.\footnote{For any $\theta \in \mathcal{C}$, $|\theta| \leq S$, 
$|A + BK(\theta)| \leq \rho <1$ and $|A_* + B_* K(\theta) | \leq M_\rho$.}
Here, $\bar{L} = \frac{1}{\sqrt{2m}}$ and $\bar{L}_\nu$ is defined in Assumption~\ref{ass:perturb}, and $G=\big(H^{-1/(d+1)}+H^{d/(d+1)}\big)\big(\frac{2Sd^{d+0.5}}{\sqrt{U}}\big)^{\frac{1}{d+1}}$. Here, we should notice that when $w \in E_t$, $\tilde \theta_s \in \mathcal {C}$ for $s\leq t-1$ while $\tilde \theta_t$ follows approximate posterior distribution without restriction to $\mathcal{C}$.


We first show that the event $F_t$ occurs with high probability. 
This result allows us to integrate the OFU-based approach into our Bayesian setting for Thompson sampling. 

\begin{prop}\label{prop:hp}
Suppose Assumptions~\ref{ass:noise}--\ref{ass:perturb} hold. Then, for any $t\geq 1$ and any $\delta>0$ such that $log(\frac{1}{\delta})\geq 2$, we have
\[
\mathrm{Pr}( E_t \cap F_t) \geq 1-4\delta.
\]
\end{prop}
\begin{proof}
Given $1\leq t \leq T$, 
fix an arbitrary time step $s$ such that $1 \leq s \leq t$. 
By Proposition \ref{prop:half_exact_true},
\[
\mathbb{E}_{\theta_s \sim \mu_s}\big[
| \theta_s - \theta_* |^p_{P_s} \mid h_{s}
\big ]^{\frac{1}{p}} \leq 2p \sqrt{\frac{8M^2n}{m^3}\log \bigg(\frac{ns(s+1)}{\delta}\bigg(\frac{\lambda_{\max,s}}{\lambda}\bigg)^{\frac{d}{2}}\bigg)+ C}
\]
holds with probability no less than $1-\frac{\delta}{s(s+1)}$.
It follows from Proposition \ref{prop:ula} and the Minkowski inequality that  for any $p \geq 2$,
\begin{equation*}
\begin{split}
    \mathbb{E}_{\tilde \theta_s \sim \tilde \mu_s}\big[
| \tilde\theta_s - \theta_* |^p_{P_s} \mid h_{s}
\big ]^{\frac{1}{p}} & \leq \mathbb{E}_{\theta_s \sim \mu_s, \tilde{\theta}_s \sim \tilde{\mu}_s}\big[
| \tilde\theta_s - \theta_s |^p_{P_s} \mid h_{s}
\big ]^{\frac{1}{p}}+ \mathbb{E}_{\theta_s \sim \mu_s}\big[
| \theta_s - \theta_* |^p_{P_s} \mid h_{s}
\big ]^{\frac{1}{p}}\\
&\leq 10\sqrt{\frac{pdn}{m}}+ 2p \sqrt{\frac{8M^2n}{m^3}\log \bigg(\frac{n s(s+1)}{\delta}\bigg(\frac{\lambda_{\max,s}}{\lambda}\bigg)^{\frac{d}{2}}\bigg)+C}
\end{split}
\end{equation*}
with probability at least  $1-\frac{\delta}{s(s+1)}$.
By the Markov inequality, we observe that for any $\epsilon>0$
\begin{equation*}
\begin{split}
    \mathrm{Pr}(|\tilde{\theta}_s-\theta_*|_{P_s}>\epsilon \mid h_{s}) &\leq \frac{\mathbb{E}_{\tilde \theta \sim \tilde \mu_s}\big[
    | \tilde\theta - \theta_* |^p_{P_s} \mid h_{s}
    \big ]}{\epsilon^p}\\
    &\leq \frac{1}{\epsilon^p}\left(10\sqrt{\frac{pdn}{m}}+ 2p \sqrt{\frac{8M^2n}{m^3}\log \bigg(\frac{ns(s+1)}{\delta}\bigg(\frac{\lambda_{\max,s}}{\lambda}\bigg)^{\frac{d}{2}}\bigg)+ C}\right)^p,
\end{split}
\end{equation*}
where the second inequality holds with probability no less than $1-\frac{\delta}{s(s+1)}$.
We now set $p=\log (\frac{1}{\delta})$ and  

\[\epsilon = e (s(s+1))^{\frac{1}{p}}\Big(10\sqrt{\frac{pdn}{m}}+ 2p \sqrt{\frac{8M^2n}{m^3}\log \big(\frac{n s(s+1)}{\delta}\big(\frac{\lambda_{\max,s}}{\lambda}\big)^{\frac{d}{2}}\big)+C}\Big).\] 
Then, $\mathrm{Pr}(|\tilde{\theta}_s-\theta_*|_{P_s} \leq \beta_s(\delta) \mid h_{s})$
with probability at least $1-\frac{\delta}{s(s+1)}$.
This implies that 
\begin{align*}
\mathrm{Pr}(|\tilde{\theta}_s - \theta_*|_{P_s} \leq \beta_s(\delta)) &= \mathbb{E}\big [\mathbb{E}[\mathbf{1}_{|\tilde  \theta_s - \theta_*|_s \leq \beta_s(\delta)}|h_s]\big ]\\
&=\mathbb{E}\big [\mathrm{Pr}(|\tilde \theta_s - \theta_*|_s \leq \beta_s(\delta)|h_s)\big ] \\
& \geq \bigg (1-\frac{\delta}{s(s+1)}\bigg )^2
\geq 1-\frac{2\delta}{s(s+1)}.
\end{align*}
Let $\Lambda_s := \{w\in \Omega_s \subset \Omega:|\tilde \theta_s-\theta_*|_{P_s}\leq \beta_s(\delta)\}$ where $\Omega_s$ denotes the set of all events before time $s$. 
Thus, 
$\mathrm{Pr} (\Lambda_s^c) \leq \frac{2\delta}{s(s+1)}$. 
Thus, we have
\begin{align*}
  \mathrm{Pr} (E_t) =   \mathrm{Pr} \bigg(\bigcap_{s=1}^{t} \Lambda_s \bigg )=1-\mathrm{Pr}\bigg (\bigcup_{s=1}^{t} \Lambda_s^c \bigg ) \geq 1-\sum_{s=1}^{t} \mathrm{Pr}(\Lambda_s^c)\geq 1- 2\delta.
\end{align*}

For $i \leq s$, we rewrite the linear system~\eqref{sys} as
\[
x_{i+1} = \Gamma_i x_i +r_i,
\]
where
\[
\Gamma_i =
\begin{cases}
\tilde{\Theta}_i^{\top}\tilde{K}(\Tilde{\theta}_i) & \mbox{if }i \notin  \mathcal{T}_s, \\
\Theta_*^{\top}\tilde{K}(\Tilde{\theta}_i) &  \mbox{if }i \in  \mathcal{T}_s
\end{cases}
\]
with $\Tilde{K}(\theta)^{\top} = \begin{bmatrix}
        I_n & {K(\theta)}^{\top} \end{bmatrix}$,
and
\[
r_i = 
\begin{cases}
(\Theta_* - \tilde{\Theta}_i)^{\top}z_i + B_*\nu_i+ w_i &\mbox{if } i \notin {\mathcal{T}_s}, \\
B_*\nu_i+ w_i &\mbox{if } i \in {\mathcal{T}_s.}
\end{cases}
\]
The system state at time $i$ can then be expressed as
\begin{equation*}
\begin{split}
    x_{s} &= \Gamma_{s-1}x_{s-1}+r_{s-1} \\
    &= \Gamma_{s-1}(\Gamma_{s-2}x_{s-2}+r_{s-2})+r_{s-1} \\
    &= \Gamma_{s-1}\Gamma_{s-2}x_{s-2} + \Gamma_{s-1}r_{s-2} + r_{s-1}\\
    & = \Gamma_{s-1}\Gamma_{s-2}\Gamma_{s-3}x_{s-3} +\Gamma_{s-1}\Gamma_{s-2}r_{s-3}+ \Gamma_{s-1}r_{s-2} + r_{s-1}\\
    & = \Gamma_{s-1}\Gamma_{s-2}\ldots\Gamma_2 r_1 + \cdots +\Gamma_{s-1}\Gamma_{s-2}r_{s-3}+\Gamma_{s-1}r_{s-2} + r_{s-1}\\
    & = \sum_{j=1}^{s-2}\bigg(\prod_{i=j+1}^{s-1}\Gamma_i\bigg)r_j + r_{s-1}.
\end{split}
\end{equation*}
Recall that $|\tilde{\Theta}_i^{\top}\tilde{K}(\Tilde{\theta}_i)|\leq \rho<1$ and $|\Theta_*^{\top}\tilde{K}(\Tilde{\theta}_i)| \leq M_{\rho}$ thanks to the construction of our algorithm.
Since $|\mathcal{T}_s|\leq d$,
we have
\[
\prod_{i=j+1}^{s-1}|\Gamma_i| \leq M_{\rho}^{d}\rho^{s-d-j-1},
\]
which implies that 
\[
|x_s| = \bigg(\frac{M_{\rho}}{\rho}\bigg)^d\sum_{j=1}^{s-2}\rho^{s-j-1}|r_{j}|+|r_{s-1}| \leq \frac{1}{1-\rho}\bigg(\frac{M_{\rho}}{\rho}\bigg)^d \max_{j\leq s}|r_j|.
\]
By the definition of $r_j$, we have
\begin{align*}
    \max_{j \leq s}|r_j| &\leq \max_{j \leq s, j \notin \mathcal{T}_s}|(\tilde{\Theta}_j-\Theta_*)^{\top}z_j| + S\max_{j \leq s}|\nu_j| +\max_{j \leq s}|w_j|.
\end{align*}
It follows from   Lemma \ref{lem:abbasi_lem_18} that
\begin{equation*}
    \max_{j\leq s, j\notin \mathcal{T}_s}|(\tilde{\Theta}_j-\Theta_*)^{\top}z_j| \leq G\Big (\max_{j\leq s}|z_j| \Big)^{\frac{d}{d+1}}\beta_{s}(\delta)^{\frac{1}{2(d+1)}}
\end{equation*}
with probability no less than $1-2\delta$ since $\mathrm{Pr}(E_s)\geq \mathrm{Pr}(E_t) \geq 1-2\delta$.

Note that our system noise is an $\bar L$-sub-Gaussian random vector, where $\bar L=\frac{1}{\sqrt{2m}}$. By Herbst's argument in \cite{ledoux1999concentration}, we have
\begin{equation}\label{eq:noise_w}
\max_{j \leq s}|w_j| \leq d \bar L\sqrt{2\log\bigg(\frac{2s^2(s+1)}{\delta}\bigg)}
\end{equation}
with probability no less than $1-\frac{\delta}{s(s+1)}$. 
Similarly, since $\nu_j$ is an $\bar L_{\nu}$-sub-Gaussian random vector, 
\begin{equation}\label{eq:noise_nu}
\max_{j \leq s}|\nu_j| \leq d \bar L_{\nu}\sqrt{2\log\bigg(\frac{2s^2(s+1)}{\delta}\bigg)}
\end{equation}
with probability no less than $1-\frac{\delta}{s(s+1)}$.
Let $\hat{E}_{w,s} \subset E_s$ and $\hat{E}_{\nu,s}\subset E_s$  denote the events satisfying~\eqref{eq:noise_w} and~\eqref{eq:noise_nu}, respectively. 
Then, on the event $\hat{E}_{w,s} \cap \hat{E}_{\nu,s}$, we obtain that
\[
|x_s|\leq \frac{1}{1-\rho}\bigg(\frac{M_{\rho}}{\rho}\bigg)^d\left(G\Big(\max_{j< s}|z_j|\Big )^{\frac{d}{d+1}}\beta_{s}(\delta)^{\frac{1}{2(d+1)}}+d (\bar L+S\bar L_{\nu})\sqrt{2\log\bigg(\frac{2s^2(s+1)}{\delta}\bigg)}\right)=\alpha_s.
\]
Hence, for $\hat{\Lambda}_t := \bigcap_{s=1}^t (\hat{E}_{w,s} \cap \hat{E}_{\nu,s})$, we have
\[
\hat{\Lambda}_t \cap E_t \subset F_t.
\]
By the union bound argument, 
\[
\mathrm{Pr} (\hat{\Lambda}_t \cap E_t) \geq 1 - \mathrm{Pr} \bigg (\bigcup_{s=1}^t (\hat{E}_{w, s}^c \cup \hat{E}_{\nu, s}^c ) \bigg ) - \mathrm{Pr} (E_t^c) \geq 1-4\delta,
\]
where the last inequality follows from 
$\mathrm{Pr} (\hat{E}_{w, s}^c )\leq \frac{\delta}{s(s+1)}$, $\mathrm{Pr} (\hat{E}_{\nu, s}^c )\leq \frac{\delta}{s(s+1)}$ and $\mathrm{Pr} (E_t^c) \leq 2\delta$. 
Consequently, we obtain that
\[
\mathrm{Pr} (E_t \cap F_t) \geq \mathrm{Pr} (\hat \Lambda_t \cap E_t \cap F_t) = \mathrm{Pr} (\hat \Lambda_t \cap E_t) \geq 1-4\delta.
\]
\end{proof}

It  immediately follows from Proposition~\ref{prop:hp}  that $\mathrm{Pr} (F_t^c) \leq 4\delta$. 
Using this property, we now prove Theorem~\ref{thm:state_poly}.

\begin{proof}[Proof of Theorem~\ref{thm:state_poly}]
We first decompose $\mathbb{E}[\max_{j\leq t}|x_j|^p]$ as
 \begin{equation}\label{state_divide}
    \mathbb{E} \Big [\max_{j\leq t}|x_j|^p \Big ] = \mathbb{E}\Big [\max_{j\leq t}|x_j|^p\mathbf{1}_{F_t} \Big ] +\mathbb{E}\Big [\max_{j\leq t}|x_j|^p\mathbf{1}_{F_t^c} \Big ].
\end{equation}
It follows from the Cauchy-Schwartz inequality
and
Proposition~\ref{prop:hp} that
\begin{align*}
    \mathbb{E}\Big [\max_{j\leq t}|x_j|^p\mathbf{1}_{F_t^c}\Big ] &\leq {\mathbb{E}[\mathbf{1}_{F_t^c}]}^{\frac{1}{2}} {\mathbb{E}\Big [\max_{j\leq t}|x_j|^{2p}\Big ]}^{\frac{1}{2}}  \leq {(4\delta)}^{\frac{1}{2}} {\mathbb{E}\Big [\max_{j\leq t}|x_j|^{2p} \Big]}^{\frac{1}{2}}.
\end{align*}
Let $D_t = \Theta_*^{\top}\tilde{K}(\Tilde{\theta}_t)$ and $r_t = B_*\nu_t + w_t$. Then, the linear system can be expressed as
\begin{align*}
    x_t &= D_{t-1}x_{t-1}+r_{t-1}  = D_{t-1}(D_{t-2}x_{t-2}+r_{t-2})+ r_{t-1}\\
    & = D_{t-1}D_{t-2}D_{t-3}x_{t-3} + D_{t-1}D_{t-2}r_{t-3}+ D_{t-1}r_{t-2}+r_{t-1}\\
    & = D_{t-1}D_{t-2}\ldots D_2r_1 + \cdots + D_{t-1}D_{t-2}r_{t-3}+ D_{t-1}r_{t-2} + r_{t-1}\\
    & = \sum_{j=1}^{t-2}\bigg(\prod_{s=j+1}^{t-1}D_s\bigg)r_j +r_{t-1}.
\end{align*}
Since $|D_t| \leq M_{\rho}$, we have
\begin{align*}
    \mathbb{E}\big [|x_t|^{2p}\big ] &= \mathbb{E}\left [\bigg |\sum_{j=1}^{t-2}\bigg(\prod_{s=j+1}^{t-1}D_s\bigg)r_j +r_{t-1} \bigg |^{2p} \right ]\\
    & \leq (t-1)^{2p-1}\mathbb{E} \left [\sum_{j=1}^{t-2}\bigg |\bigg(\prod_{s=j+1}^{t-1}D_s\bigg)r_j \bigg|^{2p}+|r_{t-1} |^{2p} \right ]\\
    & \leq (t-1)^{2p-1}\mathbb{E}\left [\sum_{j=1}^{t-1}M_{\rho}^{2p(t-j-1)}|r_j|^{2p} \right]\\
    & \leq (t-1)^{2p}\mathbb{E}\big [|r_t|^{2p}\big ]M_{\rho}^{2p(t-2)},
\end{align*}
where the second inequality follows from Jensen's inequality.

By Lemma~\ref{lem:state_log} with $\delta = \frac{1}{t^{2p+1} M_\rho^{2pt}} \leq \frac{1}{t}$, the first term on the right-hand side of \eqref{state_divide} is estimated as
\begin{align*}
    \mathbb{E}\Big [\max_{j \leq t}|x_j|^p\mathbf{1}_{F_t}\Big ] &\leq \mathbb{E}\left [C\left(\log\bigg(\frac{1}{\delta}\bigg)^3\sqrt{\log\bigg(\frac{t}{\delta}\bigg)}\right)^{p(d+1)}\mathbf{1}_{F_t}\right]\\
    &\leq C\left(\log\bigg(\frac{1}{\delta}\bigg)^3\sqrt{\log\bigg(\frac{t}{\delta}\bigg)}\right)^{p(d+1)}
\end{align*}
for some positive constant $C$ depending only on $n,n_u,\rho,M_{\rho},S,\bar{L}_{\nu}, m$ and $M$.

Finally, we obtain that
\begin{align*}
    \mathbb{E}\Big [\max_{j\leq t}|x_j|^{p}\Big ]
    &\leq C\left(\log\bigg(\frac{1}{\delta}\bigg)^3\sqrt{\log\bigg(\frac{t}{\delta}\bigg)}\right)^{p(d+1)}+\sqrt{4\delta}\sqrt{\mathbb{E}\Big[\max_{j\leq t}|x_j|^{2p}\Big]}\\
    &\leq C\left(\log\bigg(\frac{1}{\delta}\bigg)^3\sqrt{\log\bigg(\frac{t}{\delta}\bigg)}\right)^{p(d+1)}+\sqrt{4\delta}\sqrt{\sum_{j=1}^t\mathbb{E}[|x_j|^{2p}]}\\
    &\leq C\left (\log\big(t^{2p+1}M_{\rho}^{2pt}\big)^3\sqrt{\log\bigg(t^{2p+2}M_{\rho}^{2pt}\bigg)}\right)^{p(d+1)} + 2\sqrt{\mathbb{E}[|r_t|^{2p}]}\\
    & \leq C t^{\frac{7}{2}p(d+1)} + 2\sqrt{\mathbb{E}[|r_t|^{2p}]}.
\end{align*}
It follows from Jensen's inequality that 
\begin{equation}\nonumber
\begin{split}
\mathbb{E}[|r_t|^{2p}] &\leq  2^{p-1}(S^{2p}\mathbb{E}[|\nu_t|^{2p}] + \mathbb{E}[|w_t|^{2p}])\\
&\leq 2^{p-1}p!\bigg (S^{2p}(4\bar L_{\nu}^2)^p+\Big(\frac{2}{m}\Big )^p\bigg ),
\end{split}    
\end{equation}
where the second inequality holds because $\nu_t$ and $w_t$ are sub-Gaussian. 
Putting everything together, the result follows. 
\end{proof}

\subsection{Proof of Proposition~\ref{prop:pers-ex}}\label{app:prop:pers-ex}

\begin{proof}
Given $ j \in [1,k]$, let $A_*, B_*$ be the true system parameters and $s \in (t_j,t_{j+1}):=\mathcal{I}_j$. We first define the following quantities for $s\in \mathcal{I}_j$ :
\[
y_s := 
\begin{bmatrix}
    A_*x_{s-1} + B_*u_{s-1}          \\
    K_j(A_*x_{s-1} + B_*u_{s-1})
\end{bmatrix}, 
\]
where $K_j$ denotes the control gain matrix computed at the beginning of $j$th episode.

Writing
\[
L_s := 
\begin{bmatrix}
    I_n & 0\\
    K_j & I_{n_u}
\end{bmatrix},
\quad\text{and}\quad \psi_s :=
\begin{bmatrix}
    w_{s-1}\\
    \nu_{s}
\end{bmatrix},
\]
we can decompose $z_s$ as $z_s= y_s + L_s\psi_s$ by the construction of the algorithm.

For a trajectory $(z_s)_{s\geq 1}$, let us introduce a sequence of random variables up to time $s$, which is denoted by
\[
\tilde h_s:=(x_1,W_1,\nu_1,...,x_{s},W_s,\nu_s),
\]
where $W_s$ denotes randomness incurred by the ULA when triggered, hence, $W_s=0$ if $s \neq t_j$ for some $j$.
Defining the index set 
\[
\mathcal{J}_k:=\{s \in \mathcal{I}_j: j \in[1,k]\},
\]
we consider the modified filtration 
\begin{equation*}
\mathcal{F}'_s:=
\begin{cases}
\sigma(\cup_{j \leq s} \tilde h_j) \quad\text{for}\quad s \in \mathcal{J}_k-\{t_2-1,t_3-1,...,t_k-1\},\\
\sigma(\cup_{j \leq s+1} \tilde h_j) \quad\text{for}\quad s\in \{t_2-1,t_3-1,...,t_k-1\}.
\end{cases}
\end{equation*}
This way we can incorporate the information observed at $s=t_j$ with that made up to $s=t_j-1$ as seen in Figure~\ref{fig:app_e}.

\begin{figure}[h]
    \centering
    \begin{tikzpicture}[scale = 1.2]
    \tikzset{block/.style= {draw, rectangle, align=center,minimum width=1cm,minimum height=1cm},
            }
    \draw ; \node[] at (-1,-0.15) {$\ldots$};
    \draw ; \node[] at (0,-0.15) {$\mathcal{F}'_{t_j-1}$};
    \draw ; \node[] at (1,-0.15) {$||$};
    \draw ; \node[] at (2,-0.15) {$\mathcal{F}'_{t_j+1}$};
    \draw ; \node[] at (3,-0.15) {$\ldots$};
    \draw ; \node[] at (4,-0.15) {$\mathcal{F}'_{s-1}$};
    \draw ; \node[] at (5,-0.15) {$\mathcal{F}'_{s}$};
    \draw ; \node[] at (6,-0.15) {$\ldots$};
    \draw ; \node[] at (7,-0.15) {$\mathcal{F}'_{t_{j+1}-1}$};
    \draw ; \node[] at (8,-0.15) {$||$};
    \draw ; \node[] at (9,-0.15) {$\mathcal{F}'_{t_{j+1}+1}$};
    \draw ; \node[] at (10,-0.15) {$\ldots$};
    
    \draw[->] (1.55, -22 pt)node[block,right]{$y_{t_j+1}$\\$L_{t_j+1}$} -- (0,-10pt); 
    \draw[->] (4.55, -22 pt)node[block,right]{$y_{s}$\\$L_{s}$} -- (4,-10pt); 
    \end{tikzpicture}
    \caption{Filtration and  measurability of $(y_s)$ and $(L_s)$.}
    \label{fig:app_e}
\end{figure}

Yet simple but important observation is that for $\mathcal{J}_k=\{n_i:n_1<n_2<...<n_{\frac{k(k+1)}{2}}\}$ 
both stochastic processes $(L_{n_s})$, $(y_{n_s})$ are $\mathcal{F}'_{n_{s-1}}$-measurable and $(\psi_{n_s})$ is $\mathcal{F}'_{n_s}$-measurable.

To proceed we first notice that
\[
{\lambda_{\min}} \bigg(\lambda I_d+ \sum_{s=1}^{t_{k+1}-1} z_s z_s^\top\bigg) \succeq    {\lambda_{\min}} \bigg(\lambda I_d+\sum_{s \in\mathcal{J}_k} z_s z_s^\top\bigg).
\]
Invoking Lemma \ref{lem:preconditioner_expansion} with $\epsilon=\tilde\lambda=1$ and $\xi_s=L_s \psi_s$, it follows that
\begin{equation}\label{pe}
\begin{split}
&\sum_{s \in\mathcal{J}_k} z_s z_s^{\top}\quad\succeq\\
& \sum_{s \in\mathcal{J}_k}  (L_s\psi_s)(L_s\psi_s)^{\top} - \underbrace{\bigg[\sum_{s \in\mathcal{J}_k}  y_s (L_s \psi_s)^{\top}\bigg]^{\top}\bigg [ I_d+\sum_{s \in\mathcal{J}_k} y_sy_s^{\top}\bigg ]^{-1} \bigg [\sum_{s \in\mathcal{J}_k}  y_s (L_s \psi_s)^{\top}\bigg ]}_{(*)} -  I_d.
\end{split}
\end{equation}
Our goal is to find a lower bound of \eqref{pe}. To begin with, define $
\psi_{1,s} = 
\begin{bmatrix}
    w_{s-1}\\
    0
\end{bmatrix}$ and $
\psi_{2,s} = 
\begin{bmatrix}
    0\\
    \nu_{s}
\end{bmatrix}$ for $s\geq 1$ setting $w_0=0$ for simplicity. Noting that $L_s\psi_s = L_s\psi_{1,s} + \psi_{2,s}$, we apply Lemma \ref{lem:preconditioner_expansion} with $\epsilon=\frac{1}{2}$, $\tilde\lambda=1$ to obtain
\begin{equation}\label{pe_aux}
\begin{split}
    &\sum_{s \in\mathcal{J}_k}  (L_s\psi_{s})(L_s\psi_{s})^{\top}
    \succeq\sum_{s \in\mathcal{J}_k}  (L_s\psi_{1,s})(L_s\psi_{1,s})^{\top} + \frac{1}{2}\sum_{s \in\mathcal{J}_k} \psi_{2,s}\psi_{2,s}^{\top}\\
    & - 2\underbrace{\bigg [\sum_{s \in\mathcal{J}_k}  \psi_{2,s} (L_s \psi_{1,s})^{\top}\bigg]^{\top}\bigg[ I_d+\sum_{s \in\mathcal{J}_k} \psi_{2,s}\psi_{2,s}^{\top}\bigg]^{-1}\bigg[\sum_{s \in\mathcal{J}_k}  \psi_{2,s} (L_s \psi_{1,s})^{\top}\bigg]}_{(**)} - \frac{1}{2} I_d.
\end{split}
\end{equation}
The first term of \eqref{pe_aux} is written as
\begin{align*}
    \sum_{s \in\mathcal{J}_k}  (L_s\psi_{1,s})(L_s\psi_{1,s})^{\top} & = \sum_{s \in\mathcal{J}_k}
    \begin{bmatrix}
        w_{s-1} w_{s-1}^{\top} & w_{s-1}(K_{v(s)}w_{s-1})^{\top}\\
        (K_{v(s)}w_{s-1}) w_{s-1}^{\top} & (K_{v(s)}w_{s-1})(K_{v(s)}w_{s-1})^{\top}     
    \end{bmatrix}
    \\
    & =: 
    \begin{bmatrix}
        X^{\top}X & X^{\top}Y\\
        Y^{\top}X & Y^{\top}Y
    \end{bmatrix},
\end{align*}
where $v(s)$ is indicates the episode number such that $s \in \mathcal{I}_{v(s)}$. By Lemma \ref{lem:tall_matrix}, we conclude that
\begin{align}\label{bar-lambda}
    \sum_{s \in\mathcal{J}_k}  (L_s\psi_{1,s})(L_s\psi_{1,s})^{\top} 
     = 
    \begin{bmatrix}
        X^{\top}X & X^{\top}Y\\
        Y^{\top}X & Y^{\top}Y
    \end{bmatrix}
     \succeq 
    \begin{bmatrix}
        \frac{\bar\lambda}{|Y|^2+\bar\lambda}X^{\top}X & 0\\
        0 & -\bar\lambda I_{n_u}
    \end{bmatrix}
\end{align}
for any $\bar\lambda>0$, where $X = [w_{n_1-1}, \cdots, w_{n_{k(k+1)/2}-1}]^{\top}$ and $Y = [K_{v(n_1)}w_{n_1-1}, \cdots, K_{v(n_{k(k+1)/2})}w_{n_{k(k+1)/2}-1}]^{\top}$.

Next, we invoke Lemma \ref{lem:noise_tail_bound} with $\epsilon=\frac{1}{2}\lambda_{\min}(\mathbf{W})$ for $\psi_s = w_{s-1}$, $\psi_s = \nu_{s}$ respectively to characterize good noise sets. 
Choosing $\rho = \log \frac{2}{\delta}$ in Lemma \ref{lem:noise_tail_bound}, there exists $C>0$ such that for any $\delta>0$ and $k\geq C \sqrt{\log (\frac{2}{\delta}) +d\log 9}$, the following events hold with probability at least $1-\delta$:
\begin{align*}
E_{1,k} &= \bigg\{w \in \Omega : \frac{1}{4}\lambda_{\min}(\mathbf{W}) k(k+1) I_{n} \preceq \sum_{s \in\mathcal{J}_k}  w_{s-1} w_{s-1}^{\top} \preceq \frac{1}{2}(\lambda_{\max}(\mathbf{W}) +\frac{1}{2}\lambda_{\min}(\mathbf{W})) k(k+1) I_{n}\bigg\},\\
E_{2,k} &= \bigg\{\nu \in \Omega_{\nu} : \frac{1}{4}\lambda_{\min}(\mathbf{W})k(k+1) I_{n_u} \preceq \sum_{s \in\mathcal{J}_k}  \nu_{s} \nu_{s}^{\top} \preceq \frac{1}{2}(\lambda_{\max}(\mathbf{W}) +\frac{1}{2}\lambda_{\min}(\mathbf{W}))k(k+1) I_{n_u} \bigg\},
\end{align*}
where $\Omega_\nu \subset \Omega$ denotes the probability space associated with the random sequence $(\nu_s)_{s\geq 1}$ and $\Omega$ is the probability space representing all randomness in the algorithm as defined in the previous subsection. Furthermore, from the observation,
\begin{equation*}
\begin{split}
\mathrm{tr}\bigg(\sum_{s \in\mathcal{J}_k} (K_{v(s)} w_{s-1}) (K_{v(s)}w_{s-1})^\top\bigg) &\leq \sum_{s \in\mathcal{J}_k} \mathrm{tr} ((K_{v(s)} w_{s-1}) (K_{v(s)}w_{s-1})^\top)\\
&\leq M_K^2 \sum_{s \in\mathcal{J}_k} |w_{s-1}|^2 \\
&= M_K^2 \mathrm{tr}\bigg(\sum_{s \in\mathcal{J}_k} w_{s-1}w_{s-1}^\top\bigg),
\end{split}
\end{equation*}
we also have the following event {whose subevent is $E_{1,k}$}:
\[E_{3,k} =\bigg \{w \in \Omega :  \sum_{s \in\mathcal{J}_k}  (K_{v(s)} w_{s-1})(K_{v(s)} w_{s-1})^{\top} \preceq \frac{n M_K^2}{2}(\lambda_{\max}(\mathbf{W}) +\frac{1}{2}\lambda_{\min}(\mathbf{W})) k(k+1) I_{n_u}\bigg\}.\]

To proceed we choose $\bar{\lambda} = \frac{1}{8}\lambda_{\min}(\mathbf{W})k$ in \eqref{bar-lambda} and recall that $|Y|^2 = \lambda _{\max} (Y^\top Y)$. On the event $E_{1,k} \cap E_{2,k} \cap E_{3,k}$, first two terms on the right-hand side of \eqref{pe_aux} is lower bounded as
\begin{align*}
    &\sum_{s \in\mathcal{J}_k}  (L_s\psi_{1,s})(L_s\psi_{1,s})^{\top}+\frac{1}{2}\sum_{s \in\mathcal{J}_k}  \psi_{2,s}\psi_{2,s}^{\top}\\
    &\succeq
    \begin{bmatrix}
    \frac{\bar\lambda}{|Y|^2+\bar\lambda}X^{\top}X & 0\\
    0 & -\bar\lambda I_{n_u}
    \end{bmatrix} + \frac{1}{2}
   \sum_{s\in\mathcal{J}_k} 
   \begin{bmatrix}0 \\ \nu_s  
   \end{bmatrix}
    \begin{bmatrix}
        0 &\nu_s^\top
    \end{bmatrix}\\
    &\succeq
    \begin{bmatrix}
    \frac{\frac{1}{32}\lambda_{\min}^2(\mathbf{W})k^2(k+1)}{\frac{nM_K^2}{2}(\lambda_{\max}(\mathbf{W})+\frac{1}{2}\lambda_{\min}(\mathbf{W}))k(k+1)+\frac{1}{8}\lambda_{\min}(\mathbf{W})k} I_{n} &0\\ 
        0 & -\frac{1}{8}\lambda_{\min}(\mathbf{W}) k I_{n_u}
    \end{bmatrix}
    +
    \begin{bmatrix}
        0 & 0\\
        0 & \frac{1}{4}\lambda_{\min}(\mathbf{W})k(k+1)I_{n_u}
    \end{bmatrix}
    \\
    &= 
    k\begin{bmatrix}
    \frac{\lambda_{\min}^2(\mathbf{W})k(k+1)}{16nM_K^2(\lambda_{\max}(\mathbf{W})+\frac{1}{2}\lambda_{\min}(\mathbf{W}))k(k+1)+4\lambda_{\min}(\mathbf{W})k} I_{n}&0\\ 
        0 & \frac{1}{8}\lambda_{\min}(\mathbf{W})(2k+1)I_{n_u}
    \end{bmatrix}
    \\
    &\succeq C k I_{d}
\end{align*}
for some $C>0$. 

We next deal with $(*)$ in \eqref{pe} and $(**)$ in \eqref{pe_aux} together as they have the same  structure. Let us begin by defining 
\[
S_k(\psi_{2},L\psi_{1}) := \bigg[\sum_{s \in\mathcal{J}_k}  \psi_{2,s} (L_s \psi_{1,s})^{\top}\bigg]^{\top}\bigg[I_d+\sum_{s \in\mathcal{J}_k} \psi_{2,s}\psi_{2,s}^{\top}\bigg]^{-1}\bigg[\sum_{s \in\mathcal{J}_k}  \psi_{2,s} (L_s \psi_{1,s})^{\top}\bigg].
\]
Similarly,
\[
S_k(y,L\psi) := \bigg[\sum_{s \in\mathcal{J}_k}  y_s (L_s \psi_{s})^{\top}\bigg]^{\top}\bigg[I_d+\sum_{s \in\mathcal{J}_k} y_s y_s^{\top}\bigg]^{-1}\bigg[\sum_{s \in\mathcal{J}_k}  y_s (L_s \psi_{s})^{\top}\bigg].
\]

Applying Lemma \ref{lem:self-normalized_matrix} with $\rho=\log(\frac{1}{\delta})$ to the stochastic processes $(\psi_s)_{s\in \mathcal{J}_k}$ and $(y_s)_{s\in \mathcal{J}_k}$, each of the following events holds with probability at least $1-\delta$:
\begin{align*}
E_{4,k} &=\bigg \{w \in \Omega, \nu \in \Omega_{\nu}:|S_k(\psi_{2},L\psi_{1})|\leq 7\bar L^2
(M_K^2+2)\log\bigg(\frac{e^d \det(I_d+\sum_{s \in\mathcal{J}_k}\psi_{2,s}\psi_{2,s}^{\top})}{\delta}\bigg)\bigg\},\\
E_{5,k} &=\bigg \{w \in \Omega, \nu \in \Omega_{\nu}:|S_k(y,L\psi)|\leq 7\max\{\bar L, \bar L_\nu\}^2(M_K^2+2)\log\bigg(\frac{e^d \det(I_d+\sum_{s \in\mathcal{J}_k} y_s y_s^{\top})}{\delta}\bigg)\bigg\},
\end{align*}
since $\max_{s \leq t} |L_s| \leq \sqrt{M_K^2+2}$ 
with $L_s := 
\begin{bmatrix}
    I_n & 0\\
    K_j & I_{n_u}
\end{bmatrix}.$ To verify, we recall that $|L_s| = \sqrt{\lambda_{\max}(L_sL_s^{\top})}$. Here,
\[
L_sL_s^{\top} =
\begin{bmatrix}
    I_{n} &K_j^{\top}\\ 
        K_j & K_jK_j^{\top}+I_{n_u}
    \end{bmatrix}.
\]
Fixing $v=\begin{bmatrix}x^{\top} & y^{\top}\end{bmatrix}^{\top}$ such that $|v|=1$ where $x\in\mathbb{R}^n$ and  $y\in\mathbb{R}^{n_u}$, we have
\begin{align*}
    v^{\top}
    \begin{bmatrix}
    I_{n} &K_j^{\top}\\ 
        K_j & K_jK_j^{\top}+I_{n_u}
    \end{bmatrix}
    v
    &\leq  
    |x|^2+2x^{\top}K_j^{\top}y + M_K^2|y|^2 + |y|^2\\
    &\leq (M_K^2+1)(x^2+y^2)+|y|^2\\
    &\leq M_K^2+2.
\end{align*}
\begin{itemize}
\item Bound of $S_k(\psi_2,L\psi_1)$ on $E_{2,k}\cap E_{4,k}$:

On $E_{2,k}$,
\begin{align*}
\det\bigg(I_d+\sum_{s \in\mathcal{J}_k}\psi_{2,s}\psi_{2,s}^{\top}\bigg)^{\frac{1}{d}} &\leq \frac{1}{d}(d+\sum_{s \in\mathcal{J}_k}\psi_{2,s}^{\top}\psi_{2,s})\\
& = \frac{1}{d}(d+\sum_{s \in\mathcal{J}_k}|\nu_{s}|^2)\\
& \leq \frac{n_u}{2d}(\lambda_{\max}(\mathbf{W}) +\frac{1}{2}\lambda_{\min}(\mathbf{W}))k(k+1)+1\\
& \leq C k^2
\end{align*}
for some $C>0$ where the second inequality follows by
\begin{align*}
\sum_{s\in\mathcal{J}} |\nu_s|^2 =\mathrm{tr}(\sum_{s\in\mathcal{J}_k} \nu_s \nu_s ^\top) &\leq n_u \lambda_{\max}({\sum_{s\in\mathcal{J}_k} \nu_s\nu_s^\top})\\
& \leq \frac{n_u}{2}(\lambda_{\max}(\mathbf{W}) +\frac{1}{2}\lambda_{\min}(\mathbf{W}))k(k+1).
\end{align*}

Altogether, on the event $E_{2,k}\cap E_{4,k}$,
\begin{equation*}
\begin{split}
S_k(\psi_{2},L\psi_{1})&=\bigg|\bigg[\sum_{s \in\mathcal{J}_k} \psi_{2,s} (L_s \psi_{1,s})^{\top}\bigg]^{\top}\bigg[I_d+\sum_{s \in\mathcal{J}_k}\psi_{2,s}\psi_{2,s}^{\top}\bigg]^{-1}\bigg[\sum_{s \in\mathcal{J}_k} \psi_{2,s} (L_s \psi_{1,s})^{\top}\bigg]\bigg|\\
&\leq 7\bar L^2(M_K^2+2)\log\bigg(\frac{Ce^d k^{2d}}{\delta}\bigg).
\end{split}
\end{equation*}

\item Bound of $S_k(y,L\psi)$ on $F_{t_{k+1}} \cap E_{1,k} \cap E_{5,k}$: 

On $E_{1,k}$, 
\begin{align*}
&\det\bigg(I_d+\sum_{s \in\mathcal{J}_k} y_s y_s^{\top}\bigg)^{\frac{1}{d}} \\
&\leq \frac{1}{d}\bigg(d+\sum_{s \in\mathcal{J}_k}|y_s|^2\bigg)\\
& = \frac{1}{d}\bigg(d+\sum_{s \in\mathcal{J}_k}(\underbrace{|x_s-w_{s-1}|^2}_{\leq 2|x_s|^2+2|w_{s-1}|^2}+\underbrace{|K_{v(s)}(x_s-w_{s-1})|^2}_{\leq 2M_K^2|x_s|^2+2M_K^2|w_{s-1}|^2})\bigg)\\
& \leq \frac{1}{d}\bigg(d+\sum_{s \in\mathcal{J}_k}((2+2M_K^2)|x_s|^2+(2+2M_K^2)|w_{s-1}|^2)\bigg)\\
& \leq  \frac{(M_K^2+1)}{d}\bigg(\underbrace{2\sum_{s \in\mathcal{J}_k}|x_s|^2}_\text{(a)}+ \underbrace{n(\lambda_{\max}(\mathbf{W}) +\frac{1}{2}\lambda_{\min}(\mathbf{W}))k(k+1)}_\text{by taking trace in $E_{1,k}$}\bigg)+1,
\end{align*}
where the last inequality follows from
\begin{align*}
\sum_{s\in\mathcal{J}} |w_{s-1}|^2 =\mathrm{tr}\bigg(\sum_{s\in\mathcal{J}_k} w_{s-1} w_{s-1} ^\top\bigg) &\leq n \lambda_{\max}\bigg({\sum_{s\in\mathcal{J}_k} w_{s-1}w_{s-1}^\top}\bigg)\\
& \leq \frac{n}{2}\bigg(\lambda_{\max}(\mathbf{W}) +\frac{1}{2}\lambda_{\min}(\mathbf{W})\bigg)k(k+1).
\end{align*}

To bound (a) above, let us observe that $t_{k+1}=\frac{(k+1)(k+2)}{2} \leq 3k^p$ for any $p\geq 3$ and consider the event $F_{t_{k+1}} \cap E_{1,k}$. Applying Lemma~\ref{lem:state_log} with $\delta=k^{-p} \leq t_{k+1}^{-1}$, we deduce that
\begin{align*}
   \sum_{s\in\mathcal{J}_k} |x_s|^2=\sum_{s\in\mathcal{J}_k}|x_s|^2 & \leq t_{k+1} \max_{s \leq t_{k+1}}|x_s|^2\\
   &\leq t_{k+1} \bigg( C(\log k)^3 \sqrt{\log k} \bigg)^{2(d+1)}\\
   &\leq C k^2\bigg( k\sqrt{\log k}\bigg)^{2(d+1)}\\
   & \leq C k^{3d+5}
\end{align*}
for some $C>0$ depending on $p\geq 3$ and the constant from Lemma~\ref{lem:state_log}.

Therefore, on the event $F_{t_{k+1}} \cap E_{1,k}\cap E_{5,k}$, we have
\begin{align*}
    \det\bigg(I_d+\sum_{s\in\mathcal{J}_k} y_s y_s^{\top}\bigg)^{\frac{1}{d}}&\leq (M_K^2+1)\bigg(\frac{2C}{d}k^{3d+5}+ \bigg(\lambda_{\max}(\mathbf{W}) +\frac{1}{2}\lambda_{\min}(\mathbf{W})\bigg)k(k-1)\bigg)+1\\
    &\leq C k^{3d+5}
\end{align*}
for some constant $C>0$. As a result,
\begin{equation*}
\begin{split}
S_k(y,L\psi)&=\bigg|\bigg[\sum_{s\in\mathcal{J}_k} y_s (L_s \psi_{s})^{\top}\bigg]^{\top}\bigg[I_d+\sum_{s\in\mathcal{J}_k}y_sy_s^{\top}\bigg]^{-1}\bigg[\sum_{s\in\mathcal{J}_k} y_s (L_s \psi_{s})^{\top}\bigg]\bigg|\\
&\leq 7\max\{\bar L, \bar L_\nu\}^2(M_K^2+2)\log\bigg(\frac{Ce^d k^{d(3d+5)}}{\delta}\bigg).
\end{split}
\end{equation*}

\end{itemize}

Combining altogether and plugging them into \eqref{pe}, on the event $F_{t_{k+1}} \cap E_{1,k} \cap E_{2,k}\cap E_{3,k}\cap E_{4,k}\cap E_{5,k}$, one can derive that
\begin{equation*}
\begin{split}
    \lambda_{\min}(\lambda I_d + \sum_{s\in \mathcal{J}_k} z_s z_s^{\top}) 
    &\geq  \lambda+ C_1 k - C_2\log k  + C_3 \log(\delta)-C_4\\
    & \geq C k
\end{split}
\end{equation*}
for some $C_i,C>0$ with $\delta=k^{-p}$ and $k \geq k_0$ for $k_0$ large enough. In turn, we have the concentration bound for the excitation yielding that 
\begin{align*}
    &Pr\bigg(\lambda_{\min}(\lambda I_d + \sum_{s=1}^{t_{k+1}-1} z_s z_s^{\top}) \geq C k\bigg)\\
    & \geq 1-Pr(F^c_{t_{k+1}} \cup E^c_{1,k} \cup E^c_{2,k}\cup E^c_{3,k}\cup E^c_{4,k}\cup E^c_{5,k})\\
    & \geq 1-9\delta.
\end{align*}
Finally, defining the event $\bar{F}_{k+1}:=F_{t_{k+1}}\cap E_{1,k} \cap E_{2,k}\cap E_{3,k}\cap E_{4,k}\cap E_{5,k}$,
\begin{align*}
    \mathbb{E}\bigg[\frac{1}{\lambda^p_{\min,k+1}}\bigg] 
    &= \mathbb{E}\bigg[\frac{1}{\lambda^p_{\min,k+1}}\mathbbm{1}_{\bar{F}_{k+1}}\bigg] + \mathbb{E}\bigg[\frac{1}{\lambda^p_{\min,k+1}}\mathbbm{1}_{\bar{F}_{k+1}^c}\bigg]\nonumber\\
    &\leq C\mathbb{E}\bigg[ k^{-p}\mathbbm{1}_{\bar{F}_{k+1}}\bigg] + \mathbb{E}\bigg[\mathbbm{1}_{\bar{F}_{k+1}^c}\bigg]\nonumber\\
    &\leq C k^{-p} + 9\delta \leq C k^{-p},
\end{align*}
where second inequality holds from $\lambda_{\min,t} \geq \lambda \geq 1$. 
\end{proof}

\subsection{Proof of Theorem~\ref{thm:exact_appox}}
\label{app:thm:exact_appox}
\begin{proof}
It follows from \eqref{half_1} in Proposition \ref{prop:half_exact_true}  that 
\[
\mathbb{E}_{\theta_t \sim \mu_{t}}[|\theta_t-\theta_*|^p_{P_{t}}|h_{t}]\leq (2p)^p\bigg(\frac{4}{m^2}|P_{t}^{-\frac{1}{2}}\nabla_{\theta} U'_{t}(\theta_*)|^2 + \frac{4dn}{m}+ 64m+C\bigg)^{\frac{p}{2}},
\]
where $U_t'(\theta) =\sum_{s=1}^{t-1}\log p_w(x_{s+1}-\Theta^{\top}z_s)$.
Recalling $\lambda_{\min,t}=\lambda_{\min,t}(P_t)$, it follows that
\[
\lambda^{\frac{p}{2}}_{\min,t}\mathbb{E}[|\theta_t -\theta_*|^p]\leq  \mathbb{E}[|\theta_t-\theta_*|^p_{P_{t}}],
\] and hence,
\begin{align}\label{fe}
    &\mathbb{E}[\mathbb{E}_{\theta_t \sim \mu_t}[|\theta_t-\theta_*|^p|h_{t}]] \nonumber \\
    & \leq (2p)^p \sqrt{\mathbb{E}\bigg[\frac{1}{\lambda_{\min,t}^{p}}\bigg]}\sqrt{\mathbb{E}\bigg[\bigg(\frac{4}{m^2}|P_{t}^{-\frac{1}{2}}\nabla_{\theta} U'_{t}(\theta_*)|^2+  \frac{4dn}{m}+ 64m+C\bigg)^p\bigg]}\nonumber\\
    & \leq (2p)^p \sqrt{\mathbb{E}\bigg[\frac{1}{\lambda_{\min,t}^{p}}\bigg]}\sqrt{2^{p-1}\bigg(\frac{4^p}{m^{2p}}\mathbb{E}\bigg[|P_{t}^{-\frac{1}{2}}\nabla_{\theta} U'_{t}(\theta_*)|^{2p}\bigg] + \bigg( \frac{4dn}{m}+ 64m+C\bigg)^p\bigg)},
\end{align}
where the second inequality holds by Jensen's inequality and the outer expectation is taken with respect to the history at time $t$.

To bound $\mathbb{E}\bigg[|P_{t}^{-\frac{1}{2}}\nabla_{\theta} U'_{t}(\theta_*)|^{2p}\bigg]$, let us first define $Z:=\begin{bmatrix} z_1& \cdots &z_{t-1} \end{bmatrix}^\top$ and denote the $j$th component of noise $w_t$ by $w_t(j)$. A naive bound is achieved as
\begin{align}\label{navie:bd}
    |P_{t}^{-\frac{1}{2}}\nabla_{\theta} U'_{t}(\theta_*)|^{2}&= \sum_{j=1}^{n}\sum_{s',s=1}^{t-1}\frac{\partial \log p_w(w_{s'})}{\partial w_{s'}(j)}(Z(Z^{\top}Z + \lambda I_d)^{-1}Z^{\top})_{s's}\frac{\partial \log p_w(w_s)}{\partial w_s(j)}\notag\\
    &\leq \sum_{j=1}^{n}\sum_{s',s=1}^{t-1}\frac{\partial \log p_w(w_{s'})}{\partial w_{s'}(j)}(Z(Z^{\top}Z)^{-1}Z^{\top})_{s's}\frac{\partial \log p_w(w_s)}{\partial w_s(j)}\notag\\
    &\leq \sum_{j=1}^{n}\sum_{s=1}^{t-1}\bigg(\frac{\partial \log p_w(w_s)}{\partial w_s(j)}\bigg)^2 \notag\\
    &= \sum_{s=1}^{t-1}|\nabla_w \log p_w(w_s)|^2,
\end{align}
where the second inequality follows from the fact that $Z(Z^{\top}Z)^{-1}Z^{\top}$ is a projection matrix.

We now claim that $\mathbb{E}\bigg[|P_{t}^{-\frac{1}{2}}\nabla_{\theta} U'_{t}(\theta_*)|^{2p}\bigg]$ has a better bound compared to the naive one with high probability leveraging self-normalized bound for vector-valued martingale. For $s\geq 0$, let us consider the natural filtration 
\[
\mathcal{F}_s = \sigma((z_1,...,z_{s+1})),
\]
where $z_s=(x_s,u_s)$. Clearly, for $s\geq 1$, $z_s$ is $\mathcal{F}_{s-1}$-measurable and the random vector $\nabla_w \log p_w(w_s)$ is $\mathcal{F}_s$-measurable. Then for each $j \in [1,n]$, we set $\eta_s=\frac{\partial\log p_w(w_{s})}{\partial w_{s}(j)}$, $X_s=z_s$, $S_t=\sum_{s=1}^{t-1} \eta_s X_s=\sum_{s=1}^{t-1}\frac{\partial\log p_w(w_{s})}{\partial w_{s}(j)}z_s$. Here, $\eta_s$ is a $\frac{M}{\sqrt{m}}$-sub-Gaussian random variable since $v^\top \nabla_w \log p_w (w_t)$ is $\frac{M}{\sqrt{m}}$-sub-Gaussian random variable for any $v\in\mathbb{R}^n$ given when $w_t$ is sub-Gaussian (Proposition 2.18 in \cite{ledoux2001concentration}). Together with the fact that
\[
\lambda I_d + \sum_{s=1}^{t-1} X_s X_s^\top = \lambda I_d+Z^\top Z,
\]
and the result for self-normalized bound~\ref{lem:self_norm},
\begin{align*}
    &(\sum _{s=1}^{t-1} \eta_s X_s)^\top(\lambda I_d + \sum_{s=1}^{t-1} X_s X_s^\top)^{-1}(\sum _{s=1}^{t-1} \eta_s X_s)\\
    &=\sum_{s,s'=1}^{t-1}\frac{\partial \log p_w(w_{s'})}{\partial w_{s'}(j)}(Z(Z^{\top}Z + \lambda I_d)^{-1}Z^{\top})_{s's}\frac{\partial \log p_w(w_s)}{\partial w_s(j)}\notag\\
    &\leq 2\frac{M^2}{m}\log\bigg(\frac{n}{\delta}\bigg(\frac{\sqrt[n]{\det(P_{t})}}{\det(\lambda I_d)}\bigg)^{\frac{1}{2}}\bigg),
\end{align*}
holds with probability at least $1-\frac{\delta}{n}$.
Note that in the last inequality, we used the fact that $\det(\lambda I_d+Z^\top Z)=\sqrt[n]{\det(\lambda I_{dn}+\sum_{s=1}^{t-1} \mathrm{blkdiag}\{z_s z_s^\top\}_{i=1}^n)}=\sqrt[n]{\det(P_{t})}$.

By the union bound argument,
\begin{align}\label{tight:bd}
    |P_{t}^{-\frac{1}{2}}\nabla_{\theta} U'_{t}(\theta_*)|^{2} &= \sum_{j=1}^n\sum_{s,s'=1}^{t-1}\frac{\partial \log p_w(w_{s'})}{\partial w_{s'}(j)}(Z(Z^{\top}Z + \lambda I_d)^{-1}Z^{\top})_{s's}\frac{\partial \log p_w(w_s)}{\partial w_s(j)}\notag\\ 
    &\leq 2\frac{nM^2}{m}\log\bigg(\frac{n}{\delta}\bigg(\frac{\sqrt[n]{\det(P_{t})}}{\det(\lambda I_d)}\bigg)^{\frac{1}{2}}\bigg),
\end{align}
with probability at least $1- \delta$ for any $\delta>0$. Let us denote this event as $\tilde E$ so that $Pr(\tilde E) \geq 1-\delta$.

Combining the naive bound~\eqref{navie:bd} and improved bound~\eqref{tight:bd}, 
\begin{align}\label{se}
      &\mathbb{E}\bigg[|P_{t}^{-\frac{1}{2}}\nabla_{\theta} U'_{t}(\theta_*)|^{2p}\bigg]\nonumber\\ &=\mathbb{E}\bigg[\mathbbm{1}_{\tilde E}|P_{t}^{-\frac{1}{2}}\nabla_{\theta} U'_{t}(\theta_*)|^{2p}\bigg] + \mathbb{E}\bigg[\mathbbm{1}_{\tilde E^c}|P_{t}^{-\frac{1}{2}}\nabla_{\theta} U'_{t}(\theta_*)|^{2p}\bigg]\nonumber\\
      &\leq \underbrace{\mathbb{E}\bigg[\bigg(2\frac{nM^2}{m}\log\bigg(\frac{n}{\delta}\bigg(\frac{\sqrt[n]{\det(P_{t})}}{\det(\lambda I_d)}\bigg)^{\frac{1}{2}}\bigg)\bigg)^p\bigg]}_\text{by \eqref{tight:bd}} + \sqrt{\mathbb{E}\bigg[\mathbbm{1}_{\tilde E^c}\bigg]}\sqrt{\mathbb{E}\bigg[|P_{t}^{-\frac{1}{2}}\nabla_{\theta} U'_{t}(\theta_*)|^{4p}\bigg]}\nonumber\\
      &\leq \mathbb{E}\bigg[\bigg(2\frac{nM^2}{m}\log\bigg(\frac{n}{\delta}\bigg(\frac{\lambda_{\max,t}}{\lambda}\bigg)^{\frac{d}{2}}\bigg)\bigg)^p\bigg]+\sqrt{\delta}\underbrace{\sqrt{\mathbb{E}\bigg[\bigg(\sum_{s=1}^{t-1}|\nabla_w \log p_w(w_s)|^2\bigg)^{2p}\bigg]}}_\text{by \eqref{navie:bd}}.
\end{align}
We handle two terms on the right hand side separately. Recall that $g:x\rightarrow (\log x)^p$ is concave on $x\geq\max\{1,e^{p-1}\}$ whenever $p >0$. By Jensen's inequality, the first term is bounded as
\begin{equation*}
\begin{split}
    \mathbb{E}\bigg[\bigg(2\frac{nM^2}{m}\log\bigg(\frac{n}{\delta}\bigg(\frac{\lambda_{\max,t}}{\lambda}\bigg)^{\frac{d}{2}}\bigg)\bigg)^p\bigg]&=\mathbb{E}\bigg[\bigg(\frac{dnM^2}{m}\log\bigg(\left(\frac{n}{\delta}\right)^{2/d}\frac{\lambda_{\max,t}}{\lambda}\bigg)\bigg)^p\bigg]\\
    &\leq \bigg(\frac{dnM^2}{m}\bigg)^p\log\bigg(\frac{n}{\lambda\delta}\mathbb{E}[\lambda_{\max,t}]\bigg)^p\\
    &\leq \bigg(\frac{dnM^2}{m}\bigg)^p\log\bigg(\frac{n}{\lambda\delta}\mathbb{E}[\frac{1}{n}\mathrm{tr}(P_{t})]\bigg)^p\\
    &\leq \bigg(\frac{dnM^2}{m}\bigg)^p\log\bigg(\frac{n}{\lambda\delta}\mathbb{E}[d\lambda + \sum_{s=1}^{t-1}|z_s|^2]\bigg)^p\\
    &\leq \bigg(\frac{dnM^2}{m}\bigg)^p\log\bigg(\frac{n}{\lambda\delta}\bigg(d\lambda+M_K^2t\bigg(\mathbb{E}[\max_{j\leq t-1}|x_j|^2]+\mathrm{tr}(\mathbf{W}')\bigg)\bigg)\bigg)^p\\
    &\leq \bigg(\frac{dnM^2}{m}\bigg)^p\log\bigg(\frac{n}{\lambda\delta}\bigg(d\lambda+CM_K^2{t}^{7d+8}\bigg)\bigg)^p, 
\end{split}
\end{equation*}
where the last inequality holds from the Theorem~\ref{thm:state_poly}.

On the other hand, the second term of \eqref{se} can be handled similarly. Recalling Jensen's inequality,
\[
\bigg(\frac{\sum_{i}^n a_i}{n}\bigg )^{2p} \leq \frac{\sum_{i=1}^n a_i^{2p}}{n}
\]
for $a_i \in \mathbb{R}$ and $p \geq 1$, we have that
\begin{equation*}
\begin{split}
\sqrt{\delta}\sqrt{\mathbb{E}\bigg[\bigg(\sum_{s=1}^{t-1}|\nabla_w \log p_w(w_s)|^2\bigg)^{2p}\bigg]}&\leq \sqrt{\delta}\sqrt{t^{2p-1}\mathbb{E}\bigg[\sum_{s=1}^{t-1}|\nabla_w \log p_w(w_s)|^{4p}\bigg]}\\
&\leq \sqrt{\delta}t^{p}\sqrt{\mathbb{E}\bigg[|\nabla_w \log p_w(w_t)|^{4p}\bigg]}\\
&\leq \sqrt{\delta}t^{p}\sqrt{\bigg(\frac{4M^2}{m}\bigg)^{2p}(2p)!}\\
&\leq 8^p\frac{M^{2p}}{m^{p}}p^p\sqrt{\delta}t^{p},
\end{split}
\end{equation*}
where the third inequality comes from well-known fact that any $\bar L$-sub-Gaussian random vector $X$ satisfies $\E[X^{2q}] \leq q!(4{\bar L}^2)^q$ for any $q>0$.

Choosing $\delta = \frac{1}{t^{2p}}$ and combining two bounds,
\begin{align*}
    \mathbb{E}\bigg[|P_{t}^{-\frac{1}{2}}\nabla_{\theta} U'_{t}(\theta_*)|^{2p}\bigg]&\leq  \bigg(\frac{dnM^2}{m}\bigg)^p\log\bigg(\frac{n}{\lambda\delta}\bigg(d\lambda+CM_K^2{t}^{7d+8}\bigg)\bigg)^p+8^p\frac{M^{2p}}{m^{p}}p^p\sqrt{\delta}t^{p}\\
    &\leq\bigg(\frac{dnM^2}{m}\bigg)^p\log\bigg(nt^{2p}\bigg(d+\frac{CM_K^2}{\lambda}t^{7d+8}\bigg)\bigg)^p+8^p\frac{M^{2p}}{m^{p}}p^p.
\end{align*}

Finally, going back to \eqref{fe},
\begin{equation*}
\begin{split}
    &\mathbb{E}[\mathbb{E}_{\theta_t \sim \mu_{t}}[|\theta_t-\theta_*|^p|h_{t}]] \\
    & \leq (2p)^p \sqrt{\mathbb{E}\bigg[\frac{1}{\lambda_{\min,t}^{p}}\bigg]}\sqrt{2^{p-1}\bigg(\frac{4^p}{m^{2p}}\mathbb{E}\bigg[|P_{t}^{-\frac{1}{2}}\nabla_{\theta} U'_{t}(\theta_*)|^{2p}\bigg] + \bigg( \frac{4dn}{m}+ 64m+C\bigg)^p\bigg)}\\
    &\leq (2p)^p \sqrt{\mathbb{E}\bigg[\frac{1}{\lambda_{\min,t}^{p}}\bigg]}\\
    &\quad\quad \times\sqrt{\frac{2^{3p-1}(dn)^pM^{2p}}{m^{3p}}\log\bigg(nt^{2p}\bigg(d+\frac{CM_K^2}{\lambda}t^{7d+8}\bigg)\bigg)^p+ \frac{2^{6p-1}}{m^{3p}}M^{2p}p^p+\bigg( \frac{4dn}{m}+ 64m+C\bigg)^p}\\
    &\leq \bigg((2p)^pC\sqrt{\frac{2^{3p-1}(dn)^pM^{2p}}{m^{3p}}\log\bigg(nt^{2p}\bigg(d+\frac{CM_K^2}{\lambda}t^{7d+8}\bigg)\bigg)^p+ \frac{2^{6p-1}}{m^{3p}}M^{2p}p^p+\bigg( \frac{4dn}{m}+ 64m+C\bigg)^p}\bigg)t^{-\frac{p}{4}},
\end{split}
\end{equation*}
where last inequality holds thanks to Proposition \ref{prop:pers-ex}.

For the concentration of the approximate posterior, we invoke Jensen's inequality to derive
\begin{equation}\nonumber
\begin{split}
    \mathbb{E}\bigg[\mathbb{E}_{\tilde{\theta}_t \sim \tilde{\mu}_t}\big[| \tilde{\theta}_t - \theta_* |^p | h_{t}\big ]\bigg] &= \mathbb{E}\bigg[\mathbb{E}_{\theta_t \sim \mu_t, \tilde{\theta}_t \sim \tilde{\mu}_t}\big[| \tilde{\theta}_t - \theta_* |^p\big | h_{t}]\bigg]\\
    &\leq 2^{p-1}\mathbb{E}\bigg[\mathbb{E}_{\theta_t \sim \mu_t, \tilde{\theta}_t \sim \tilde{\mu}_t}\big[| \theta_t- \tilde{\theta}_t |^p | h_{t} \big ]\bigg]+2^{p-1}\mathbb{E}\bigg[\mathbb{E}_{\theta_t \sim \mu_t, \tilde{\theta}_t \sim \tilde{\mu}_t}\big[| \theta_t - \theta_* |^p|h_{t} \big ]\bigg]\\
    &\leq 2^{p-1}\mathbb{E}\bigg[\frac{D_p}{(\sqrt{\lambda_{\min,t}})^p}\bigg]+2^{p-1}C\bigg(t^{-\frac{1}{4}}\sqrt{\log t}\bigg)^p\\
    &\leq C\bigg(t^{-\frac{1}{4}}\sqrt{\log t}\bigg)^p,
\end{split}
\end{equation}
where the second inequality comes from Proposition \ref{prop:ula} and the concentration result of exact posterior above.
\end{proof}

\subsection{Proof of Theorem~\ref{thm:improved}}
\label{app:thm:improved}
\begin{proof}
At $k$th episode, for timestep $t \in [t_k,t_{k+1})$, $x_{t}$ is written as
\begin{equation}\label{state_norm}
    x_{t+1} = (A_*+B_*K(\tilde{\theta}_{t}))x_{t}+ r_t,
\end{equation}
where $r_t = B_*\nu_t +w_t$.
Squaring and taking expectations on both sides of the equation above with respect to noises, the prior and randomized actions,
\begin{equation}\label{sb2}
    \mathbb{E}[|x_{t+1}|^2] \leq \mathbb{E}[|D_{t}|^{2}|x_{t}|^2]+ \mathbb{E}[|r_{t}|^2],
\end{equation}
where $D_{t} = A_*+B_*K(\tilde{\theta}_{t})$.

Since $\theta_*$ is stabilizable, it is clear to see that there exists small $\epsilon_0>0$ for which $|\theta-\theta_*|\leq \epsilon_0$ implies that $|A_*+B_* K(\theta)|\leq \Delta<1$ for some $\Delta>0$. Splitting $\mathbb{E}[|D_t|^2|x_t|^2]$ around the true system parameter $\theta_*$,
\[
\mathbb{E}[|D_{t}|^2|x_{t}|^2] = \underbrace{\mathbb{E}[|D_{t}|^2|x_{t}|^2\mathbbm{1}_{|\tilde{\theta}_{t}-\theta_*| \leq \epsilon_0}]}_{(i)} +\underbrace{\mathbb{E}[|D_{t}|^2|x_{t}|^2\mathbbm{1}_{|\tilde{\theta}_{t}-\theta_*| > \epsilon_0}]}_{(ii)}. 
\]
One can see that (i) is bounded by $\Delta^2 \mathbb{E}[|x_{t}|^2]$ by the construction. For (ii), we note that $|D_{t}| \leq M_\rho$ by Assumption \ref{ass:init_bound}. Using Cauchy-Schwartz inequality, (ii) is bounded as
\begin{equation}\label{second_1}
\mathbb{E}[|D_{t}|^2|x_{t}|^2\mathbbm{1}_{|\tilde{\theta}_{t}-\theta_*| > \epsilon_0}]] \leq M_{\rho}^2\sqrt{Pr(|\tilde{\theta}_{t}-\theta_*|>\epsilon_0)}\sqrt{\mathbb{E}[|x_{t}|^4]}.
\end{equation}
By Markov's inequality,
\begin{align*}
    Pr(|\tilde{\theta}_{t}-\theta_*|>\epsilon_0) &\leq \frac{\mathbb{E}[|\tilde{\theta}_{t}-\theta_*|^p]}{\epsilon_0^p}\\
    &\leq C\bigg(t^{-\frac{1}{4}}\sqrt{\log t}\bigg)^p,
\end{align*}
where the last inequality holds for $t\geq t_0$ thanks to Theorem \ref{thm:exact_appox}, and $C$ is a positive constant depending only on $p$ and $\epsilon_0$. 
Taking $p$ large enough to satisfy $p>28(d+1)$, Theorem \ref{thm:state_poly} yields that 
\begin{align*}
M_{\rho}^2\sqrt{Pr(|\tilde{\theta}_{t}-\theta_*|>\epsilon_0)}\sqrt{\mathbb{E}[|x_{t}|^4]}
    &\leq M_{\rho}^2C \bigg(t^{-\frac{1}{4}}\sqrt{\log t}\bigg)^pt^{7(d+1)} <C
\end{align*}
for some $C>0$.

Therefore, $\mathbb{E}[|x_{t+1}|^2]$ is estimated as
\[
\mathbb{E}[|x_{t+1}|^2] \leq \Delta^2\mathbb{E}[|x_{t}|^2]+C+\mathbb{E}[|r_{t}|^2].
\]
As $r_t$ is sub-Gaussian, we also have $\mathbb{E}[|r_t|^2]$ is bounded, and hence, 
\[
\mathbb{E}[|x_{t}|^2] <C
\]
for all $t \in [1,T]$ and $C>0$ by the recursive relation.

To handle the fourth moment, we take the fourth power on both sides and expectation to \eqref{state_norm} to obtain
\begin{align*}
    &\mathbb{E}[|x_{t+1}|^4]\\
    &\leq \mathbb{E}[|D_{t}x_{t}|^4]+\underbrace{4\mathbb{E}[|D_{t}x_{t}|^2(D_{t}x_{t})^{\top}w_{t}]}_{=0}+6\mathbb{E}[|D_{t}x_{t}|^2|r_{t}|^2] +4\mathbb{E}[|D_{t}x_{t}||r_{t}|^3] + \mathbb{E}[|r_t|^4]\\
    &\leq [|D_{t}|^4|x_{t}|^4\mathbbm{1}_{|\tilde{\theta}_{t}-\theta_*| \leq \epsilon_0}]+\mathbb{E}[|D_{t}|^4|x_{t}|^4\mathbbm{1}_{|\tilde{\theta}_{t}-\theta_*| \geq \epsilon_0}]+\underbrace{6M_{\rho}^2\mathbb{E}[|r_{t}|^2]\mathbb{E}[|x_{t}|^2] +4M_{\rho}\mathbb{E}[|r_{t}|^3]\mathbb{E}[|x_{t}|]+  \mathbb{E}[|r_t|^4]}_{<C}\\
    & \leq \Delta^4\mathbb{E}[|x_{t}|^4]+M_{\rho}^4\sqrt{Pr(|\tilde{\theta}_{t}-\theta_*| \geq \epsilon_0)}\sqrt{\mathbb{E}[|x_{t}|^8]}+C,
\end{align*}
since $\mathbb{E}[|x_t|^2] \leq C$. We recall Theorem \ref{thm:state_poly} once again with $p$ satisfying $p>56(d+1)$ to deduces that
\begin{align*}
M_{\rho}^2\sqrt{Pr(|\tilde{\theta}_{t}-\theta_*|>\epsilon_0)}\sqrt{\mathbb{E}[|x_{t}|^8]}
    &\leq M_{\rho}^2C\bigg(t^{-\frac{1}{4}}\sqrt{\log t}\bigg)^pt^{14(d+1)} \leq C
\end{align*}
for some $C>0$. 

Hence,
\begin{align*}
    \mathbb{E}[|x_{t+1}|^4] \leq \Delta^4\mathbb{E}[|x_{t}|^4]+C,
\end{align*}
and, one can conclude that 
\[
\mathbb{E}[|x_{t}|^4] <C
\]
for some $C>0$.
\end{proof}

\subsection{Proof of Theorem~\ref{thm:regret}}\label{app:thm:regret}
It follows from \cite{abeille2017thompson} that $J$ is Lipschitz continuous on $\mathcal{C}$ with a Lipschitz constant $L_J > 0$. We then estimate one of the key components of regret.

\begin{lem}\label{lem:r1}
Suppose that Assumptions \ref{ass:noise}, \ref{ass:init_bound} and \ref{ass:perturb} hold.
Recall that $\bar{\Theta}_* \in \mathbb{R}^{d \times n}$ denote the 
matrix of the true parameter random variables, $\tilde{\Theta}_k  \in \mathbb{R}^{d \times n}$ is the matrix of the parameters sampled in episode $k$, and $z_t := (x_t, u_t) \in \mathbb{R}^d$.
Then, the following inequality holds:

\begin{equation}\nonumber
\begin{split}
&R_1:= \mathbb{E}\bigg[
\sum_{k=1}^{n_T} \sum_{t=t_k}^{t_{k+1} - 1} z_t^\top [
\bar{\Theta}_*  P^*_k  \bar{\Theta}_*^\top
- \tilde{\Theta}_k P^*_k \tilde{\Theta}_k^\top 
]z_t
\bigg ]\\
& \leq 8M_{P^*} S\sqrt{D}(C M_K^2+32\bar{L}_{\nu}^2) n_T,
\end{split}
\end{equation}
where $P^*_k := P^*(\tilde{\theta}_k)$ is the symmetric positive definite solution of the ARE~\eqref{ric} with $\theta := \tilde{\theta}_k$, and $n_T$ is the last episode for time horizon $T$.
\end{lem}

\begin{proof}[Proof of Lemma~\ref{lem:r1}]
We first observe that for any $\theta$ which satisfies $|\theta| \leq S$,
\[
|z_t|  = | (x_t, u_t) | = | (x_t, K(\theta) x_t+ \nu_t) |
= \left | \begin{bmatrix}
I_n\\
K(\theta)
\end{bmatrix} x_t + \begin{bmatrix}
    0 \\
    I_{n_u}
\end{bmatrix}\nu_t
\right | \leq M_K |x_t| + |\nu_t|,
\]
and
\[
| {P^*_k}^{1/2} \Theta^\top z_t | \leq M_{P^*}^{1/2} S |z_t|, 
\]
where $M_{P^*}$ satisfies $|P^*(\theta)| \leq M_{P^*}$ for all $\theta \in \mathcal{C}$.
We then consider 
\begin{equation}\label{eq:lem3:1}
\begin{split}
 |{P^*_k}^{1/2} \bar{\Theta}_*^\top z_t |^2 - |{P^*_k}^{1/2} \tilde{\Theta}_k^\top z_t |^2 &= (|{P^*_k}^{1/2} \bar{\Theta}_*^\top z_t | + |{P^*_k}^{1/2} \tilde{\Theta}_k^\top z_t |) (|{P^*_k}^{1/2} \bar{\Theta}_*^\top z_t | - |{P^*_k}^{1/2} \tilde{\Theta}_k^\top z_t |)\\
 &\leq (|{P^*_k}^{1/2} \bar{\Theta}_*^\top z_t | + |{P^*_k}^{1/2} \tilde{\Theta}_k^\top z_t |) | {P^*_k}^{1/2} (\bar{\Theta}_* - \tilde{\Theta}_k)^\top z_t |\\
 &\leq 2 M_{P^*}  S |z_t| |  (\bar{\Theta}_* - \tilde{\Theta}_k)^\top z_t |.
 \end{split}
\end{equation}
Note that 
\[
\Theta^\top z_t = \begin{bmatrix}  {\Theta}(1) & \cdots & {\Theta} (n)\end{bmatrix}^\top z_t \in \mathbb{R}^n.
\]
Thus, with $<x,y>$ denoting the inner product of two vectors $x,y\in \mathbb{R}^d$,
\begin{equation}\label{eq:lem3:2}
\begin{split}
|  (\bar{\Theta}_* - \tilde{\Theta}_k)^\top z_t |^2 &=  \sum_{i=1}^n | <(\bar{\Theta}_* - \tilde{\Theta}_k)(i), z_t>|^2\\
&\leq \sum_{i=1}^n| (\bar{\Theta}_* - \tilde{\Theta}_k) (i)|^2 |z_t|^2\\
&\leq |z_t|^2 \sum_{i=1}^n | (\bar{\Theta}_* - \tilde{\Theta}_k) (i)|^2\\
&= |z_t|^2 |\bar{\theta}_* - \tilde{\theta}_k|^2.
\end{split}
\end{equation}
Combining \eqref{eq:lem3:1} and \eqref{eq:lem3:2} yields that
\begin{equation}\label{eq:lem3:3}
\begin{split}
R_1&\leq 2 M_{P^*} S  \mathbb{E} \bigg [
 \sum_{k=1}^{n_T} \sum_{t=t_k}^{t_{k+1} -1} |z_t|^2|\bar{\theta}_* - \tilde{\theta}_k|
\bigg ] \\
&\leq 4 M_{P^*} S\bigg( M_K^2 \mathbb{E} \bigg [
 \sum_{k=1}^{n_T} \sum_{t=t_k}^{t_{k+1} -1} |x_t|^2|\bar{\theta}_* - \tilde{\theta}_k|
\bigg ] +   \mathbb{E} \bigg [
 \sum_{k=1}^{n_T} \sum_{t=t_k}^{t_{k+1} -1} |\nu_t|^2|\bar{\theta}_* - \tilde{\theta}_k|
\bigg ]\bigg).
\end{split}
\end{equation}
Invoking the Cauchy-Schwarz inequality, we have
\[
\mathbb{E} [|x_t|^2 |\bar{\theta}_* - \tilde{\theta}_k| ]
\leq \sqrt{\mathbb{E}[|x_t|^4] \mathbb{E}[|\bar{\theta}_* - \tilde{\theta}_k|^2 ]}.
\]
It follows from the tower rule together with Proposition~\ref{prop:ula} that
\[
\sqrt{\mathbb{E}[|\bar{\theta}_* - \tilde{\theta}_k|^2 ]}
= \sqrt{ \mathbb{E} [\mathbb{E}_{\bar\theta_* \sim \mu_k, \tilde{\theta}_k \sim \tilde{\mu}_k} [|\bar{\theta}_* - \tilde{\theta}_k|^2  | h_{t_k}]] } \leq \sqrt{\frac{D}{\max\{\lambda_{\min,k},t_k\}}}
\leq \sqrt{\frac{D}{t_{k}}},
\]
where $D = 114\frac{dn}{m}$.
Similarly, second term of \eqref{eq:lem3:3} is bounded as
\begin{equation}\nonumber
\begin{split}
\mathbb{E} \bigg [
 \sum_{k=1}^{n_T} \sum_{t=t_k}^{t_{k+1} -1} |\nu_t|^2|\bar{\theta}_* - \tilde{\theta}_k|
\bigg ]
&\leq \sum_{k=1}^{n_T} \sum_{t=t_k}^{t_{k+1} -1} \sqrt{\mathbb{E}[|\nu_t|^4]} \sqrt{\mathbb{E}[|\bar{\theta}_* - \tilde{\theta}_k|^2 ]}\\
&\leq 32\bar{L}_{\nu}^2\sum_{k=1}^{n_T} \sum_{t=t_k}^{t_{k+1} -1}\sqrt{\mathbb{E}[|\bar{\theta}_* - \tilde{\theta}_k|^2 ]}\\
&\leq 32\bar{L}_{\nu}^2\sqrt{D}\sum_{k=1}^{n_T} \sum_{t=t_k}^{t_{k+1} -1}\frac{1}{\sqrt{t_k}}.
\end{split}    
\end{equation}
Now putting these together with Theorem \ref{thm:improved}, we obtain
\begin{equation}
\begin{split}
R_1&\leq 4M_{P^*} S\sqrt{D}(C M_K^2+32\bar{L}_{\nu}^2) \sum_{k=1}^{n_T} \frac{T_k}{\sqrt{t_k}}.
\end{split}
\end{equation}
Finally, to bound $\sum_{k=1}^{n_T} \frac{T_k}{\sqrt{t_k}}$, we recall that 
$T_k = k+1$ and $t_k = t_{k-1} + T_{k-1}$. 
Thus, $t_k = \frac{T_k(T_k-1)}{2}$.
Then, the sum $\sum_{k=1}^{n_T} \frac{T_k}{\sqrt{t_k}}$ is bounded as follows:
\begin{equation}\label{eq:lem1:t}
    \sum_{k=1}^{n_T} \frac{T_k}{\sqrt{t_k}} \leq \sum_{k=1}^{n_T} \frac{\sqrt{2}T_k}{\sqrt{T_k(T_k-1)}}\leq \sum_{k=1}^{n_T} 2 = 2n_T.
\end{equation}
Therefore, the result follows. 
\end{proof}

\begin{proof}[Proof of Theorem~\ref{thm:regret}]
Combining Theorem~\ref{thm:improved} and Lemma~\ref{lem:r1}, we finally prove Theorem~\ref{thm:regret}, which yields the $O(\sqrt{T})$ regret bound. Recall that the system parameter sampled in Algorithm~\ref{alg:ula} is denoted by $\tilde{\theta}_k$, which is used in obtaining the control gain matrix $K_k = K(\tilde{\theta}_k)$ for $t \in [t_k,t_{k+1})$. Let $P^*_k := P^* (\tilde{\theta}_k)$ for brevity and $\Tilde{u}_t = K_k x_t$ be an optimal action for $\tilde{\theta}_k$. Fix an arbitrary $t \in [t_k, t_{k+1})$. Then, the Bellman equation {\cite{bertsekas2011dynamic}} for $t$ in episode $k$ is given by
\begin{equation}\label{eq:main1}
\begin{split}
&J(\tilde{\theta}_k) + x_t^\top P^*_k x_t\\
 &= 
x_t^{\top} Q x_t + \Tilde{u}_t^{\top} R \Tilde{u}_t + \mathbb{E} [ (\tilde{A}_kx_t+\Tilde{B}_k\Tilde{u}_t+ w_t)^\top P^*_k (\tilde{A}_kx_t+\Tilde{B}_k\Tilde{u}_t + w_t) \mid h_t]\\
&= x_t^{\top} Qx_t + \Tilde{u}_t^{\top}R \Tilde{u}_t +  (\tilde{A}_kx_t+\Tilde{B}_k\Tilde{u}_t )^\top P^*_k(\tilde{A}_kx_t+\Tilde{B}_k\Tilde{u}_t) + \mathbb{E} [ w_t^\top P^*_k w_t \mid h_t],
\end{split}
\end{equation}
where the expectation is taken with respect to $w_t$, and
the second inequality holds because the mean of $w_t$ is zero.
On the other hand, the observed next state is expressed as
\[
x_{t+1} = \bar{\Theta}_*^\top z_t + w_t,
\]
where $\bar{\Theta}_* \in \mathbb{R}^{d \times n}$ is the matrix of the true parameter random variables. 
We then notice that
\begin{equation}\label{eq:main2}
\begin{split}
&\mathbb{E} [ w_t^\top P^*_k w_t \mid h_t]=\mathbb{E} [ x_{t+1}^\top P^*_k x_{t+1} \mid h_t] - 
(\bar{\Theta}_*^\top z_t )^\top P^*_k (\bar{\Theta}_*^\top z_t ).
\end{split}
\end{equation}
Plugging \eqref{eq:main2} into \eqref{eq:main1} and rearranging it,
\begin{equation}\label{eq:main3}
\begin{split}
x_t^{\top} Q x_t + \Tilde{u}_t^{\top} R \Tilde{u}_t  &= J(\tilde{\theta}_k) + x_t^\top P^*_k x_t 
-\mathbb{E} [ x_{t+1}^\top P^*_k x_{t+1} \mid h_t]\\
&+ (\bar{\Theta}_*^\top z_t )^\top P^*_k (\bar{\Theta}_*^\top z_t )
- (\tilde{A}_kx_t+\Tilde{B}_k\Tilde{u}_t )^\top P^*_k (\tilde{A}_kx_t+\Tilde{B}_k\Tilde{u}_t). 
\end{split}
\end{equation}
Since $\Tilde{u}_t = u_t - \nu_t$, we derive that
\begin{equation}\label{eq:main4}
    \Tilde{u}_t^{\top}R\Tilde{u}_t =u_t^{\top}Ru_t -\nu_t^{\top} R \Tilde{u}_t-\Tilde{u}_t^{\top}R\nu_t - \nu_t^{\top} R \nu_t,
\end{equation}
and
\begin{equation}\label{eq:main5}
\begin{split}
  (\tilde{A}_kx_t+\Tilde{B}_k\Tilde{u}_t )^\top P^*_k (\tilde{A}_kx_t+\Tilde{B}_k\Tilde{u}_t) &= (\tilde{\Theta}_k^\top z_t )^\top P^*_k (\tilde{\Theta}_k^\top z_t )-(\Tilde{B}_k\nu_t)^{\top}P^*_k(\Tilde{A}_kx_t)-(\Tilde{A}_kx_t)^{\top}P^*_k(\Tilde{B}_k\nu_t)\\
    &\quad\quad-(\Tilde{B}_k\nu_t)^{\top}P^*_k(\Tilde{B}_k\tilde{u}_t)-(\Tilde{B}_k\tilde{u}_t)P^*_k(\Tilde{B}_k\nu_t)- \nu_t^{\top} \Tilde{B}_k^{\top}P^*_k\Tilde{B}_k\nu_t.
\end{split}
\end{equation}
Combining \eqref{eq:main3}, \eqref{eq:main4} and \eqref{eq:main5}, we conclude that 
\begin{equation*}
\begin{split}
\mathbb{E}[c (x_t, u_t)] &= \mathbb{E}[x_t^{\top} Q x_t + u_t^{\top} R u_t] \\
&= J(\tilde{\theta}_k) + x_t^\top P^*_k x_t 
-\mathbb{E} [ x_{t+1}^\top P^*_k x_{t+1} \mid h_t]\\
&+ (\bar{\Theta}_*^\top z_t )^\top P^*_k (\bar{\Theta}_*^\top z_t )
- (\tilde{\Theta}_k^\top z_t )^\top P^*_k (\tilde{\Theta}_k^\top z_t ) + \mathbb{E}[\nu_t^{\top} \Tilde{B}_k^{\top}P^*_k\Tilde{B}_k\nu_t] +\mathbb{E}[\nu_t^{\top} R \nu_t],
\end{split}
\end{equation*}
where the expectation is taken with respect to $w_t$ and $\nu_t$.

Using this expression and observing $t_{n_T} \leq T \leq t_{n_T+1} - 1$, the expected regret of Algorithm~\ref{alg:ula} is decomposed as
\begin{equation}\nonumber
\begin{split}
R(T) 
& =  \mathbb{E} \bigg [
\sum_{k=1}^{n_T} \sum_{t=t_k}^{t_{k+1} - 1} ( c (x_t, u_t) - J (\bar{\theta}_*))
\bigg ] - \mathbb{E} \bigg [ \sum_{t=T+1}^{t_{n_T+1} - 1} ( c (x_t, u_t) - J (\bar{\theta}_*))  \bigg ]\\&:= R_1 + R_2 + R_3 + R_4 + R_5,
\end{split}
\end{equation}
where
\begin{equation}\nonumber
\begin{split}
R_1 &= \mathbb{E}\bigg[
\sum_{k=1}^{n_T} \sum_{t=t_k}^{t_{k+1} - 1} z_t^\top (
\bar{\Theta}_* P^*_k \bar{\Theta}_*^\top 
- \tilde{\Theta}_k P^*_k\tilde{\Theta}_k^\top 
) z_t 
\bigg ],\\
R_2 &= \mathbb{E}\bigg[
\sum_{k=1}^{n_T} \sum_{t=t_k}^{t_{k+1} - 1} (
x_t^\top P^*_k x_t - 
\mathbb{E} [x_{t+1}^\top P^*_k x_{t+1} | h_t]
)
\bigg ],\\
R_3 &= \mathbb{E} \bigg [
\sum_{k=1}^{n_T} T_k (J (\tilde{\theta}_k) - J(\bar{\theta}_*))
\bigg ],\\
R_4 &=\mathbb{E}\bigg[\sum_{k=1}^{n_T} \sum_{t=t_k}^{t_{k+1} - 1}(\nu_t^{\top} \Tilde{B}_k^{\top}P^*_k\Tilde{B}_k\nu_t +\nu_t^{\top} R \nu_t)\bigg],\\
R_5 &= \mathbb{E} \bigg [ \sum_{t=T+1}^{t_{n_T+1} - 1} (  J (\bar{\theta}_*) - c (x_t, u_t))  \bigg ].
\end{split}
\end{equation}
To obtain the exact regret bound, we include $R_5$  which is not considered in \cite{ouyang2019posterior}.  
By Lemma~\ref{lem:r1}, $R_1$ is bounded as
\[
R_1 \leq 8M_{P^*} S\sqrt{D}(C M_K^2+32\bar{L}_{\nu}^2) n_T.
\]
Since $T_k = k+1$, we have
\begin{equation}\nonumber
\begin{split}
T &\geq 1 + \sum_{k=1}^{n_T-1} T_{k} =  \frac{n_T(n_T+1)}{2} \geq\frac{n_T^2}{2},
\end{split}
\end{equation}
which implies that 
\begin{equation}\label{eq:thm:n}
n_T \leq \sqrt{2 T}.
\end{equation} 
Therefore, we conclude that 
\[
R_1 \leq 8\sqrt{2}M_{P^*} S\sqrt{D}(C M_K^2+32\bar{L}_{\nu}^2) \sqrt{T}.
\]

Regarding $R_2$, we use the tower rule $\mathbb{E} [ \mathbb{E} [ X_t | h_t]] = \mathbb{E} [X_t]$ to obtain 
\begin{equation}\nonumber
\begin{split}
R_2 
&= \mathbb{E}\bigg[
\sum_{k=1}^{n_T} \sum_{t=t_k}^{t_{k+1} - 1} (
x_t^\top P^*_k x_t - 
x_{t+1}^\top P^*_k x_{t+1} 
)
\bigg ] \\
&= \mathbb{E} \bigg [
\sum_{k=1}^{n_T} (
x_{t_k}^\top P^*_k x_{t_k} - 
x_{t_{k+1}}^\top P^*_k x_{t_{k+1}}
)
\bigg ]\\
& \leq \mathbb{E} \bigg [
\sum_{k=1}^{n_T} 
x_{t_k}^\top P^*_k x_{t_k}
\bigg ]\\
&\leq  \mathbb{E} \bigg [
\sum_{k=1}^{n_T} 
M_{P^*} | x_{t_k}|^2
\bigg ]\\
&\leq M_{P^*} C n_T \quad (\because \mbox{Theorem~\ref{thm:improved}})\\
&\leq M_{P^*} C \sqrt{2T},
\end{split}
\end{equation}
where the last inequality follows from \eqref{eq:thm:n}. 

We also need to deal with $R_3$ carefully. What is different from the analysis presented in \cite{ouyang2019posterior}, the term simply vanishes using the intrinsic property of probability matching of Thompson sampling as exact posterior distributions are used. However, in our analysis, approximate posterior is considered instead so a different approach is required. To cope with this problem, we adopt the notion of Lipschitz continuity of $J$ for estimation. Specifically, 
\begin{equation}\nonumber
\begin{split}
R_3 &\leq \mathbb{E} \bigg [
\sum_{k=1}^{n_T} T_k |J (\tilde{\theta}_k) - J(\bar{\theta}_*)| 
\bigg ]\\
&\leq \mathbb{E} \bigg [
\sum_{k=1}^{n_T} T_k L_J  |\tilde{\theta}_k -  \bar{\theta}_*| 
\bigg ]\\
&= \sum_{k=1}^{n_T}  T_k L_J \mathbb{E} \big [ \mathbb{E} [
  |\tilde{\theta}_k -  \bar{\theta}_*|  | h_{t_k} ]
\big ] \\
& \leq \sum_{k=1}^{n_T}  T_k L_J \mathbb{E} \big [ \mathbb{E} [
  |\tilde{\theta}_k -  \bar{\theta}_*|^2  | h_{t_k} ]^{\frac{1}{2}}
\big ]\\
&\leq  \sum_{k=1}^{n_T} L_J \sqrt{D} T_k  \frac{1}{\sqrt{t_k}},
\end{split}
\end{equation}
where $L_J$ is a Lipschitz constant of $J$ and the last inequality follows from Proposition~\ref{prop:ula} with $D = 114\frac{dn}{m}$. 

Using the bound~\eqref{eq:lem1:t} of $\sum_{k=1}^{n_T} \frac{T_k}{\sqrt{t_k}}$ in the proof of Lemma~\ref{lem:r1}, we have
\begin{equation}\nonumber
\begin{split}
R_3 &\leq 2  L_J \sqrt{D} n_T\\
&\leq 2\sqrt{2} L_J \sqrt{D} \sqrt{T}.
\end{split}
\end{equation}
By the definition of $\nu_t$, $R_4$ is bounded as
\begin{equation}\nonumber
\begin{split}
R_4 &=\mathbb{E}\bigg[\sum_{k=1}^{n_T} \sum_{t=t_k}^{t_{k+1} - 1}(\nu_t^{\top} \Tilde{B}_k^{\top}P^*_k\Tilde{B}_k\nu_t +\nu_t^{\top} R \nu_t)\bigg]\\
&\leq \mathbb{E}\bigg[\sum_{k=1}^{n_T}\sum_{t=t_k}^{t_{k+1} - 1}(S^2M_{P^*}+|R|)|\nu_t|^2\bigg]\\
&= \sum_{k=1}^{n_T}(S^2M_{P^*}+|R|)\mathrm{tr}(\mathbf{W}')\\
&\leq (S^2M_{P^*}+|R|)\mathrm{tr}(\mathbf{W}')n_T\\
&\leq (S^2M_{P^*}+|R|)\mathrm{tr}(\mathbf{W}')\sqrt{2T},
\end{split}
\end{equation}
where $M_{P^*}$ satisfies $P^*(\theta)\leq M_{P^*}$ for $\theta\in\mathcal{C}$. 
Lastly, $R_5$ is bounded as
\begin{equation}\nonumber
\begin{split}
R_5 &= \mathbb{E} \bigg [ \sum_{t=T+1}^{t_{n_T+1} - 1} (  J (\bar{\theta}_*) - c (x_t, u_t))  \bigg ]\\
&\leq \mathbb{E} \bigg [ \sum_{t=T+1}^{t_{n_T+1} - 1}  J (\bar{\theta}_*) \bigg ]\\
&\leq (t_{n_T+1} - T-1 )M_J\\
&\leq (T_{n_T}-1) M_J \quad (\because t_{n_T} \leq T \leq t_{n_T +1} - 1)\\
&= M_J n_T \\
&\leq M_J \sqrt{2T},
\end{split}
\end{equation}
where $M_J$ satisfies $J(\theta)\leq M_J$ for $\theta \in \mathcal{C}$.
Putting all the bounds together, we conclude that
\begin{equation}\nonumber
\begin{split}
R(T) &\leq C \sqrt{T}, 
 \end{split}
\end{equation}
and thus the result follows. One novelty in our analysis is that the concentration of approximate posterior is naturally embedded into the analysis, which eventually drops the $\log T$ term in the resulting regret.
\end{proof}

\section{Lemmas}

\subsection{Self-normalization lemma}\label{app:lem:self_norm}

\begin{lem}[Theorem 1 \cite{abbasi2011improved}, self-normalized bound for vector-valued martingales]\label{lem:self_norm}
Let $(\mathcal{F}_s)_{s=1}^\infty$ be a filtration. Let $(\eta_s)_{s=1}^\infty$ be a real-valued stochastic process such that $\eta_s$ is $\mathcal{F}_s$-measurable and $\eta_s$ is conditionally $R$-sub-Gaussian for some $R>0$. Let $(X_s)_{s=1}^\infty$ be an $\mathbb{R}^d$-valued stochastic process such that $X_{s}$ is $\mathcal{F}_{s-1}$-measurable. For any $t\geq 0$, define
\[
V_t = \lambda I_d + \sum_{s=1}^t X_s X_s^\top ,\quad S_t = \sum _{s=1}^t \eta_s X_s,
\]
where $\lambda >0$ is given constant. Then, for any $\delta>0$, the inequality
\[
|S_t|_{V_t^{-1}}^2 \leq 2R^2 \log \bigg( \frac{1}{\delta} \sqrt{\frac{\det(V_t)}{\det(\lambda I_d)}}\bigg), \quad t \geq 0
\]
holds with probability no less than $1-\delta$. 
\end{lem}

\subsection{Maximum norm bound}
\begin{lem}[Lemma 5 in \cite{abbasi2011regret}]\label{lem:state_log}
For any $t = 1, \ldots, T$, the following inequality holds:
\[
\mathbf{1}_{F_t}\max_{j\leq t}|x_j| \leq C \left(\log\bigg(\frac{1}{\delta}\bigg)^3\sqrt{\log\bigg(\frac{t}{\delta}\bigg)}\right)^{d+1}
\]
for some constant $C > 0$ depending only on $d, m, \rho, M_\rho, \bar{L}_\nu$ and $S$.
\end{lem}

\begin{proof}
On the event $F_t$, define $X_t: = \max_{j\leq t}|x_j| \leq \alpha_t$. Here, we may assume that $X_t \geq 1$ as the result above holds with some $C>0$ large enough when $X_t <1$. 

Recall that 
\[
\alpha_t=\frac{1}{1-\rho}\bigg(\frac{M_{\rho}}{\rho}\bigg)^d\bigg(G(\max_{j \leq  t}|z_j|)^{\frac{d}{d+1}}\beta_{t}(\delta)^{\frac{1}{2(d+1)}}+d(\bar L+S\bar L_{\nu})\sqrt{2\log\bigg(\frac{2t^2(t+1)}{\delta}\bigg)}\bigg),
\]
and $\alpha_t$ is monotone increasing in $F_t$. From
\begin{align*}
X_t=\max_{j\leq t}|x_j| \leq \alpha_t,
\end{align*}
in $F_t$, we derive that
\begin{equation}\label{eq:lem7}
X_t \leq G_1 \beta_t(\delta) X_t^{\frac{d}{d+1}} + G_2 \sqrt{\log\bigg(\frac{t}{\delta}\bigg)}
\end{equation}
by choosing constants $G_i$'s appropriately. Let us recall $\beta_t(\delta)$ which is given as
\[
\beta_t(\delta) = e(t(t+1))^{-1/\log \delta}\bigg(10\sqrt{\frac{dn}{m}\log \bigg(\frac{1}{\delta}\bigg)}+ 2\log \bigg(\frac{1}{\delta}\bigg) \sqrt{\frac{8M^2n}{m^3}\log \bigg(\frac{nt(t+1)}{\delta}\bigg(\frac{\lambda_{\max,t}}{\lambda}\bigg)^{\frac{d}{2}}\bigg)+ C}\bigg)\bigg).
\]
For $\delta \leq \frac{1}{t}$, 
\begin{equation*}
    \begin{split}
        (t(t+1))^{-1/\log \delta} &\leq (t(t+1))^{1/\log t}\\ 
        &\leq (2t^2)^{1/\log t}\\
        &=2^{1/\log t} t^{2/ \log t}\\
        &\leq e^3.
    \end{split}
\end{equation*}
As a result, 
\[
\beta_t(\delta)\leq e^4\bigg(10\sqrt{\frac{dn}{m}\log \bigg(\frac{1}{\delta}\bigg)}+ 2\log \bigg(\frac{1}{\delta}\bigg) \sqrt{\frac{8M^2n}{m^3}\log \bigg(\frac{nt(t+1)}{\delta}\bigg(\frac{\lambda_{\max,t}}{\lambda}\bigg)^{\frac{d}{2}}\bigg)+ C}\bigg)\bigg)=:\beta_t'(\delta).
\]

In turn,~\eqref{eq:lem7} implies that
\[
X_t \leq G_1 \beta'_t(\delta) X_t^{\frac{d}{d+1}} + G_2 \sqrt{\log\bigg(\frac{t}{\delta}\bigg)}.
\]
We now claim that one further has
\begin{equation}\label{eq:lem_eq}
X_t \leq \bigg(G_1 \beta'_t(\delta)  + G_2 \sqrt{\log\bigg(\frac{t}{\delta}\bigg)}\bigg)^{d+1},
\end{equation}
when $G_1 \beta'_t(\delta) + G_2 \sqrt{\log\left(\frac{t}{\delta}\right)} \geq 1$. To see this, set
\[
f(x)=x- \alpha x^\frac{d}{d+1} - \beta
\]
with $\alpha=G_1 \beta'_t(\delta)$ and $\beta=G_2 \sqrt{\log\bigg(\frac{t}{\delta}\bigg)}$. Here, we may assume that $\alpha+\beta \geq 1$ by adjusting the constants. Clearly, $f(x)$ is increasing when $x>\big(\frac{\alpha d}{d+1}\big)^{d+1}$ and $\frac{\alpha d}{d+1}<\alpha$. Since $\alpha+\beta \geq 1$,
\[
f((\alpha+\beta)^{d+1}) = \beta(\alpha+\beta)^d - \beta \geq 0, 
\]
and it follows that $x \leq (\alpha+\beta)^{d+1}$ whenever $f(x)\leq 0$. Therefore, the claim follows. 

To proceed let us estimate $\beta'_t(\delta)$. We first see that the preconditioner $P_t$ satisfies 
\begin{equation}\label{lem7:eq2}
 \lambda_{\max,t} \leq \frac{1}{n}\mathrm{tr}(P_t) = d\lambda+\sum_{s=1}^{t-1} |z_s|^2 \leq d\lambda + M_K^2 t X_t^2+td \bar L_{\nu}\sqrt{2\log\bigg(\frac{2t^2(t+1)}{\delta}\bigg)},
\end{equation}
where $M_K$ satisfies $|[I \quad K(\theta)^\top ]| \leq M_K$ for $\theta \in \mathcal{C}$. Using this relation, one derives that 
\begin{equation}\label{ineq:lem}
    \begin{split}
    \beta_t'(\delta) 
    &= G_1\sqrt{\log\bigg(\frac{1}{\delta}\bigg)} + G_2\log\bigg(\frac{1}{\delta}\bigg)\sqrt{G_3\log X_t + G_4\log\bigg(\frac{t}{\delta}\bigg)+C}\\
    &\leq G_1\sqrt{\log\bigg(\frac{1}{\delta}\bigg)} + G_2\log\bigg(\frac{1}{\delta}\bigg)\sqrt{\log X_t}+ G_3\log\bigg(\frac{1}{\delta}\bigg)\sqrt{\log\bigg(\frac{t}{\delta}\bigg)}+G_4\log\bigg(\frac{1}{\delta}\bigg)
    \end{split}
\end{equation}
for appropriately chosen $G_i>0$. Here, $G_i$'s represent different constants whenever it appears for brevity.

Define $a_t := X_t^{\frac{1}{d+1}} \geq 1$. Combining~\eqref{eq:lem_eq} and~\eqref{ineq:lem},
\[
a_t \leq  G_1\log\bigg(\frac{1}{\delta}\bigg)\sqrt{\log a_t }+ G_2\log\bigg(\frac{1}{\delta}\bigg)\sqrt{\log\bigg(\frac{t}{\delta}\bigg)}.
\]
To finish the proof, we claim the following.

\textbf{Claim]} Given $c_1,c_2 \geq 1$, when $x \geq 1$ satisfies
\[
x \leq c_1 \sqrt{\log x } +c_2,
\]
then $x\leq C c_1^2 c_2$ where $C$ is independent of $c_1$ and $c_2$.
\begin{proof}[Proof of the Claim]
Let
\[
f(x)=x-c_1\sqrt{\log x} -c_2.
\] 
From 
\begin{align*}
f(x) \geq x-c_1\sqrt{x}-c_2=
(\sqrt{x}-\frac{c_1+\sqrt{c_1^2+4c_2}}{2})(\sqrt{x}-\frac{c_1-\sqrt{c_1^2+4c_2}}{2}),
\end{align*}
$f(x)\leq 0$ implies that $x\leq C c_1^2 c_2$ from some $C>0$ which is independent of $c_1$ and $c_2$.
\end{proof}
Finally, setting
\[
c_1=G_1 \log\bigg(\frac{1}{\delta}\bigg)\text{ and } c_2 = \log\bigg(\frac{1}{\delta}\bigg) \sqrt{\log \bigg(\frac{t}{\delta}\bigg)},
\]
we deduce that  
\[
a_t \leq G_1^2\log\bigg(\frac{1}{\delta}\bigg)^3\sqrt{\log\bigg(\frac{t}{\delta}\bigg)}.
\]
\end{proof}

\subsection{Lemmas for Theorem~\ref{thm:state_poly}}

Recall the setup and notation in Section~\ref{app:polystate}.

\begin{lem}\label{lem:abbasi_lem_18}
For any $t = 1, \ldots, T$, on the event $E_t$
\begin{equation*}
\max_{s\leq t,s\notin \mathcal{T}_t} |M_s^\top z_s| \leq GZ_t^{\frac{d}{d+1}} \beta_t(\delta)^{\frac{1}{2(d+1)}},
\end{equation*}
where $G=\big(H^{-1/(d+1)}+H^{d/(d+1)}\big)\big(\frac{2Sd^{d+0.5}}{\sqrt{U}}\big)^{\frac{1}{d+1}}$ and $Z_t=\max_{s\leq t}|z_s|.$
\end{lem}

\begin{proof}
We note that the following inequalities hold on the event $E_t$:
\begin{equation*}
\begin{split}
    \beta_t(\delta)\geq |\tilde{\theta}_t-\theta_*|_{P_t} &= \sum_{i,i'=1}^{d}\sum_{j,j'=1}^{n}(\tilde{\theta}_t-\theta_*)_{d(j-1)+i}P_{d(j-1)+i,d(j'-1)+i'}(\tilde{\theta}_t-\theta_*)_{d(j'-1)+i'}\\
    & = \sum_{i,i'=1}^{d}\sum_{j,j'=1}^{n} (\tilde{\Theta}_t-\Theta_*)_{ij}(I_n)_{jj'}\bigg (\sum_{s=1}^{t-1}z_sz_s^{\top}+\lambda I_d\bigg )_{ii'}(\tilde{\Theta}_t-\Theta_*)_{i'j'}\\
    & = \sum_{i,i'=1}^{d}\sum_{j=1}^{n} (\tilde{\Theta}_t-\Theta_*)^{\top}_{ji}\bigg( \sum_{s=1}^{t-1}z_sz_s^{\top}+\lambda I_d\bigg )_{ii'}(\tilde{\Theta}_t-\Theta_*)_{i'j}\\
    &= \mathbf{tr}\left (M_t^\top \bigg (\sum_{s=1}^{t-1} z_s z_s^\top +\lambda I_d\bigg )M_t\right)\\
    &\geq \max_{1\leq s < t} |M_t^\top z_s|^2.
\end{split}
\end{equation*}
The rest of the proof follows that of Lemma 18 in~\cite{abbasi2011improved} and we provide the details for completeness.

Let us assume that $\epsilon<1$ for this moment and get back to this part later with a particular choice of $\epsilon$. From \eqref{lem:17}, we obtain,
\[
\sqrt{U}\epsilon^{d}|\pi(z_s,\mathcal{B}_s)|\leq \sqrt{i(s)}\max_{1 \leq i \leq i(s)}|M_{\tilde t_i}^{\top}z_s|,
\]
which implies that
\begin{equation}
    |\pi(z_s,\mathcal{B}_s)|\leq \sqrt{\frac{d}{U}}\frac{1}{\epsilon^{d}}\max_{1 \leq i \leq i(s)}|M_{\tilde t_i}^{\top}z_s|.
\end{equation}
Using \eqref{perpend} and \eqref{lem:17},
\begin{align*}
    |M_s^{\top}z_s| &= |(\pi(M_s,\mathcal{B}^\perp_s)+\pi(M_s,\mathcal{B}_s))^{\top}(\pi(z_s,\mathcal{B}^\perp_s)+\pi(z_s,\mathcal{B}_s))|\\
    &=|\pi(M_s,\mathcal{B}^\perp_s)^{\top}\pi(z_s,\mathcal{B}^\perp_s)+\pi(M_s,\mathcal{B}_s)^{\top}\pi(z_s,\mathcal{B}_s)|\\
    &\leq |\pi(M_s,\mathcal{B}^\perp_s)^{\top}\pi(z_s,\mathcal{B}^\perp_s)|+|\pi(M_s,\mathcal{B}_s)^{\top}\pi(z_s,\mathcal{B}_s)|\\
    &\leq d\epsilon|z_s|+2S\sqrt{\frac{d}{U}}\frac{1}{\epsilon^{d}}\max_{1 \leq i \leq i(s)}|M_{\tilde t_i}^{\top}z_s|.
\end{align*}
Since $Z_t$ is increasing in $t$, we have
\begin{align*}
    \max_{s\leq t,s\notin \mathcal{T}_{t}}|M_s^{\top}z_s| &\leq  d\epsilon Z_t+2S\sqrt{\frac{d}{U}}\frac{1}{\epsilon^{d}}\max_{s\leq t,s\notin \mathcal{T}_{t}}\max_{1 \leq i \leq i(s)}|M_{\tilde t_i}^{\top}z_s|.
\end{align*}
Recalling the definition of $i(s)$, the condition $s\notin \mathcal{T}_t$ and $1\leq i \leq i(s)$ implies that $s< \tilde t_i$. Therefore, for $\delta<1$,
\begin{align*}
\max_{s\leq t,s\notin \mathcal{T}_{t}}\max_{1 \leq i \leq i(s)}|M_{\tilde t_i}^{\top}z_s| &\leq \max_{i}\max_{s<\tilde t_i}|M_{\tilde t_i}^{\top}z_s| \\
& \leq \beta_t(\delta)^{\frac{1}{2}}.
\end{align*}
Hence, we deduce that
\begin{equation}\label{eq:mszs}
    \max_{s\leq t,s\notin \mathcal{T}_{t}}|M_s^{\top}z_s| 
    \leq d\epsilon Z_t+2S\sqrt{\frac{d}{U}}\frac{1}{\epsilon^{d}}\beta_{t}(\delta)^{\frac{1}{2}}.
\end{equation}
Let us choose $\epsilon  = \bigg(\frac{2S\beta_t(\delta)^{1/2}}{Z_td^{1/2}U^{1/2}H}\bigg)^{1/(d+1)}$ with the choice of $H>\max\{16,\frac{4S^2\tilde M^2}{dU_0}\}$.

To further simplify~\eqref{eq:mszs}, 
\begin{align*}
     \max_{s\leq t,s\notin \mathcal{T}_{t}}|M_s^{\top}z_s|& \leq \big(H^{-1/(d+1)}+H^{d/(d+1)}\big)\bigg(\frac{2S\beta_t(\delta)^{1/2}Z_t^d d^{d+1/2}}{U^{1/2}}\bigg)^{1/(d+1)}\\
     &\leq GZ_t^{\frac{d}{d+1}}\beta_t(\delta)^{\frac{1}{2(d+1)}}.
\end{align*}

Now let us show $\epsilon < 1$, which is the part we postponed at the beginning of the proof.  Since 
$H> \frac{4S^2\tilde{M}^2}{dU_0}$, a direct computation yields that
\[
\bigg(\frac{4S^2{\tilde M}^2}{dU_0H}\bigg)^{\frac{1}{2(d+1)}}<1.
\]
Noting that $\lambda_{\max,t}\leq \frac{1}{n}\mathrm{tr}(P_t)=d\lambda+\sum_{s=1}^{t-1} |z_s|^2 \leq d\lambda+t|Z_t|^2$,
\begin{align*}
\frac{\beta_t(\delta)}{Z_t} &\leq e(t(t+1))^{-1/\log{\delta}}\\
&\quad \quad \times \bigg(10\sqrt{\frac{dn}{m}\log{\bigg(\frac{1}{\delta}\bigg)}}+ 2\log\bigg(\frac{1}{\delta}\bigg) \sqrt{\frac{8M^2n}{m^3}\log \bigg(\frac{nt(t+1)}{\delta}\bigg(\frac{\lambda_{\max,t}}{\lambda}\bigg)^{\frac{d}{2}}\bigg)+C}\bigg)/Z_t\\
&\leq 
\sup_Y e(t(t+1))^{-1/\log{\delta}}\\
&\quad \quad \times \bigg(10\sqrt{\frac{dn}{m}\log{\bigg(\frac{1}{\delta}\bigg)}}+ 2\log\bigg(\frac{1}{\delta}\bigg) \sqrt{\frac{8M^2n}{m^3}\log \bigg(\frac{nt(t+1)}{\delta}\bigg(d+\frac{tY^2}{\lambda}\bigg)^{\frac{d}{2}}\bigg)+C}\bigg)/Y \\
&\leq 
\sup_Y e(T(T+1))^{-1/\log{\delta}}\\
&\quad \quad \times \bigg(10\sqrt{\frac{dn}{m}\log{\bigg(\frac{1}{\delta}\bigg)}}+ 2\log\bigg(\frac{1}{\delta}\bigg) \sqrt{\frac{8M^2n}{m^3}\log \bigg(\frac{nT(T+1)}{\delta}\bigg(d+\frac{TY^2}{\lambda}\bigg)^{\frac{d}{2}}\bigg)+C}\bigg)/Y\\
&=\tilde{M}.
\end{align*}

Therefore, $\beta_t(\delta) \leq \tilde M Z_t$
holds for all $t$ and consequently,
\[
\epsilon  = \bigg(\frac{2S\beta_t(\delta)^{1/2}}{Z_td^{1/2}U^{1/2}H}\bigg)^{1/(d+1)} = \bigg(\frac{2S\beta_t(\delta)}{Z_td^{1/2}U_0^{1/2}H^{1/2}}\bigg)^{1/(d+1)} \leq \bigg(\frac{2S\tilde{M}}{d^{1/2}U_0^{1/2}H^{1/2}}\bigg)^{1/(d+1)}<1.
\]
\end{proof}

\subsection{Lemmas for Proposition~\ref{prop:pers-ex}}

\begin{lem}[Lemma 10 in \cite{jedra2022minimal}]\label{lem:preconditioner_expansion}
Let $(z_s)_{s=1}^\infty$, $(y_s)_{s=1}^\infty$ and $(\xi_s)_{s=1}^\infty$ be three sequences of vectors in $\R^d$, satisfying the linear relation $z_s = y_s + \xi_s$ for all $s\geq 0$. Then, for all $\tilde\lambda>0$, all $t\geq 1$ and all $\epsilon \in (0,1]$, we have
\[
 \sum_{s=1}^t z_s z_s^{\top} \succeq \sum_{s=1}^t \xi_s\xi_s^{\top} + (1-\epsilon)\sum_{s=1}^t y_s y_s^{\top} - \frac{1}{\epsilon}\bigg(\sum_{s=1}^t y_s  \xi_s^{\top}\bigg)^{\top}\bigg(\tilde\lambda I_d+\sum_{s=1}^{t}y_sy_s^{\top}\bigg)^{-1}\bigg(\sum_{s=1}^t y_s \xi_s^{\top}\bigg) -  \epsilon \tilde\lambda I_d.
\]
\end{lem}

\begin{lem}[Lemma 12 in \cite{jedra2022minimal}]\label{lem:tall_matrix}
For two matrices $X$,$Y$ with the same number of rows and any $\bar\lambda>0$, we have
\[
\begin{bmatrix}
X^\top X & X^\top Y\\
Y^\top X & Y^\top Y
\end{bmatrix} \succeq
\begin{bmatrix}
\frac{\bar\lambda}{|Y|^2+\bar\lambda}X^\top X & 0\\
0 & -\bar\lambda I_{n_u}
\end{bmatrix}.
\]
\end{lem}
\begin{proof}
Since
\begin{align*}
    &\begin{bmatrix}
    X^\top Y(Y^{\top}Y+\bar\lambda I_d)^{-1} Y^{\top} X & X^{\top}Y \\
    Y^\top X & Y^{\top}Y+\bar\lambda I_{n_u}
    \end{bmatrix}\\
    &=
    \begin{bmatrix}
    X^\top Y(Y^{\top}Y+\bar\lambda I_d)^{-1/2}\\ 
    (Y^{\top}Y+\bar\lambda I_d)^{1/2}
    \end{bmatrix}
    \begin{bmatrix}
    (Y^{\top}Y+\bar\lambda I_d)^{-1/2}Y^\top X& 
    (Y^{\top}Y+\bar\lambda I_d)^{1/2}
    \end{bmatrix}\\
    &\succeq 0,
\end{align*}
it is straightforward to check that
\begin{align*}
    &\begin{bmatrix}
     X^{\top}X & X^{\top}Y \\
    Y^\top X & Y^{\top}Y\\
    \end{bmatrix}\\
    &\succeq 
    \begin{bmatrix}
        X^{\top}X-X^{\top}Y(Y^{\top}Y+\bar\lambda I_{n_u})^{-1}Y^{\top}X & 0\\
        0 & -\bar\lambda I_{n_u}\\
    \end{bmatrix}\\
    &= 
    \begin{bmatrix}
        X^{\top}(I-Y(Y^{\top}Y+\bar\lambda I_{n_u})^{-1}Y^{\top})X & 0\\
        0 & -\bar\lambda I_{n_u}\\
    \end{bmatrix}\\
    &\succeq
    \begin{bmatrix}
    \frac{\bar\lambda}{|Y|^2+\bar\lambda}X^\top X & 0\\
    0 & -\bar\lambda I_{n_u}
    \end{bmatrix},
\end{align*}
where the last inequality follows from the singular value decomposition and the relation
\[
I-Y(Y^{\top}Y+\bar\lambda I_{n_u})^{-1}Y^{\top} \succeq \frac{\bar\lambda}{|Y|^2+\bar\lambda}I.
\]
\end{proof}

\begin{lem}[\cite{vershynin2018high}]\label{lem:epsilon-net}
Let $W \in \mathbb{R}^{d\times d}$ be a random matrix and $\epsilon \in (0,\frac{1}{2})$ and $\mathcal{M}$ be $\epsilon$-net in $S^{d-1}$ with minimal cardinality. Then, for any $\rho>0$,
\[
Pr(|W|>\rho) \leq \bigg(\frac{2}{\epsilon}+1\bigg)^d \max_{x\in \mathcal{M}} \mathrm{Pr}(|x^\top W x|>(1-2\epsilon) \rho).
\]
\end{lem}

\begin{lem}[Modification of Proposition 8 in \cite{jedra2022minimal}]\label{lem:noise_tail_bound}
Let $(\psi_s)_{s=1}^{\infty}$ be a sequence of independent, zero mean, $\bar L$-sub-Gaussian and $\mathcal{F}_{s}$-measurable random vector in $\R^{d}$. Then, for all $\rho'>0$, $0<\epsilon<1$ and $t \geq \max(\frac{16^2\bar{L}^4}{\epsilon^2}, \frac{16\bar{L}^2}{\epsilon})(\rho' +d \log 9)$,
\[
\mathrm{Pr}\bigg((\lambda_{\min}(\mathbb{E}[\psi_t \psi_t^{\top}])- \epsilon)tI_d\preceq \sum_{s=1}^t \psi_s \psi_s^{\top} \preceq (\lambda_{\max}(\mathbb{E}[\psi_t \psi_t^{\top}])+ \epsilon)tI_d\bigg)\geq 1-2e^{-\rho'}.
    \]
\end{lem}
\begin{proof}
Here, $\psi_s$ is zero-mean, $\bar L$-sub-Gaussian random vector satisfying
\[
\mathbb{E}[\exp(\theta^\top \psi_s)]\leq \exp\bigg(\frac{|\theta|^2 {\bar L}^2}{2}\bigg)
\]
for any vector $\theta \in \mathbb{R}^d$. Then for any unit vector $x$, $Y:=x^\top \psi_s$ is zero-mean, $\bar L$-sub-Gaussian, and hence, it follows that 
\[
\mathbb{E}[\exp \lambda (Y^2-\mathbb{E}[Y^2])] \leq \exp(16 \lambda^2 \bar L^4) 
\]
for any $|\lambda| \leq \frac{1}{4 \bar L^2}$ which follows from Appendix B in~\cite{honorio2014tight}.

With $Z_s:= Y_s^2 - \mathbb{E}[Y_s^2]$,
\begin{align*}
\mathbb{E}\bigg [\exp \bigg (\lambda\sum_{s=1}^t Z_s \bigg)\bigg ] &= \Pi_{s=1}^t \mathbb{E}[\exp(\lambda Z_s)]\\ 
& \leq \exp(16 t \lambda^2 \bar L^4),
\end{align*}
and therefore,
\[
\mathbb{E}\bigg[\exp\bigg(\lambda\sum_{s=1}^t (x^{\top}\psi_s)^2 - \lambda\sum_{s=1}^t  \mathbb{E}[(x^{\top}\psi_s)^2]\bigg)\bigg] \leq \exp(16 t \lambda^2 \bar L^4).
\]
Invoking Markov inequality, for any $\rho>0$, 
\[
Pr\bigg(\sum_{s=1}^t (x^{\top}\psi_s)^2 - \sum_{s=1}^t  \mathbb{E}[(x^{\top}\psi_s)^2] > \rho\bigg)  \leq \exp(16 t \lambda^2 \bar L^4-\lambda \rho)
\]
for any $|\lambda| \leq \frac{1}{4 \bar L^2}$. Choosing $\lambda=\min\{\frac{1}{4\bar L^2},\frac{\rho}{32 t \bar L^4 }\}$, we derive that
\[
\mathrm{Pr}\bigg(\sum_{s=1}^t (x^{\top}\psi_s)^2 - \sum_{s=1}^t  \mathbb{E}[(x^{\top}\psi_s)^2] > \rho \bigg)  \leq \exp\bigg(-\min\bigg \{\frac{\rho}{8 \bar L^2},\frac{\rho^2}{64t \bar L^4} \bigg \} \bigg).
\]
Similarly,
\[
\mathrm{Pr}\bigg(\sum_{s=1}^t  \mathbb{E}[(x^{\top}\psi_s)^2]-\sum_{s=1}^t (x^{\top}\psi_s)^2 > \rho
\bigg)  \leq \exp\bigg(-\min\bigg \{\frac{\rho}{8 \bar L^2},\frac{\rho^2}{64t \bar L^4}\bigg \}\bigg).
\]
Altogether,
\[
\mathrm{Pr}\bigg(\bigg|\sum_{s=1}^t (x^{\top}\psi_s)^2-\sum_{s=1}^t\mathbb{E}[(x^{\top}\psi_s)^2]\bigg| > \rho\bigg)  \leq 2\exp\bigg(-\min\bigg \{\frac{\rho}{8 \bar L^2},\frac{\rho^2}{64t \bar L^4}\bigg \}\bigg).
\]

Now we apply Lemma \ref{lem:epsilon-net} with $\epsilon = \frac{1}{4}$ and $W= \sum_{s=1}^{t}(\psi_s \psi_s^\top - \mathbb{E}[\psi_s \psi_s^\top]$), we have
\[
\mathrm{Pr}\bigg(\bigg|\sum_{s=1}^t \psi_s \psi_s^{\top} - \sum_{s=1}^t  \mathbb{E}[\psi_s \psi_s^{\top}]\bigg|> \rho \bigg) \leq 2\cdot9^d\exp \bigg(-\min\bigg \{\frac{\rho}{16 \bar L^2},\frac{\rho^2}{256t \bar L^4}\bigg \}\bigg).
\]

Upon substitution $\exp(-\rho')=9^d\exp (-\min \{\frac{\rho}{16 \bar L^2},\frac{\rho^2}{256t \bar L^4}\})$, or equivalently,
\[
16 \bar L^2 (\rho'+d \log 9)=\min\bigg \{\rho,\frac{\rho^2}{16 t \bar L^2} \bigg \},
\]
and solving for $\rho$, we further obtain that
\[
\mathrm{Pr}\bigg(\bigg|\sum_{s=1}^t \psi_s \psi_s^{\top} - \sum_{s=1}^t  \mathbb{E}[\psi_s \psi_s^{\top}]\bigg|> 16 \bar{L}^2 t\max \bigg \{\sqrt{\frac{\rho'+d \log 9}{t}},\frac{\rho'+d \log 9}{t}\bigg \}\bigg)\leq 2\exp(-\rho').
\]
Now for $t \geq \max \{\frac{16^2\bar{L}^4}{\epsilon^2}, \frac{16\bar{L}^2}{\epsilon}\}(\rho' +d \log 9)$, we have that
\[
\frac{\rho'+d\log 9}{t} \leq \frac{1}{\max \Big\{\frac{16^2 \bar L^4}{\epsilon^2},\frac{16 \bar L^2}{\epsilon}\Big\}}\leq \frac{\epsilon}{16 \bar L^2},
\]
and
\[
\sqrt{\frac{\rho'+d\log 9}{t}} \leq \frac{1}{\max \Big \{\frac{16 \bar L^2}{\epsilon},\sqrt{\frac{16 \bar L^2}{\epsilon}\Big\}}}\leq \frac{\epsilon}{16 \bar L^2},
\]
which implies that
\[
\epsilon t \geq 16 \bar{L}^2 t\max \bigg\{\sqrt{\frac{\rho'+d \log 9}{t}},\frac{\rho'+d \log 9}{t}\bigg\}.
\]
Therefore,
\[
\mathrm{Pr}\bigg(\bigg|\sum_{s=1}^t \psi_s \psi_s^{\top} - \sum_{s=1}^t  \mathbb{E}[\psi_s \psi_s^{\top}]\bigg|> \epsilon t\bigg)\leq 2\exp(-\rho').
\]
Since $\psi_s \psi_s^{\top}$ is symmetric, the inequality $\big|\sum_{s=1}^t \psi_s \psi_s^{\top} - \sum_{s=1}^t  \mathbb{E}[\psi_s \psi_s^{\top}]\big|\leq \epsilon t$ implies that
\[
\lambda_{\max}^2 \bigg (\sum_{s=1}^t \psi_s \psi_s^{\top} - \sum_{s=1}^t  \mathbb{E}[\psi_s \psi_s^{\top}]\bigg )\leq \epsilon^2 t^2,
\]
and
\[
\lambda_{\min}^2 \bigg (\sum_{s=1}^t \psi_s \psi_s^{\top} - \sum_{s=1}^t  \mathbb{E}[\psi_s \psi_s^{\top}]\bigg )\leq \epsilon^2 t^2.
\]
As a result, 
\begin{align*}
(\lambda_{\min}(\mathbb{E}[\psi_t \psi_t^{\top}])- \epsilon)tI_d&\preceq \sum_{s=1}^t\mathbb{E}[\psi_s \psi_s^{\top}]-\epsilon t I_d\\
&\preceq \sum_{s=1}^t \psi_s \psi_s^{\top}\\
&\preceq \sum_{s=1}^t\mathbb{E}[\psi_s \psi_s^{\top}] + \epsilon t I_d\\
&\preceq (\lambda_{\max}(\mathbb{E}[\psi_t \psi_t^{\top}])+ \epsilon)tI_d.
\end{align*}
\end{proof}

\begin{lem}[Proposition 9 in \cite{jedra2022minimal}]\label{lem:self-normalized_matrix}
Let $\mathcal{F}_s$ be a filtration and $(\psi_s)_{s=1}^{\infty}$ be a sequence of independent, zero mean, $\bar L$-sub-Gaussian and $\mathcal{F}_{s}$-measurable random vectors in $\R^{d}$. Let $(L_s)_{s=1}^{\infty}$ be a sequence of random matrices in $\R^{d\times d}$ such that $\mathcal{F}_{s-1}$-measurable and $|L_s|<\infty$. Let $(y_s)_{s=1}^{\infty}$ be a sequence of $\mathcal{F}_{s-1}$-measurable random variables in $\R^{d}$. Then for all positive definite matrix $V\succ 0$, the following self-normalized matrix process defined by 
\[
S_t(y,L\psi) = \bigg (\sum_{s=1}^t y_s (L_s \psi_s)^{\top}\bigg)^{\top} \bigg(V+\sum_{s=1}^ty_sy_s^{\top}\bigg)^{-1}\bigg (\sum_{s=1}^t y_s(L_s \psi_s)^{\top}\bigg )
\]
satisfies
\[
\mathrm{Pr}\bigg[|S_t(y,L\psi)|>\bar{L}^2(\max_{1\leq s \leq t}|L_s|^2)\bigg(2\log \bigg(\det\bigg(I_d+V^{-1}\sum_{s=1}^ty_sy_s^{\top} \bigg)\bigg)+ 4\rho +7d\bigg)\bigg] \leq e^{-\rho}
\]
for all $\rho,t \geq 1$.
\end{lem}

\section{Empirical Analyses}\label{app:exp}

We test the performance of our algorithm with Gaussian mixture noises specified in Sections \ref{app:gm} and \ref{app:agm}. 
The source code for our TSLD-LQ implementation is available online: \url{https://github.com/Jiwhan-Park/tsld}.
The true system parameter $\Theta_*$ is chosen as follows: 
\begin{itemize}
\item for $n=n_u=3$,
\begin{equation*}
A_*=
  \begin{bmatrix}
    0.3  & 0.1  & 0.2           \\
    0.1 & 0.4   & 0\\
    0 & 0.7 & 0.6
    
  \end{bmatrix}
,\quad B_*=
  \begin{bmatrix}
    0.5  & 0.4  & 0.5           \\
    0.6 & 0.3   & 0\\
    0.3 & 0 & 0.2
  \end{bmatrix},
\end{equation*}
\item for $n=n_u=5$, 
\begin{equation*}
A_*=
  \begin{bmatrix}
    0.3  & 0.6 & 0.2& 0.3 &0.1\\
    0 & 0.1 & 0.4 &0 &0.6\\
    0.1 & 0.5 & 0.3 & 0 & 0.2\\
    0.4 & 0 & 0.3 & 0.3 & 0\\
    0.3&0.3&0.1&0.4&0.4
  \end{bmatrix},\quad
    B_*=
  \begin{bmatrix}
    0.5  & 0.4 & 0.2& 0.5 &0.4\\
    0.6 & 0 & 0.3 &0.1 &0.3\\
    0.5 & 0 & 0 &0.1 &0.2\\
    0.1 & 0.5 & 0 & 0.2 &0.4\\
    0.2 & 0.1 & 0.6 & 0 & 0
  \end{bmatrix},
\end{equation*}
\item for $n=n_u=10$,
\begin{equation*}
A_*=
  \begin{bmatrix}
    0.6  & 0.6 & 0.5& 0 &0.1 & 0.4 & 0.3 & 0.3 & 0.3 & 0.4\\
    0.3&0.2&0.6&0&0.1&0&0.2&0.5&0.2&0\\
    0&0.6&0&0.3&0.4&0&0.5&0.4&0.1&0.3\\
    0.4&0.1&0.5&0.6&0.6&0.5&0.1&0.1&0.6&0\\
    0.5&0.1&0.2&0&0.1&0.1&0.1&0&0.6&0.4\\
    0.1&0.2&0.2&0.1&0.2&0&0.5&0.2&0.5&0.7\\
    0.3&0.6&0.1&0.6&0.1&0&0.3&0.4&0.6&0.3\\
    0.3&0&0.5&0.2&0.2&0.7&0.4&0.1&0.4&0.3\\
    0&0.3&0.3&0.5&0.3&0.5&0.1&0&0.1&0.5\\
    0.3&0&0&0.5&0&0.2&0.4&0.4&0&0.5
  \end{bmatrix},
\end{equation*}
\begin{equation*}
    B_*=
  \begin{bmatrix}
    0.5  & 0.4 & 0.2& 0.5 &0.4&0&0.8&0.1&0.3&0.7\\
    0.1 & 0.4 & 0.6 &0 &0.5 &0&0.3&0.1&0.3&0.2\\
    0 & 0.5 & 0 &0.6 &0.6 &0.5&0&0&0.1&0.2\\
    0.4 & 0.4 & 0.3 & 0.5 &0 & 0.1 &0.5 &0 &0.2 &0.4\\
    0.2 & 0.1 & 0.4 & 0 & 0 & 0.7 & 0.1 &0.1 &0.5 &0.3\\
    0.4 &0.5 &0 &0.6 &0 & 0.4 & 0.6 &0.1 &0.4 &0.5\\
    0.3&0.5&0&0.3&0.1&0.7&0.2&0&0.4&0.6\\
    0.2&0&0.1&0.6&0.2&0.7&0&0.1&0.4&0.4\\
    0&0.2&0.2&0.2&0&0&0&0.3&0.1&0.4\\
    0.2&0.5&0.1&0.3&0&0.5&0.4&0.4&0.2&0.3
  \end{bmatrix}.
\end{equation*}
\end{itemize}
 For the quadratic cost, $Q=2I_{n}$, $R=I_{n_u}$ are used where $n=3,5,10$. True system parameters $(A_*,B_*)$ satisfy $\rho(A_*+B_*K)=0.3365$ for $n=n_u=3$, $0.3187$ for $n=n_u=5$, and $0.3839$ for $n=n_u=10$, where $K$ denotes the control gain matrix associated with $(A_*,B_*)$. For the admissible set $\mathcal{C}$, we choose $S=20$, $M_J=20000$, and $\rho=0.99$ for both  cases regardless of the type of noise. We also sample action perturbation $\nu_s$ from $\mathcal{N}(0,\frac{1}{10000}I_{n_u})$ at the end of each episode. Finally, the prior is set to be Gaussian distribution with covariance $0.2 I_n$ for $n=n_u=3$, $n=n_u=5$ (or $\lambda=5$), and with covariance $0.1 I_n$ for $n=n_u=10$ (or $\lambda=10$). The mean of each component is set to be $0.5$.

\subsection{Regret}



\begin{figure}[h]
\centering
\begin{tabular}{ccc}
\includegraphics[width=2in,clip=false,trim=0.1cm 0.1cm 0.4cm 0.6cm]{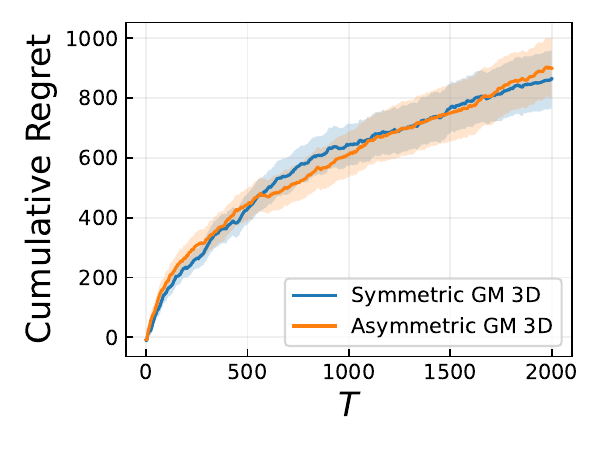}&
\includegraphics[width=2in,clip=false,trim=0.1cm 0.1cm 0.4cm 0.6cm]{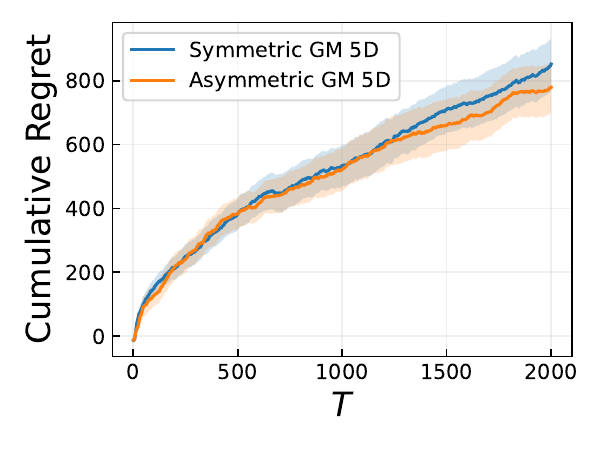}&
\includegraphics[width=2in,clip=false,trim=0.1cm 0.1cm 0.4cm 0.6cm]{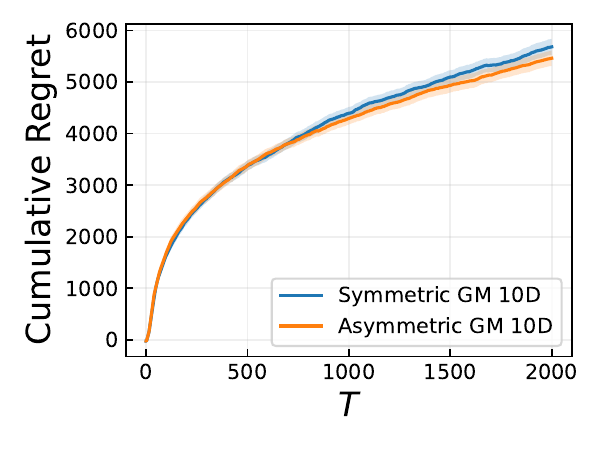}
\end{tabular}
\caption{Expected cumulative regret $R(T)$ over a time horizon $T$ using the Gaussian mixture noise for $n=n_u=3$ (left),
for $n=n_u=5$ (center), for $n=n_u=10$ (right).}\label{fig:main}
\end{figure}

We test our method with both symmetric and asymmetric Gaussian mixture noises specified in Sections~\ref{app:gm} and \ref{app:agm} respectively. 
As shown in Figure~\ref{fig:main}, the proposed algorithm achieves an $O(\sqrt{T})$ regret bound even when the noise is asymmetric.


\subsection{Effect of the preconditioner on the number of iterations}\label{sub:effect}

\begin{table}[htbp]
\caption{The number of iterations required for the naive ULA and preconditioned ULA
  when  $n=n_u=3$.}
    \centering
    \begin{tabular}{|c|c|c|c|c|}
        \hline
        Time horizon $T$ & 500 &  1000 &  1500 & 2000 \\
        \hline
        Naive ULA & $6.3\times 10^5$ &$1.8\times 10^6$  &$3.4\times 10^6$  &$5.1\times 10^6$ \\
        \hline
        Preconditioned ULA &$7.1\times 10^3$ &$1.2\times 10^4$ &$1.7\times 10^4$ & $2.1\times 10^4$ \\
        \hline
    \end{tabular}
    \label{comparison_iterations}
\end{table}

Table~\ref{comparison_iterations} shows the number of iterations computed according to Theorem~\ref{thm:naive_ula} (naive ULA) and Algorithm~\ref{alg:ula} (preconditioned ULA).
We observe a significant reduction in the number of iterations required for the sampling process when the preconditioned ULA is employed, in comparison to the naive ULA. This empirical evidence confirms that our algorithm achieves the regret bound utilizing fewer computational resources.

%
%

\subsection{Additional Analyses on Gaussian Mixture Noise}\label{app:additional}

\begin{figure}[h]
\centering
\begin{tabular}{ccc}
\includegraphics[width=2in,clip=false,trim=0.1cm 0.1cm 0.4cm 0.6cm]{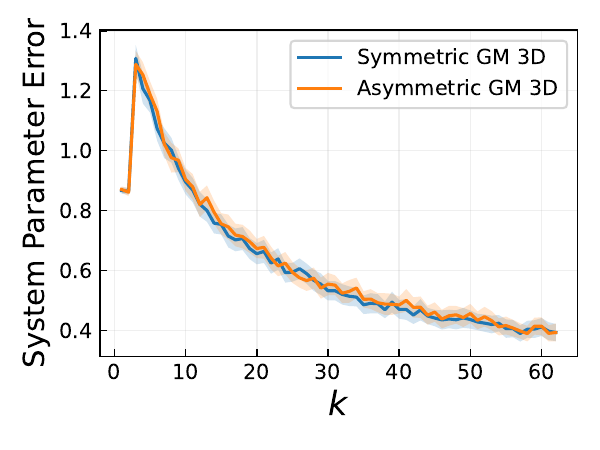}&
\includegraphics[width=2in,clip=false,trim=0.1cm 0.1cm 0.4cm 0.6cm]{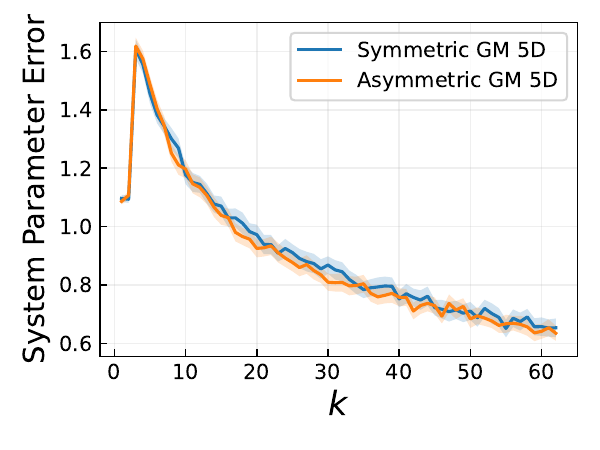}&
\includegraphics[width=2in,clip=false,trim=0.1cm 0.1cm 0.4cm 0.6cm]{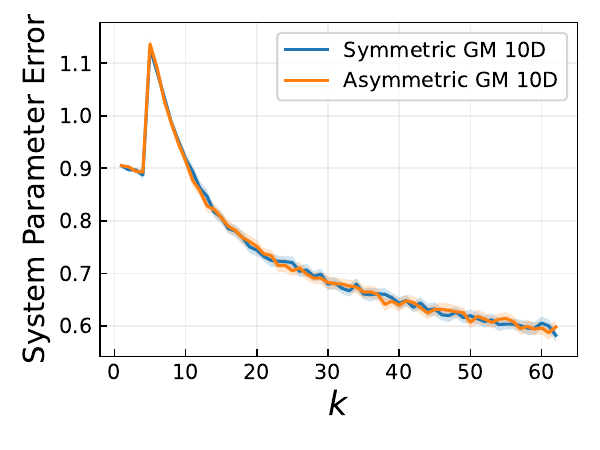}
\end{tabular}
\caption{System parameter error $|\tilde\theta_k-\theta_*| / |\theta_*|$ over episode $k$ using the Gaussian mixture noise for $n=n_u=3$ (left),
for $n=n_u=5$ (center), for $n=n_u=10$ (right).}
\label{fig:moderr}
\end{figure}

Figure~\ref{fig:moderr} shows the error between sampled and true system parameters over episode, which demonstrates its $\tilde O(t^{-\frac{1}{4}})$ convergence proved in Theorem~\ref{thm:exact_appox}.

\begin{figure}[h]
\centering
\begin{tabular}{ccc}
\includegraphics[width=2in,clip=false,trim=0.1cm 0.1cm 0.4cm 0.6cm]{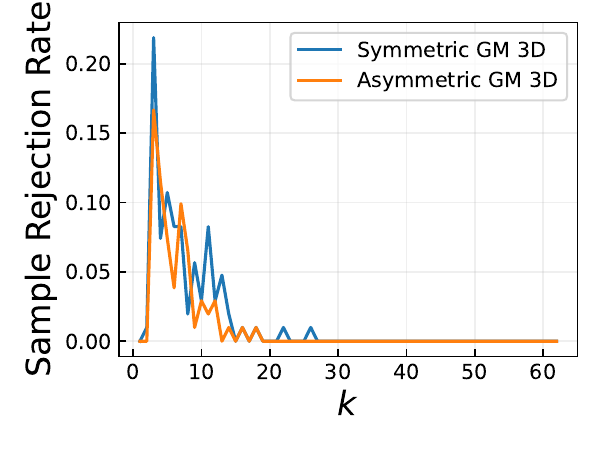}&
\includegraphics[width=2in,clip=false,trim=0.1cm 0.1cm 0.4cm 0.6cm]{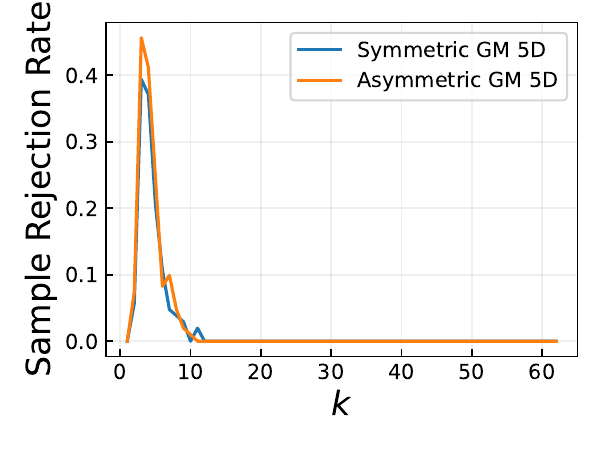}&
\includegraphics[width=2in,clip=false,trim=0.1cm 0.1cm 0.4cm 0.6cm]{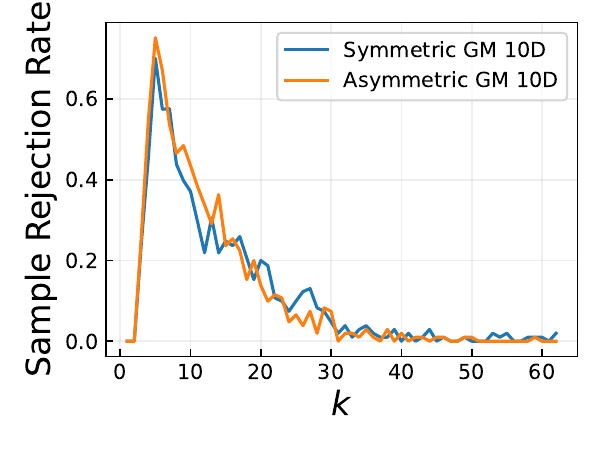}
\end{tabular}
\caption{Sample rejection rate over episode using the Gaussian mixture noise for $n=n_u=3$ (left),
for $n=n_u=5$ (center), for $n=n_u=10$ (right).}
\label{fig:rejection}
\end{figure}

The sample rejection rate of Figure~\ref{fig:rejection} is computed as $n_\mathrm{rej}/(n_\mathrm{acc}+n_\mathrm{rej})$ where $n_\mathrm{rej}$ is the total number of rejections at the episode and $n_\mathrm{acc}$ is the total number of accepted samples at the episode, which is equal to the number of simulations carried out. This result empirically shows the existence of a small positive constant $\epsilon$ that satisfies $\mathrm{Pr}(\tilde\theta_k\in\mathcal{C})\geq1-\epsilon$.

Execution time illustrated in Table~\ref{tab:gm_time} is measured on an Intel Xeon W-2295 (3.00GHz) platform equipped with an NVIDIA RTX 3090 GPU.

\begin{table}[htbp]
\caption{The mean and standard deviation of execution time of 2000 time steps of Algorithm~\ref{alg:ula} in seconds for the Gaussian mixture noise. The left column is the mean and the right column is the standard deviation for each system dimension value.}
    \centering
    \begin{tabular}{|c|c|c|c|c|c|c|}
        \hline
        System dimension $n=n_u$ & \multicolumn{2}{c|}{3} &  \multicolumn{2}{c|}{5} &  \multicolumn{2}{c|}{10} \\
        \hline
        Symmetric & $1.9\times 10^3$ & $6.2\times 10^2$ & $7.5\times 10^2$ & $1.3\times 10^2$ & $1.9\times 10^3$ & $1.5\times 10^2$ \\
        \hline
        Asymmetric & $2.2\times10^3$ & $7.7\times10^2$ & $7.0\times 10^2$ & $1.1\times 10^2$ & $2.0\times10^3$ & $1.1\times10^2$ \\
        \hline
    \end{tabular}
    \label{tab:gm_time}
\end{table}

\subsection{Gaussian mixture noise}    \label{app:gm}
We consider a Gaussian mixture noise which is given by
\begin{equation*}
    p_w(w_t) = \frac{1}{2(2\pi)^{n/2}}\Big(e^{\frac{-|w_t-a|^2}{2}}+e^{\frac{-|w_t+a|^2}{2}}\Big ),
\end{equation*}
where $a = [\frac{1}{2},\frac{1}{2},\frac{1}{2}]^{\top}$, $[\frac{1}{4},\frac{1}{4},\frac{1}{4},\frac{1}{4},\frac{1}{4}]^{\top}$ and $[\frac{1}{4},\frac{1}{4},\frac{1}{4},\frac{1}{4},\frac{1}{4},\frac{1}{4},\frac{1}{4},\frac{1}{4},\frac{1}{4},\frac{1}{4}]^{\top}$ for $n=3$, $5$ and $10$ respectively.
Taking gradients,
\[
-\nabla \log p_w(w_t) = w_t -a + \frac{2a}{1+e^{2w_t^{\top}a}},
\]
and
\begin{equation*}
    \begin{split}
        -\nabla^2 \log p_w(w_t) &= I_n - 4aa^{\top} \frac{e^{2w_t^{\top}a}}{(1+e^{2w_t^{\top}a})^2}\\
        &\succeq I_n-aa^{\top}\\
        &\succeq (1-|a|^2)I_n.
    \end{split}
\end{equation*}
Therefore, the first condition in Assumption \ref{ass:noise} is satisfied for $n=3$, $5$ and $10$:
\[
\frac{1}{4}I_3\preceq -\nabla^2 \log p_w(w_t)\preceq I_3,
\]
\[
\frac{11}{16}I_5\preceq -\nabla^2 \log p_w(w_t)\preceq I_5,
\]
\[
\frac{3}{8}I_{10}\preceq -\nabla^2 \log p_w(w_t)\preceq I_{10}.
\]

\begin{figure}[h]
    \centering
    \begin{tabular}{cc}
\includegraphics[width=0.7\textwidth,clip=false,trim=0.2cm 0.1cm 0.4cm 0.6cm]{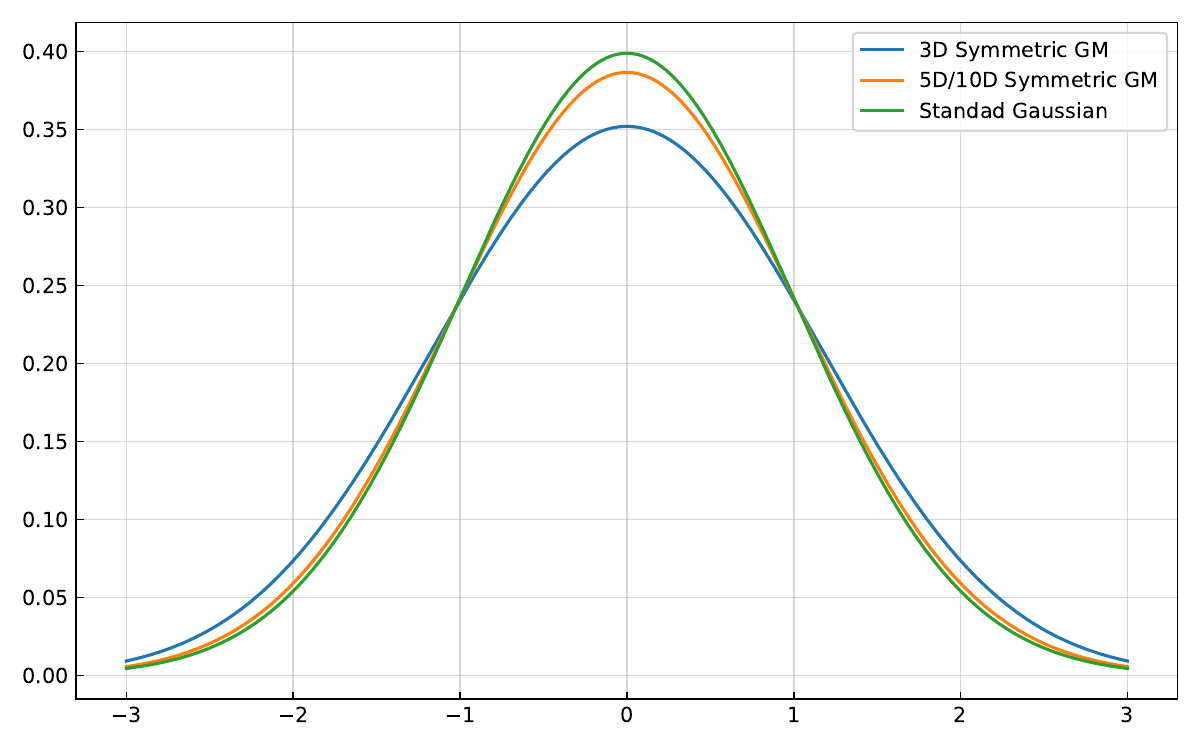}
    \end{tabular}
    \caption{Comparison between symmetric Gaussian mixture noise and the standard Gaussian noise.}\label{fig:symmetric_gm}
\end{figure}

Figure~\ref{fig:symmetric_gm} demonstrates the comparison between the marginal distribution for some selected dimension of our symmetric Gaussian mixture noise and the standard Gaussian noise.

\subsection{Asymmetric Gaussian mixture noise}\label{app:agm}
\begin{figure}[h]
    \centering
    \begin{tabular}{cc}
\includegraphics[width=0.7\textwidth,clip=false,trim=0.2cm 0.1cm 0.4cm 0.6cm]{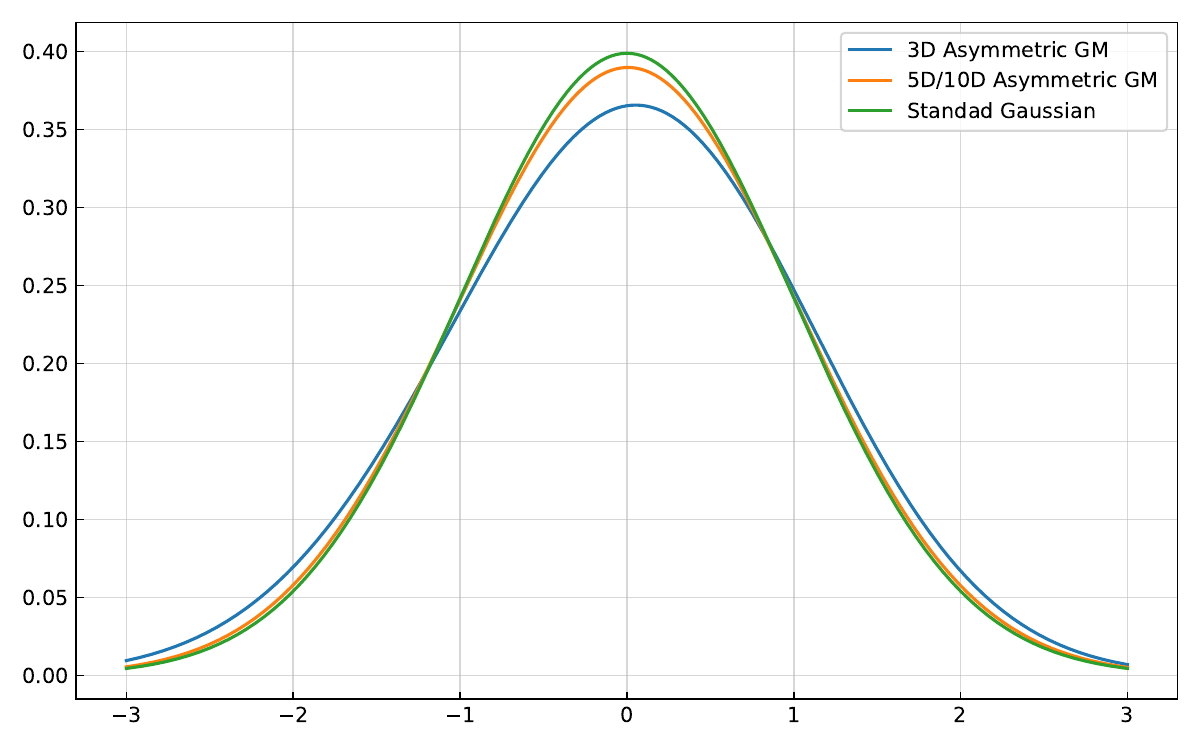}
    \end{tabular}
    \caption{Comparison between asymmetric Gaussian mixture noise and the standard Gaussian noise.}\label{fig:asymmetric_gm}
\end{figure}

We consider an asymmetric Gaussian mixture noise which is given by
\begin{equation*}
    p_w(w_t)=\frac{1}{(2\pi)^{n/2}}\left((1-\gamma) e^{\frac{-|w_t-\gamma a|^2}{2}} + \gamma e^{\frac{-|w_t+(1-\gamma) a|^2}{2}}\right),
\end{equation*}
where $\gamma=\frac{1}{4}$ and $a = [1,1,1]^{\top}$, $[\frac{1}{2},\frac{1}{2},\frac{1}{2},\frac{1}{2},\frac{1}{2}]^{\top}$ and $[\frac{1}{2},\frac{1}{2},\frac{1}{2},\frac{1}{2},\frac{1}{2},\frac{1}{2},\frac{1}{2},\frac{1}{2},\frac{1}{2},\frac{1}{2}]^{\top}$ for $n=3$, $5$ and $10$ respectively. Taking gradients,
\begin{equation*}
    -\nabla \log p_w(w_t)=w_t-\gamma a+\frac{\gamma a}{\gamma + (1-\gamma)ke^{w_t^\top a}},
\end{equation*}
and
\begin{equation*}
    \begin{split}
        -\nabla^2 \log p_w(w_t)&=I_n-\gamma(1-\gamma)aa^\top\frac{ke^{w_t^\top a}}{(\gamma + (1-\gamma)ke^{w_t^\top a})^2}\\
        &\succeq I_n-\frac{1}{4}aa^\top\\
        &\succeq \left(1-\frac{|a|^2}{4}\right)I_n,
    \end{split}
\end{equation*}
where $k=\exp((1-2\gamma)|a|^2/2)$. Therefore, the first condition in Assumption~\ref{ass:noise} is satisfied for $n=3$, $5$, and $10$ as in Section~\ref{app:gm}. Note that if we set $\gamma=\frac{1}{2}$, we recover the symmetric Gaussian mixture noise defined in Section~\ref{app:gm}. Figure~\ref{fig:asymmetric_gm} demonstrates the comparison between the marginal distribution for some selected dimension of our symmetric Gaussian mixture noise and the standard Gaussian noise.

\bibliographystyle{IEEEtran}

\bibliography{ref}

\end{document}